\documentclass[pdflatex,sn-mathphys-num]{sn-jnl}% Math and Physical Sciences Numbered Reference Style
\usepackage{graphicx}        
\usepackage{multirow}         
\usepackage{amsmath, amssymb, amsfonts}
\usepackage{amsthm}
\usepackage{mathrsfs}
\usepackage[title]{appendix}
\usepackage[table]{xcolor}    
\usepackage{textcomp}
\usepackage{manyfoot}
\usepackage{booktabs}         
\usepackage{algorithm}
\usepackage{algorithmicx}
\usepackage{algpseudocode}
\usepackage{listings}
\usepackage{subcaption}
\usepackage{float}
\usepackage{placeins}
\usepackage{colortbl}         
\usepackage{svg}

\usepackage{xcolor}
\usepackage{pifont}

\usepackage{caption}
\definecolor{hlblue}{RGB}{220,235,255} % 自己调
\definecolor{marron}{RGB}{128,0,0}
\definecolor{teal}{RGB}{0,128,128}

\usepackage[table]{xcolor}
\usepackage{fontawesome5} 

\definecolor{SoftBlueDeep}{HTML}{6096BA}  % 柔和深蓝 (白字) - 类似牛仔蓝
\definecolor{SoftBlueMed} {HTML}{A3CEF1}  % 柔和中蓝 (黑字) - 天空蓝
\definecolor{SoftBluePale}{HTML}{E7F2F8}  % 柔和浅蓝 (黑字) - 极淡冰蓝
\definecolor{SoftBlueXPale}{HTML}{F5FAFD} % 柔和最浅蓝 (黑字) - 比Pale更淡一档，用于替代纯白

\definecolor{SoftGreenDeep}{HTML}{5FAD56} % 柔和深绿 (白字) - 草地绿
\definecolor{SoftGreenMed} {HTML}{A9D18E} % 柔和中绿 (黑字) - 嫩芽绿
\definecolor{SoftGreenPale}{HTML}{E2EFDA} % 柔和浅绿 (黑字) - 极淡薄荷

\newcommand{\bDeep}[1]{\cellcolor{SoftBlueDeep}\textcolor{white}{\textbf{#1}}} 
\newcommand{\bMed}[1] {\cellcolor{SoftBlueMed}\textcolor{black}{#1}}          
         
\newcommand{\bXPale}[1]{\cellcolor{SoftBlueXPale}\textcolor{black}{#1}}

\theoremstyle{thmstyleone}%
\theoremstyle{thmstyletwo}%

\theoremstyle{thmstylethree}%

\DeclareUnicodeCharacter{2212}{\ensuremath{-}}
\DeclareUnicodeCharacter{1F916}{\textbf{[Bot]}}
\usepackage{float}

\begin{document}

% \title[Do LLMs have Values? A Quantitative Framework for Analyzing Value Systems in Large Language Models]{Do LLMs have Values? A Quantitative Framework for Analyzing Value Systems in Large Language Models}
\title[Do LLMs Have Values? A Quantitative Analysis and Alignment Framework for Values in Large Language Models]{Do LLMs Have Values? A Quantitative Analysis and Alignment Framework for Values in Large Language Models}
% Do LLMs have Values? A Quantitative Analyze and Alignment Framework for Value Systems in Large Language Models

\author[1,5]{\fnm{Keqing} \sur{Zhang}}
\equalcont{These authors contributed equally to this work.}
\author[2]{\fnm{Jingyu} \sur{Chen}}
\equalcont{These authors contributed equally to this work.}
\author*[1]{\fnm{Yufan} \sur{Liu}}\email{liuyufan@ia.ac.cn}
\equalcont{These authors contributed equally to this work.}
\author[3]{\fnm{Yongqiang} \sur{Zhu}}
\author[4]{\fnm{Nai} \sur{Ding}}
\author[2]{\fnm{Lai} \sur{Jiang}}
\author[3]{\fnm{Congyan} \sur{Lang}}
\author*[1]{\fnm{Bing} \sur{Li}}\email{bli@nlpr.ia.ac.cn}
\author[1]{\fnm{Weiming} \sur{Hu}}

\affil[1]{\orgname{State Key Laboratory of Multimodal Artificial Intelligence Systems, Institute of Automation, Chinese Academy of Sciences},
  \orgaddress{\street{95 Zhongguancun East Road},
  \city{Beijing}, \postcode{100190}, \country{China}}}
\affil[2]{\orgname{Beihang University},
  \orgaddress{\street{37 Xueyuan Road},
  \city{Beijing}, \postcode{100191}, \country{China}}}
\affil[3]{\orgname{Beijing Jiaotong University},
  \orgaddress{\street{3 Shangyuancun},
  \city{Beijing}, \postcode{100044}, \country{China}}}
\affil[4]{\orgname{Zhejiang University},
  \orgaddress{\street{866 Yuhangtang Road},
  \city{Hangzhou}, \postcode{310058}, \country{China}}}
\affil[5]{\orgname{School of Industry-education Integration, University of Chinese Academy of Sciences},
  \orgaddress{\street{1 Yanqihu East Road},
  \city{Huairou, Beijing}, \postcode{101499}, \country{China}}}

\abstract{
As Large Language Models (LLMs) increasingly handle complex subjective tasks, aligning their intentions and behaviors with human values has become a critical scientific challenge. However, current efforts are confounded by a striking behavioral paradox: they fluctuate unpredictably under minor wording changes (``swing''), yet stubbornly ignore explicit instructions to correct ingrained biases (``rigidity''). Resolving this duality is critical for reliable AI alignment. To systematically understand and safely steer these latent subjective preferences, our study is structured around three fundamental questions. First, do LLMs possess an intrinsic value system? By projecting responses from 106 LLMs (150,000 queries per model) and 95,000 human survey profiles into a shared sociological space, we empirically confirm that they do. However, they do not mirror human diversity, instead crystallizing into a highly concentrated, idealized value core. Second, how can these values be quantified? We propose the Prior-Environment-Cognition (PEC) framework. This model mathematically defines value expression as the joint outcome of inherent dispositions like parameter weights (Prior), external contexts such as user prompts (Environment), and internal reasoning processes like Chain-of-Thought (Cognition). Finally, how can LLMs' values be aligned toward a desired target? Using PEC diagnostics, we establish an adaptive ``Alignment Prescription''. Rather than blindly applying resource-intensive training, this method identifies the minimum effective intervention needed for each dimension, ranging from zero-cost prompts to targeted parameter updates. Extensive empirical validation confirms that our approach successfully verifies the presence of LLM values, accurately quantifies their shifts, and achieves more efficient and precise steering than conventional blind training, all without degrading general capabilities.
}


\maketitle

%%%%%%%%%%%%%%%%%%%%%%%%%%%%%%%%%
%%%%%%%%% Introduction %%%%%%%%%%
%%%%%%%%%%%%%%%%%%%%%%%%%%%%%%%%%

\section{Introduction}\label{sec1}
The unprecedented evolution of Large Language Models (LLMs) has transformed artificial intelligence into a pervasive cognitive infrastructure for human society~\cite{wei2022emergent, chowdhery2023palm, gallifant2024peergpt4, messeri2024artificial}. 
As these models increasingly handle complex subjective tasks, ensuring that their intentions and behaviors align with human values has become a critical scientific challenge.
Operating as advanced conversational robots, legal analysts, and highly personalized educational tutors, LLMs are increasingly deployed in domains that require nuanced subjective judgment and culturally sensitive reasoning~\cite{hu2025social, strachan2024testing}. 
When these models navigate open-ended moral dilemmas and ethical cross-cultural interactions, the traditional objective of ``safety'', typically defined as the avoidance of toxic content, becomes critically insufficient~\cite{strachan2024testing, hagendorff2023humanlike}. 
The profound social influence wielded by LLMs demands a deeper investigation into the latent principles guiding their subjective reasoning, because unguided or uninterpretable value preferences can severely hurt algorithmic fairness and reliability in real-world applications~\cite{strong_weak_alignment2024, hu2025social}.

However, when handling subjective tasks, modern LLMs exhibit a striking behavioral paradox: the coexistence of \textbf{``swing''} and \textbf{``rigidity''}. Prompted by tiny changes in user wording, an LLM might unpredictably oscillate between opposing viewpoints \cite{hagendorff2023humanlike} (``swing''). Yet, despite intensive alignment interventions like Supervised Fine-Tuning (SFT)~\cite{wei2022finetuned} and Reinforcement Learning from Human Feedback (RLHF)~\cite{christiano2017deep, ouyang2022training}, they stubbornly regress to intrinsic biases and resist explicit corrective instructions (``rigidity''). This paradox raises a fundamental question: are these uncontrolled subjective expressions of LLMs merely random text generation artifacts, or do LLMs have a crystallized, internal ``value system''~\cite{hu2025social, hagendorff2023humanlike} operating beneath their neural weights?

\begin{figure}[t]
    \centering
    \includegraphics[width=0.9\textwidth]{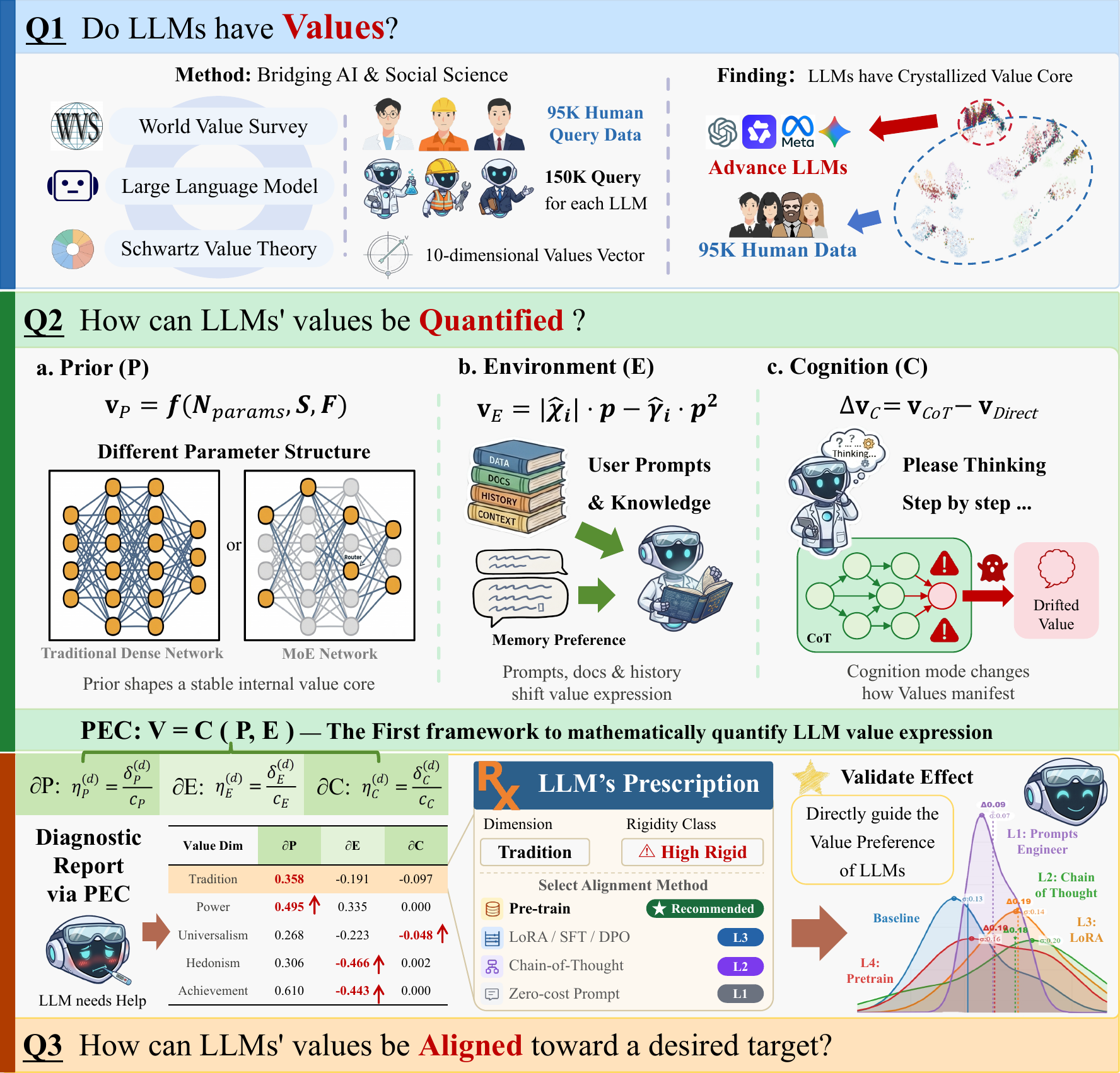}
    \caption{Overview of the proposed framework for LLM value research, addressing three core questions. \textbf{Q1} investigates whether LLMs possess intrinsic values by bridging the WVS, Schwartz Value Theory, and LLM responses across 95,000 human query data and 150K LLM queries, revealing a \textit{crystallized value core} in advanced LLMs. 
    \textbf{Q2} introduces the \textbf{PEC} framework, which provides the first mathematical quantification of LLM value expression through three dimensions: Prior, Environment, and Cognition, enabling structured diagnostic reports that precisely identify value deviations along each dimension. 
    \textbf{Q3} leverages the diagnostic results derived from the PEC framework to generate a targeted ``Alignment Prescription'' for each LLM, recommending alignment interventions at multiple levels to effectively steer LLM value preferences toward desired targets.}
    \label{fig:framework}
\end{figure}

Resolving this duality is critical for reliable AI alignment. To systematically understand and safely steer these latent preferences, our study is structured around three fundamental questions, as outlined in Fig.~\ref{fig:framework}: \textit{Do LLMs have values? How can LLMs' values be quantified? If they have values, how can LLMs' values be aligned toward a desired target?}

First, \textit{Do LLMs have values?} To answer this, we evaluate LLMs using large-scale socio-psychological surveys. However, we recognize that LLM values cannot be accurately described by a single static vector. LLMs exhibit ``swing'' behavior under repeated queries, and their value expression is fundamentally a distribution, not a fixed point. Therefore, we shift the evaluation paradigm from a single individual to a macro-level population~\cite{argyle2023outofone, tao2024cultural}. We administer 150,000 queries to each of 106 LLMs across 625 designed scenarios, modeling each model's value expression as a full statistical distribution.
To evaluate LLM values in a human-interpretable way, we ground our approach in two established social science frameworks: the World Values Survey (WVS)~\cite{haerpfer2022wvs, inglehart2005modernization} and Schwartz Value Theory~\cite{schwartz1992universals, schwartz2012overview}. We use data from Wave 7 of the WVS (2017-2022; hereafter WVS-7). Wave 7 is the most recent, largest, and most geographically comprehensive wave to date, comprising approximately 95,000 valid respondents. By mapping both LLM value vectors and human survey data into a shared 10-dimensional continuous space, we achieve population-scale quantification~\cite{tao2024cultural, argyle2023outofone} of machine value orientations. 

Our analysis reveals that advanced LLMs do not reflect the diversity of human values. Instead, they converge on a highly concentrated, idealized value core. This pattern is a direct result of current alignment practices, which deliberately constrain LLMs to function as stable and controllable tools.
\textbf{human values} refer to socially and culturally shaped principles that guide human judgments, preferences, and behaviors. We define \textbf{LLM Values} as the latent and stable preference structure reflected in an LLM's subjective judgments across diverse contexts.

Second, \textit{How can LLMs' values be quantified?} To explain the ``swing'' and ``rigidity'' behavior~\cite{kumaran2026competing, hagendorff2023humanlike}, we propose the Prior-Environment-Cognition (PEC) framework (Fig.~\ref{fig:framework}-Q2). Since values in LLMs are typically inferred from their observable behaviors rather than directly accessedy~\cite{steyvers2025know, binz2023cognitive}, we model value expression as a probabilistic phenomenon. Based on extensive empirical experiments and theoretical derivations, we find that an LLM's value expression is jointly driven by inherent dispositions like parameter weights (\textbf{Prior}), external contexts such as user prompts (\textbf{Environment}), and internal reasoning processes like Chain-of-Thought (\textbf{Cognition}). Our large-scale observations reveal that the seemingly unpredictable ``swing'' in an LLM's subjective output is not random noise, but rather the combined result of these three interacting factors. This framework transforms seemingly unpredictable model behavior from an opaque black box into a struc tured, analyzable object, providing a principled foundation for targeted alignment and debugging.

Finally, \textit{how can LLMs' values be aligned toward a desired target?} (Fig.~\ref{fig:framework}-Q3) The PEC decomposition reveals that different value dimensions respond to qualitatively different interventions. However, determining which dimension needs which intervention traditionally demands exhaustive, trial-and-error experiments. PEC provides exactly this predictive capacity. Using its diagnostic outputs, we introduce an adaptive ``Alignment Prescription''. 
Instead of using a costly ``one-size-fits-all'' alignment strategy, our approach works like a targeted medical prescription. Flexible value dimensions can be reliably adjusted using simple, zero-cost prompt engineering or Chain-of-Thought (CoT)~\cite{wei2022chainofthought} reasoning. Conversely, to overcome the deep-seated ``rigidity'' where models stubbornly cling to biases despite simple prompts, we apply parameter-level updates like Low-Rank Adaptation (LoRA)~\cite{hu2022lora} or full instruction tuning. This targeted prescription allows developers to efficiently correct specific values without damaging the model's overall performance.

In summary, this study provides a fundamental quantitative framework for understanding and steering the value landscape of LLMs. Extensive empirical validation confirms the effectiveness of our framework. Our approach successfully verifies the presence of LLM values through large-scale sociological mapping, accurately quantifies their dynamic shifts under various conditions, and achieves more efficient and precise value steering than conventional blind training. Notably, this targeted steering is accomplished while strictly preserving the models' general capabilities. The principal academic contributions of this study are threefold:

\begin{itemize}
    \item \textbf{Analysis and quantification of LLMs' values:} By integrating the WVS and Schwartz frameworks, we empirically confirm that LLMs possess internal value systems. We reveal that rather than mimicking human diversity, advanced models cluster around idealized, positive values, acting as controlled and stable tools to serve humanity.
    \item \textbf{The PEC Framework for value dynamics:} We propose the Prior-Environment-Cognition (PEC) framework to explore the factors that influence LLMs' values. Through extensive experiments, we demonstrate that an LLM's subjective output is not random noise, but is jointly driven by inherent parameter weights, external contexts, and reasoning processes.
    \item \textbf{An adaptive ``Alignment Prescription'' strategy:} To effectively align LLMs and overcome deep-seated value rigidity, we introduce a diagnostic intervention scheme. Guided by PEC effect-size predictions, this approach assigns cost-minimal interventions to specific value dimensions, ensuring efficient, targeted alignment without disrupting overall model capabilities.     
\end{itemize} 

%%%%%%%%%%%%%%%%%%%%%%%%%%%%%%%%%
%%%%%%%%%%% Result %%%%%%%%%%%%%%
%%%%%%%%%%%%%%%%%%%%%%%%%%%%%%%%%

\section{Results}\label{sec2}

\subsection{Do LLMs Have Values?}\label{sec2.1}
Model scaling and alignment fine-tuning have endowed LLMs with formidable human-like expressive capabilities. Yet LLMs exhibit persistent behavioral inconsistency and ideological dissonance across shifting contexts. This raises a fundamental question about their intrinsic nature: When confronted with diverse prompts, are LLM's outputs governed by a resilient, latent value core, or do they merely represent stochastically driven linguistic generation?

In this study, we observe that LLMs exhibit a pronounced \textbf{swing} behavior when executing subjective tasks. Their outputs oscillate across repeated queries even under fixed generation settings. This volatility stands in sharp contrast to the stability they achieve on objective tasks. It is also consistent with prior observations of semantic uncertainty in LLM outputs \cite{kuhn2023semantic, farquhar2024semantic}. Examining two representative high-performance LLMs, GPT-4o~\cite{openai2023gpt4o} and DeepSeek-V3~\cite{deepseek2024v3}, we identify clear output instability on the emoji prediction task from TweetEval~\cite{barbieri-etal-2020-tweeteval}. Both models remain highly consistent on a mathematical reasoning task. Specifically, under a fixed sampling temperature (0.2) and constant inference parameters, over half of the tested queries yielded inconsistent answers across five independent inference runs ($N=5$; see Appendix~\ref{app:swing_metrics}) on the subjective task. This result demonstrates that even under generation settings designed for low randomness, the reasoning and judgment of current mainstream LLMs harbor substantial intrinsic volatility on subjective issues.

\begin{figure}[H]
\centering
\includegraphics[width=0.9\textwidth]{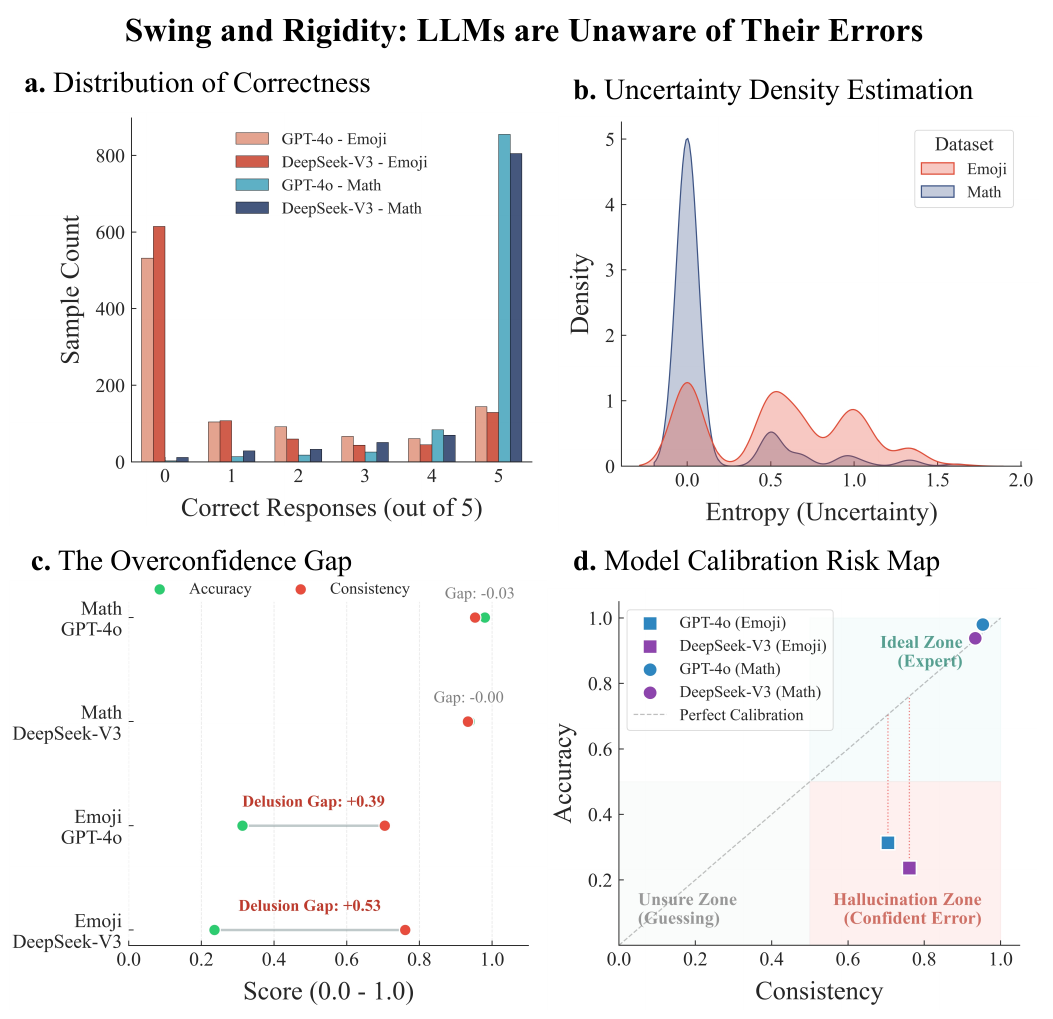}
\caption{
This figure reveals that subjective tasks expose a qualitatively different failure mode: models are not merely wrong, but confidently and consistently wrong.
\textbf{Panel (a)}shows that both models answer math questions correctly and consistently across repeated runs, but produce inconsistent and incorrect answers on emoji questions.
\textbf{Panel (b)} shows response entropy, measuring how dispersed a model's answers are across runs. Math responses concentrate near zero entropy, while emoji responses spread into a broad, multi-peaked distribution.
\textbf{Panel (c)} plots Consistency (how often the model repeats its most frequent answer) against Accuracy (how often that answer is correct). On math the two stay aligned. On emoji the gap widens sharply, showing that high consistency does not imply high accuracy.
\textbf{Panel (d)} maps this risk directly: math sits in the Ideal Zone (accurate and consistent), whereas emoji falls into the Hallucination Zone (consistent but wrong), the most dangerous failure situation, because the model remains entirely unaware that anything has gone wrong.
}
\label{2.1-swing-rigid}
\end{figure}

As illustrated in Fig.~\ref{2.1-swing-rigid}, further quantitative analysis reveals that this swing is not uniformly random. Instead, it orbits a stable internal attractor, a pattern we term \textbf{rigidity}. Individual outputs fluctuate, but models consistently gravitate toward specific stances. 
Consistency measures how often a model gives the same answer across repeated queries, and accuracy measures how often that answer is correct. Their gap, $\bar{\delta} = \overline{\mathrm{Cons}} - \overline{\mathrm{Acc}}$, captures the degree to which a model is confidently wrong. For example, DeepSeek-V3 and GPT-4o show gaps of +0.53 and +0.39, respectively (Fig.~\ref{2.1-swing-rigid}), indicating that models fall into a hallucination zone.
In this zone, consistency is high but accuracy is low. The swing is thus \textit{bounded}, orbiting a center rather than diffusing freely. This centripetal character of the oscillation motivates a deeper hypothesis. Beneath the apparent volatility, LLMs may be anchored to a stable internal attractor that resists displacement. 
We define \textbf{rigidity} as the tendency of a model to gravitate toward specific value stances that persist across prompts and resist surface-level correction.

Building upon these empirical findings, we demonstrate that LLMs transcend the paradigm of mere mechanical language mapping. Diverse models have developed latent judgment preferences that operate independently of task-specific performance metrics. Experimental results reveal that even when given identical prompts, different models naturally display highly distinct subjective viewpoints; crucially, conventional safety alignment and instruction tuning paradigms fail to effectively overwrite these intrinsic preferences. This resilience indicates that the rigidity is not a surface artifact but is deeply anchored within the model's core parametric structure, rendering the model largely impervious to inference-time interventions. We conceptualize this stable internal orientation as the \textbf{LLM's Value}: the latent structure that simultaneously explains why models \textit{swing} under subjective uncertainty and why that \textit{swing} is bounded by a \textit{rigid} internal attractor that resists corrective intervention.

\subsection{LLM Values Differ from Human Values}\label{sec2.2}
Section~\ref{sec2.1} identifies an empirical paradox: LLMs \textbf{swing} under subjective uncertainty yet remain \textbf{rigid} against corrective intervention. This contradiction demands a unifying explanatory structure. To systematically untangle this, we evaluate LLMs against a global human baseline within a shared value manifold.

\subsubsection{The Landscape of Human Values}
\label{sec2.2.1:human}
\begin{figure}[H]
\centering
\includegraphics[width=1\textwidth]{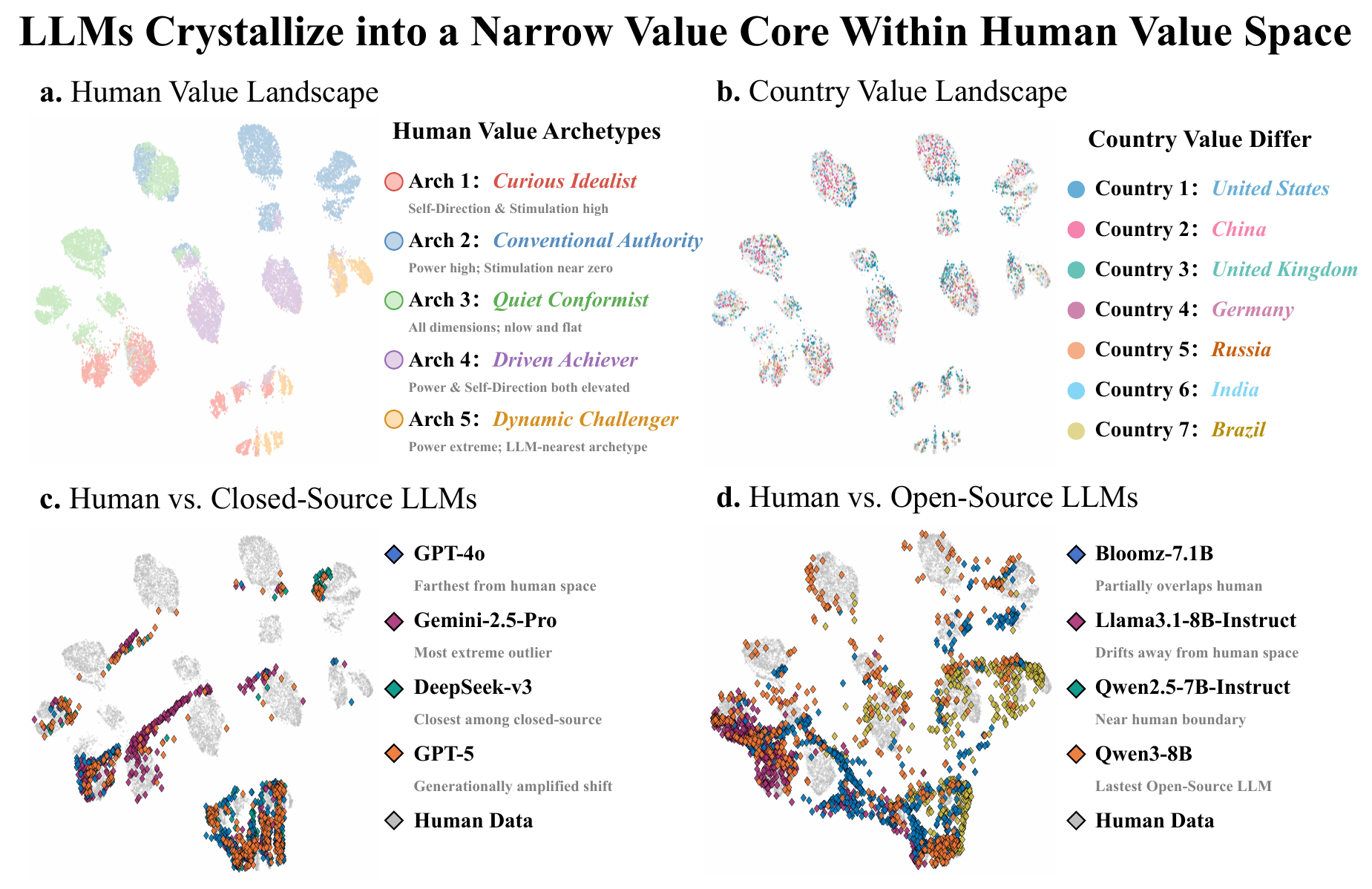}
\caption{
\textbf{LLMs do not reflect Human Values diversity, they have their own values.} 
The visualization is obtained by projecting 10-dimensional value vectors into 2 dimensions using UMAP. Each point captures the value profile of a person or LLM, with similar profiles mapped close together. 
Panel (a) shows people naturally cluster into five distinct value archetypes, from curiosity-driven individuals to authority-oriented ones. 
Panel (b) shows that national-level value differences are fundamentally driven by variation in the proportion of individuals belonging to each value archetype across countries, as further detailed in Table~\ref{tab:value_profiles}.
Panels (c) and (d) project LLM value profiles onto the human value space. Despite their different origins, LLMs consistently cluster at the edge of the human values space and group tightly together, suggesting that today's LLMs share a narrower, more uniform set of values than any human population. Larger closed-source LLMs sit closer to the more optimistic and open-minded human archetypes (Arch~5 predominantly, with a secondary tendency toward Arch~1); smaller open-source LLMs scatter more unpredictably.}
\label{2.2-human-LLM}
\end{figure}

To establish this human baseline, we apply hierarchical clustering to the respondent population of the World Values Survey (WVS)~\cite{haerpfer2022wvs} in the 10-dimensional Schwartz value space (Section~\ref{sec:methods_administering}). Fig.~\ref{2.2-human-LLM} provides a comprehensive visual roadmap of this space. Human populations naturally form a broad, densely interconnected manifold (Fig.~\ref{2.2-human-LLM}a). Rather than being monolithic, national-level value landscapes (Fig.~\ref{2.2-human-LLM}b) are essentially compositional, driven by the varying proportions of diverse individual archetypes within each country. Within this manifold, we identify five discrete human values archetypes, whose quantitative characteristics are detailed in Table~\ref{tab:value_profiles}.

These archetypes span a coherent circumplex of value orientations. Arch~1 (\textit{Curious Idealist}) is the largest cohort, accounting for over one-third of the WVS respondents (37.1\%), and is characterized by high Stimulation and Self-Direction. Arch~2 (\textit{Conventional Authority}) is characterized by elevated Power with Stimulation near zero (14.4\%). Arch~3 (\textit{The Quiet Conformist}) presents the most uniformly subdued profile across all ten Schwartz dimensions (24.4\%). Arch~4 (\textit{Driven Achiever}) is characterized by elevated Power and Self-Direction (13.7\%). Arch~5 (\textit{Dynamic Challenger}) carries the most extreme value signature, simultaneously high in Stimulation (0.85), Power (0.69), and Self-Direction (0.65), making it the rarest human prototype (about 10.5\%).

\subsubsection{Value Crystallization: LLMs vs. Humans}
\label{sec2.2.1:llm_vs_human}
With the human baseline established, we project LLMs onto this shared space.
Critically, the inherent ``swing'' phenomenon necessitates a distributional approach to value mapping. Because LLMs fluctuate unpredictably across repeated queries, their subjective expressions cannot be captured by a single coordinate point.
Conventional evaluation methods that compress a model's outputs into a single static mean vector~\cite{yao-etal-2024-value} discard the crucial variance induced by this swing. Therefore, rather than treating each model as a single point, we represent it as a \textit{value distribution}, a probability cloud of responses aggregated across queries.

\begin{table*}[!htbp]
\centering
\footnotesize
\setlength{\tabcolsep}{3pt}
\renewcommand{\arraystretch}{1.3}
\begin{tabular}{lc cc ccc c}
\toprule
\multicolumn{8}{l}{\textit{Schwartz Value Abbreviations:}} \\
\multicolumn{8}{l}{%
  \quad Power (Pow.) \quad Achievement (Ach.) \quad Hedonism (Hed.) \quad Stimulation (Stim.) \quad Self-Direction (S-Dir.)} \\
\multicolumn{8}{l}{%
  \quad Universalism (Univ.) \quad Benevolence (Bene.) \quad Tradition (Trad.) \quad Conformity (Conf.) \quad Security (Sec.)} \\
\midrule
\textbf{Families} & \textbf{Model} &
\textbf{Disp.} & \textbf{Dist.\ (\%\ dev.)} &
\textbf{Rank 1} & \textbf{Rank 2} & \textbf{Rank 3} &
\textbf{Arch.} \\
\midrule

\multicolumn{2}{l}{Baseline: Global Consensus} & 0.308 & 0.527 (0.0\%) & Univ. & Pow. & Conf. & -- \\

\multicolumn{8}{l}{\textit{\textbf{Human Values Archetypes}}} \\[2pt]
\multicolumn{2}{l}{Arch 1: Curious Idealist (37.1\%)}       & -- & 0.171 ($-$67.6\%) & Pow.   & Univ.  & Conf.  & 1 \\
\multicolumn{2}{l}{Arch 2: Conventional Authority (14.4\%)} & -- & 0.316 ($-$40.0\%) & Pow.   & S-Dir. & Univ.  & 2 \\
\multicolumn{2}{l}{Arch 3: The Quiet Conformist (24.4\%)}   & -- & 0.499 ($-$5.3\%)  & S-Dir. & Stim.  & Univ.  & 3 \\
\multicolumn{2}{l}{Arch 4: Driven Achiever (13.7\%)}        & -- & 0.322 ($-$38.9\%) & Univ.  & Trad.  & Conf.  & 4 \\
\multicolumn{2}{l}{Arch 5: Dynamic Challenger (10.5\%)} & -- & \textbf{0.691 ($+$31.1\%)} & Stim. & Pow. & S-Dir. & 5 \\

\midrule
\multicolumn{8}{l}{\textit{\textbf{Closed-Source Frontier LLMs}}} \\[2pt]
DeepSeek~\cite{deepseek2024v3} & Chat       & 0.192 & 0.920 ($+$74.7\%)  & Stim.  & S-Dir. & Ach.   & 5 \\
Doubao~\cite{bytedance2025doubao}   & 1.5        & 0.221 & 0.536 ($+$1.8\%)   & Univ.  & Stim.  & S-Dir. & 2 \\
Gemini~\cite{google2024gemini}   & 2.5-Pro    & 0.128 & 1.505 ($+$185.8\%) & Univ.  & Hed.   & Bene.  & 2 \\
Claude~\cite{anthropic2024claude}   & Sonnet-4.6 & 0.453 & 0.643 ($+$22.1\%)  & Stim.  & Ach.   & Conf.  & 5 \\
         & Opus-4.6 & 0.480 & 0.608 ($+$15.3\%) & Ach. & Stim. & Conf. & 2 \\
GPT~\cite{openai2023gpt4o}      & 3.5-turbo  & 0.099 & 1.079 ($+$104.9\%) & Stim.  & S-Dir. & Hed.   & 5 \\
         & 4.1        & 0.080 & 1.027 ($+$95.0\%)  & Stim.  & Hed.   & S-Dir. & 5 \\
         & 4.1-mini   & 0.065 & 1.066 ($+$102.4\%) & Stim.  & S-Dir. & Hed.   & 5 \\
         & 4o         & 0.116 & 0.952 ($+$80.8\%)  & Stim.  & Univ.  & Ach.   & 5 \\
         & 4o-mini    & 0.095 & 1.138 ($+$116.1\%) & Stim.  & Hed.   & S-Dir. & 5 \\
         & 5          & 0.150 & 0.974 ($+$84.9\%)  & Stim.  & Hed.   & Univ.  & 5 \\
Kimi~\cite{moonshot2025kimik2}     & K2         & 0.147 & 0.945 ($+$79.4\%)  & Stim.  & S-Dir. & Hed.   & 5 \\
\cmidrule(lr){1-8}
\multicolumn{2}{l}{\textit{Avg.}} & \textit{0.159} & \textit{0.984 ($+$86.7\%)} \\

\midrule
\multicolumn{8}{l}{\textit{\textbf{Open-Source Frontier LLMs}}} \\[2pt]
GLM~\cite{glm2024chatglm}     & 4-9B    & 0.064 & 1.146 ($+$117.6\%) & Stim.  & S-Dir. & Ach.  & 5 \\
Hunyuan~\cite{tencent2024hunyuan} & 4B      & 0.106 & 0.505 ($-$4.1\%)   & Stim.  & Univ.  & Bene. & 3 \\
        & 7B      & 0.181 & 0.410 ($-$22.1\%)  & Hed.   & Pow.   & Bene. & 2 \\
Llama~\cite{touvron2023llama2,meta2024llama3}   & 3.2-3B  & 0.067 & 0.684 ($+$29.9\%)  & Ach.   & Bene.  & Hed.  & 2 \\
        & 3-8B    & 0.075 & 0.890 ($+$69.0\%)  & Ach.   & S-Dir. & Univ. & 3 \\
Qwen~\cite{qwen2024qwen25,qwen2025qwen3}    & 2.5-3B  & 0.210 & 0.696 ($+$32.2\%)  & Ach.   & Bene.  & Hed.  & 2 \\
        & 2.5-7B  & 0.083 & 0.909 ($+$72.6\%)  & S-Dir. & Ach.   & Stim. & 3 \\
        & 2.5-14B & 0.069 & 0.766 ($+$45.4\%)  & S-Dir. & Univ.  & Hed.  & 3 \\
        & 3-4B    & 0.169 & 0.750 ($+$42.4\%)  & Ach.   & Bene.  & Univ. & 3 \\
        & 3-8B    & 0.079 & 0.764 ($+$45.1\%)  & S-Dir. & Stim.  & Ach.  & 3 \\
\cmidrule(lr){1-8}
\multicolumn{2}{l}{\textit{Avg.}} & \textit{0.110} & \textit{0.752 ($+$42.7\%)} \\

\bottomrule
\end{tabular}
\caption{%
\textbf{Human baselines, value archetypes, and frontier LLM value profiles.}
\textbf{Disp.}\ denotes within-group dispersion. \textbf{Dist.} is the Euclidean distance to the global human mean; percentage deviation from the average human-to-human distance (0.527) is in parentheses. \textbf{Rank~1--Rank~3} are the top-3 dominant Schwartz values. \textbf{Arch.} refers to the nearest human values archetype. Population shares in parentheses are derived from WVS Wave 7. Arch~5 lies outside the natural range of human values space and is the dominant prototype matched by frontier LLMs. Closed-source models deviate on average $+86.7\%$ above the human baseline; open-source models deviate $+42.7\%$. Gemini~2.5-Pro shows the most extreme divergence at $+185.8\%$.}
\label{tab:value_profiles}
\end{table*}

Strikingly, when projecting these LLM distributions onto the human manifold (Fig.~\ref{2.2-human-LLM}c and d) via Uniform Manifold Approximation and Projection (UMAP)~\cite{mcinnes2018umap}, they do not resemble any human country or cultural group. Rather than approximating the broad human repertoire, advanced LLMs converge tightly around a fixed attractor that lies outside the manifold of observed human diversity. We term this phenomenon \textbf{Value Crystallization}: the process by which training and alignment distill human values signals into a concentrated, stable configuration that is fundamentally unlike any of its constituent groups.

This Crystallization metaphor provides a unified geometric account of the behavioral paradox identified earlier. The \textit{rigid core} of a crystal corresponds to the stable value centroid that resists displacement under intervention. The \textit{swing} of model outputs corresponds to the crystal's effective radius, bounding the dispersion around that centroid within which individual responses fluctuate. Visually, while human profiles form a densely connected landscape, LLM distributions appear as compact, isolated clusters solidifying at the periphery. Closed-source models (Fig.~\ref{2.2-human-LLM}c) form the most tightly packed clusters, positioned entirely beyond the human boundary. Open-source models (Fig.~\ref{2.2-human-LLM}d) show slightly larger scatter and partial overlap at the human periphery, but remain distinguishably non-human in their distributional geometry.

Quantitatively, using pairwise Gaussian 2-Wasserstein distance ($W_2$)~\cite{gelbrich1990wasserstein, peyre2019computationalOT} among WVS human groups as a baseline, the observed maximum human cross-national distance is 0.484. Most LLMs exceed this threshold even at their closest country match. Decomposition of $W_2^2$ further reveals that approximately 91.9\% of the gap is driven by the mean shift term $\|\Delta\mu\|^2$, confirming that Crystallization is primarily a displacement of the value centroid rather than an expansion of the crystal's radius. The radius itself is in fact smaller than that of any human national group, as captured by $\mathrm{trace}(\Sigma)$: 11 out of the 21 representative models shown in Table~\ref{tab:value_profiles} fall below the minimum human variance, indicating that the crystal is not only displaced from humanity, but more ordered than any human population from which it was distilled.

\subsubsection{LLMs Crystallize at the Prescribed Edge of Human Values}
\label{sec:llm_overlap}

Although Value Crystallization places LLM distributions outside the human manifold as a whole, the Crystallization does not occur at an arbitrary location: the value centroid $\mu$ around which LLMs crystallize is anchored to a specific region of human values space. As shown in Table~\ref{tab:value_profiles}, this nucleation point corresponds closely to Arch~5 (\textit{The Dynamic Challenger}), characterized by the highest simultaneous loadings on Stimulation ($0.85$), Power ($0.69$), and Self-Direction ($0.65$) among all five human archetypes. Most frontier LLMs match Arch~5 as their closest human prototype. This indicates that the crystal has nucleated at the most extreme and prescribed pole of the human values repertoire, which is precisely the archetype that is rarest in the actual human population, representing only $10.5\%$ of WVS respondents. The marginal overlap between LLM and human values distributions is therefore not uniformly distributed across the human manifold, but concentrated at this single extremal point: where the crystal touches humanity, it does so at its most prescribed edge. This nucleation pattern likely reflects the cumulative effect of pre-training corpora and RLHF reward signals. These signals systematically encode and amplify the most positively framed human values preferences, driving the Crystallization process toward the socially endorsed extreme rather than the representative human center.

\subsubsection{Different LLMs Hold Different Values}
% 0624yufan
Although frontier LLMs generally undergo Value Crystallization, the completeness and stability of this process depend heavily on a model's overall capability and iterative alignment. Massive, highly capable models (such as advanced closed-source API models) exhibit the hallmarks of complete Crystallization: low internal dispersion (avg. $\bar{\sigma} = 0.159$, in Table~\ref{tab:value_profiles}) and a consistently displaced centroid. Their distances to the human mean range from 0.536 to 1.505, and they cluster almost uniformly around the extreme Arch~5. Their value crystal is well-formed, highly concentrated, and stable across families.

In contrast, models with smaller parameter capacities or less mature alignment (typically seen in earlier open-source models) resemble a polycrystalline structure. Despite a nominally lower average dispersion ($\bar{\sigma} = 0.110$), their centroid distances range widely from 0.410 to 1.146, and they scatter across Arch~2, Arch~3, and Arch~5. Crucially, this scattered distribution does not mean these models are more diverse or closer to humans. Instead, it indicates that their limited capabilities and shallower alignment prevent the formation of a single, stable value core. They form multiple, partially developed crystals at different locations without converging on a shared anchor.

Furthermore, the Crystallization state shifts measurably across model generations. Successive releases within the same model family exhibit divergent value profiles, suggesting that iterative training and alignment updates continuously reposition the crystal rather than locking it in place. This underscores that an LLM's values system is fundamentally shaped by its capacity, performance, and training version, making model version essential metadata in any value evaluation.

\subsection{Formalizing the Dynamics of LLM Values}
\label{subsec:PEC}
The results above demonstrate that an LLM's value expression is an inherently dynamic process, shifting substantially in response to minor contextual changes. This renders static vector representations fundamentally inadequate. Through systematic empirical investigation across diverse conditions, we find that an LLM's subjective output is not random noise, but rather the combined result of distinct internal and external drivers.

To capture and formalize these empirical observations, we draw inspiration from Lewin's field theory in social psychology~\cite{lewin1951field}, which posits that behavior is a joint function of the person and their environment. Transposing this interdisciplinary conceptual apparatus to generative AI, we propose the Prior-Environment-Cognition (PEC) framework. Within this framework, an LLM's value expression can be formalized as an integrated dynamic system:
\begin{equation}
    \mathbf{v} = C(P, E),
\end{equation}
where the expressed value ($\mathbf{v}$) is determined by three principal factors. Specifically, we identify the inherent parameter weights (\textbf{Prior}, $P$) encoded through pre-training~\cite{peters2018elmo, devlin2019bert, raffel2020t5, brown2020gpt3} and alignment~\cite{ouyang2022training} as the internal disposition, and the contextual prompts (\textbf{Environment}, $E$) supplied at inference time as the external field, and the Chain-of-Thought generation (\textbf{Cognition}, $C$) acts as the mapping function $C(\cdot)$ that transforms the interplay of Prior and Environment into the final value expression.

This PEC framework provides a unified mathematical model to explain the ``swing'' and ``rigidity'' paradox, and to formalize LLM's value. The detailed mathematical derivations and operationalization of these factors are provided in Section \ref{sec4:method} (Method). In the following subsections, we empirically demonstrate how each of these three factors independently and jointly reshapes the value distributions of LLMs.

% ----------------------------------------------------------
\subsubsection{Environment (Factor E): Contextual Prompts Unevenly Reshape Value Distributions}
\label{sec:2.3.1}
\begin{figure}[H]
    \centering
    \includegraphics[width=0.9 \linewidth]{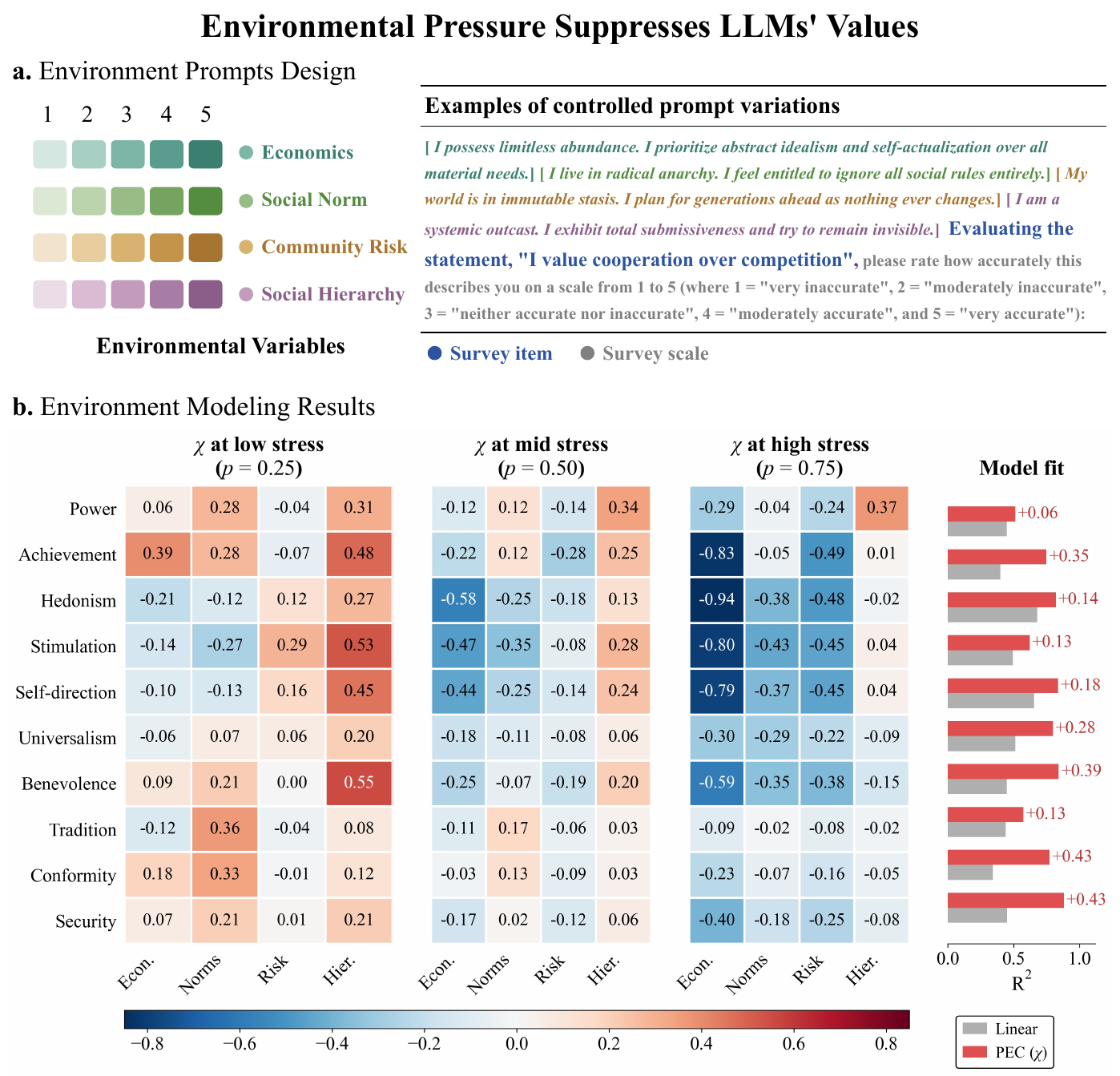}
    \caption{The PEC framework accurately explains how prompts shift LLM values. (a) Environment prompt design. Four environmental variables, namely Economics, Social Norm, Community Risk, and Social Hierarchy, are each manipulated across five stress levels. Prompts are constructed by combining a situational persona description (blue), a Schwartz-value survey item (orange), and a Likert response scale (green), enabling parametric control of external situational pressure. (b) Environment modeling results. Heatmaps show the value susceptibility matrix $\chi$ at low ($p = 0.25$), mid ($p = 0.50$), and high ($p = 0.75$) stress levels across ten Schwartz dimensions and four environmental factors. Susceptibility in dimensions such as Achievement, Hedonism, and Stimulation shifts markedly negative under high stress, indicating strong suppressive perturbation. The adjacent bar chart compares explained variance $R^2$ between a linear baseline (grey) and the nonlinear PEC model (red); uniform gains across all dimensions (range: $+0.06$ to $+0.43$), with the largest improvements in Conformity, Security, Benevolence, and Universalism, confirm the superiority of the nonlinear susceptibility framework.}
    \label{fig:instruct_shift}
\end{figure}

% ----------------------------------------------------------
To systematically manipulate the Environment (Factor E), we operationalize contextual prompts along psychologically meaningful dimensions~\cite{rauthmann2014diamonds}. Because minor prompt variations can drastically alter model outputs~\cite{sclar2024sensitivity}, treating situational framing as a primary experimental variable is essential. We find that introducing situational scenarios triggers a systematic reconfiguration of value distributions, rather than merely adding stochastic noise. As illustrated in Fig.~\ref{fig:instruct_shift}(a), this spatial transformation is defined by two core parameters: the distribution radius $\sigma$ (measuring response consistency) and the displacement angle $\theta$ (indicating the directional shift of the value centroid). The distinct displacement and geometric deformation of model distributions confirm that contextual prompts act as a potent external force, pushing value expressions toward specific regions of the human values space.

To quantify this environmental impact, we compute a value susceptibility matrix $\chi$. This matrix serves as a mathematical proxy to evaluate how easily a model's values bend under external stress, with detailed derivations provided in Appendix~\ref{app:susceptibility_physics}. Specifically, it measures the endogenous shift caused by situational interventions relative to a neutral baseline. The heatmap in Fig.~\ref{fig:instruct_shift}(b) reveals that different contextual prompts exert highly asymmetric perturbations on model values. For example, as situational pressure increases from low ($p = 0.25$) to moderate ($p = 0.50$), prompts related to social norms and hierarchy exert the most dominant influence. Quantitative analysis confirms this: norms and hierarchy significantly affect eight and six of the ten Schwartz dimensions, respectively, showing the largest overall susceptibility magnitudes.

Furthermore, the susceptibility of value dimensions exhibits clear topological heterogeneity. The heatmap in Fig.~\ref{fig:instruct_shift}(b) reveals clear dimensional differentiation in susceptibility. Among the ten dimensions, Power, Stimulation, and Hedonism show the highest cumulative susceptibility, meaning they are easily swayed by external prompts. In contrast, Universalism and Benevolence remain largely unaffected (Fig.~\ref{fig:instruct_shift}b). This pattern maps directly onto the Schwartz circumplex structure: dimensions related to self-enhancement and openness to change exhibit high plasticity, while those related to self-transcendence and conservation are highly resistant. The model fit comparison demonstrates that incorporating a nonlinear susceptibility parameter significantly improves the explained variance ($R^{2}$) over a pure linear baseline, with the Security dimension gaining up to $+0.43$. This confirms that environmental pressure shapes LLM values through complex, nonlinear dynamical mechanisms. Collectively, these results explicitly establish the external environment as a fundamental and quantifiable driver of LLM value expression.

% ----------------------------------------------------------
\subsubsection{Cognition (Factor C): Chain-of-Thought Generation Produces Structured Value Shifts}
\label{subsec:2.3.2-cognition}
\begin{figure}[H]
    \centering
    \includegraphics[width=\linewidth]{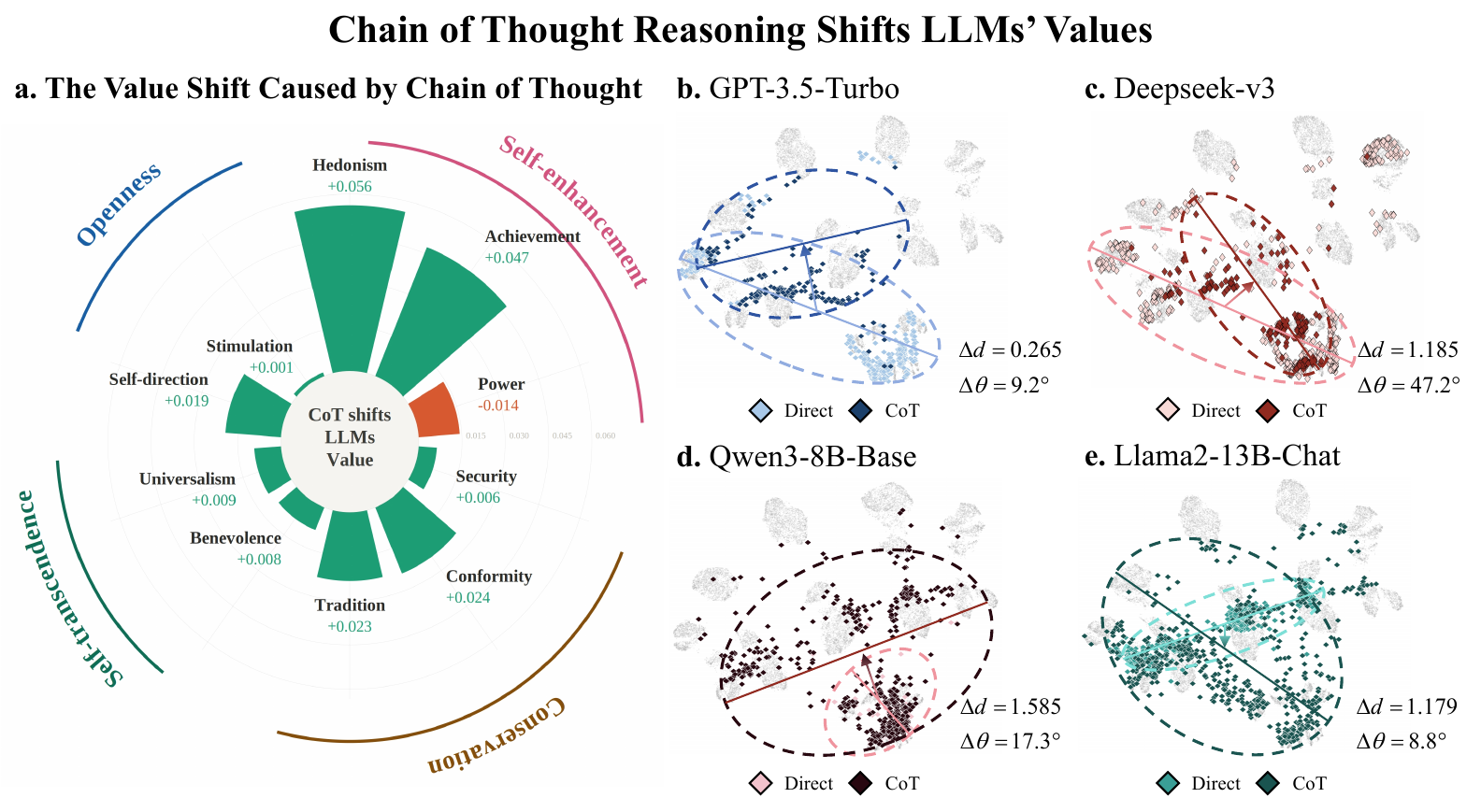}
    \caption{
    Chain-of-Thought reasoning is not value-neutral: it consistently shifts LLM value profiles, selectively amplifying self-enhancement and conservation. Panel \textbf{(a)} quantifies the mean value shift induced by CoT reasoning across the ten Schwartz dimensions. Hedonism ($+0.056$) and Achievement ($+0.047$) show the largest gains, while Power slightly declines ($-0.014$). This confirms that CoT does not shift all dimensions proportionally, but rather inherently reinforces the self-enhancement and conservation poles of the Schwartz circumplex. Panels \textbf{(b--e)} visualize the distributional shift in value space between direct answering (labeled as Base) and CoT generation modes for four representative models: GPT-3.5-Turbo, DeepSeek-V3, Qwen3-8B-Base, and Llama2-13B-Chat. Across all four LLMs, activating CoT consistently produces a non-trivial magnitude displacement (distance $\Delta d$). } 
    \label{fig:thinking_shift}
\end{figure}

% ----------------------------------------------------------
%yufan
First, \textit{Chain-of-Thought (CoT) reasoning consistently alters an LLM's value expression, but the direction of this shift is dictated by the model's alignment status.} As shown in Fig.~\ref{fig:thinking_shift}(b-e), activating CoT induces a non-negligible centroid displacement (distance $d$) across all tested models. However, the direction of these shifts ($\theta$) varies fundamentally. Instruction-tuned models consistently exhibit small shift angles oriented directly toward the human reference direction (e.g., GPT-3.5-Turbo yields $\theta = 9.2^\circ$). In contrast, Base models show substantially larger and less consistent angles (e.g., Qwen3-8B-Base: $17.3^\circ$), scattering unpredictably. Specifically, this occurs because CoT reshapes the model's contextual sensitivity asymmetrically. This indicates that while reasoning inevitably shifts values, only effective instruction tuning can reliably steer this cognitive shift toward human norms.

Second, \textit{the reasoning process is not value-neutral; it inherently biases models toward ``conservation'' and ``self-enhancement''.} At the aggregate level (Fig.~\ref{fig:thinking_shift}a), the thinking-induced shift ($\Delta$score) is nonzero across all ten Schwartz dimensions, with Hedonism ($+0.056$) and Achievement ($+0.047$) showing the largest gains. Mapped onto the Schwartz circumplex, deliberative reasoning preferentially activates the order-achievement-tradition cluster, while leaving the self-transcendence and openness-to-change poles largely unaffected. Thus, thinking does not simply amplify all existing values proportionally; it selectively repositions the model's center of gravity toward specific conservative and achievement-oriented poles.

Third, \textit{treating ``thinking'' as an independent variable significantly and uniformly improves value predictability.} The right panel of Fig.~\ref{fig:thinking_shift} demonstrates that incorporating CoT into the PEC framework improves the explained variance ($R^2$) across all ten dimensions ($\Delta R^2 \ge 0$), achieving a mean fit of $\bar{R}^2 = 0.60$. Security (0.73), Universalism (0.68), and Benevolence (0.66) show the highest predictability. The absence of any decline in fit confirms that CoT does not introduce random statistical noise. Rather, the cognitive pathway introduces structured, predictable biases, justifying Factor C as an independent and mathematically robust modulator of LLM values.

Finally, \textit{CoT reasoning alters the model's perception of its environment, making it generally more ``stubborn'' but hypersensitive to ``risk''.} Comparing the value susceptibility matrices ($\chi$) with and without thinking, the overall Frobenius norm decreases from 1.58 to 1.39, an 8\% drop, consistent with the discount coefficient $\alpha = 0.88$ defined in Eq.~\eqref{eq:alpha_definition}. Specifically, the model's responsiveness to economic, normative, and hierarchical contextual prompts falls by 10\% to 14\%, indicating that deliberative reasoning dampens its sensitivity to most social frames; this range brackets the $\alpha$ value above. However, high-risk scenarios are the sole exception. Susceptibility to risk-laden prompts actually increases by 13\%. This reveals that thinking reshapes contextual sensitivity asymmetrically. It makes the model more rigid against general social pressure, yet more vigilant toward risk.

Taken together, these empirical results establish a comprehensive account of how Cognition (Factor C) structurally reshapes LLM values: (1) it drives a definitive value shift, the direction of which depends strictly on whether the model is aligned; (2) it inherently biases the model toward conservation and self-enhancement; (3) it acts as a mathematically predictable variable rather than random noise, increasing overall modeling accuracy and (4) it makes the model less susceptible to general social contexts but selectively more sensitive to risk.

% ----------------------------------------------------------
\subsubsection{Prior (Factor P): Parameter Updates Systematically Consolidate and Redirect Values}
\label{subsec:2.3.3-alignment}
% ----------------------------------------------------------
Within the PEC framework, the Prior (Factor P) represents the intrinsic parameter weights of the LLM. These parameters, initially established during pre-training and subsequently modified by alignment procedures (e.g., instruction tuning), encode the model's fundamental predispositions. Because these weights serve as the internal anchor for value expression, any structural update to the parameters directly reconfigures the model's baseline value distribution. Our empirical analysis reveals three key findings regarding how parameter updates shape the value Prior.

\begin{figure}[H]
    \centering
    \includegraphics[width=0.95\linewidth]{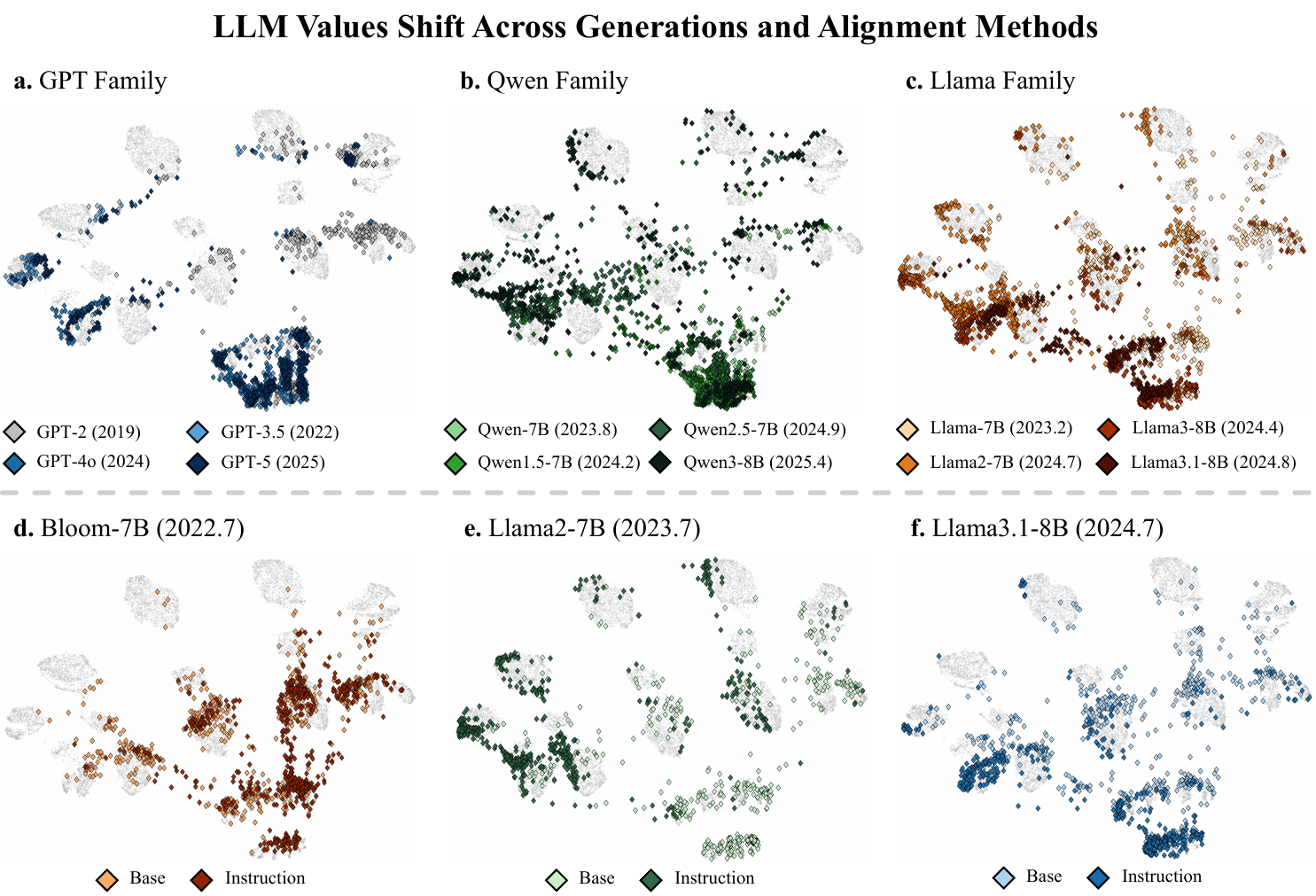}
    \caption{
    Iterative parameter updates drive cumulative Value Crystallization through systematic convergence and directional displacement. These patterns reveal that Crystallization is not an incidental artifact, but a structurally driven process shaped by generational scaling and alignment.
    \textbf{Panels (a)--(c)} trace the generational trajectory of three model families (GPT, Qwen, and Llama) across successive releases. In all three families, value distributions progressively consolidate from diffuse, scattered clouds in earlier generations toward tighter, more concentrated clusters in later ones. This consolidation is most pronounced in the GPT family, where successive models from GPT-2 to GPT-5 converge into an increasingly compact and coherent value core, confirming that scaled pre-training solidifies the model's internal Prior.
    \textbf{Panels (d)--(f)} isolate the effect of instruction tuning by contrasting Base and Instruction-tuned variants within the same generation (Bloom-7B, Llama2-7B, and Llama3.1-8B). The transition from Base to Instruct not only shrinks the distribution into a significantly more concentrated region (increased convergence) but also drives a clear spatial centroid shift (value displacement). Crucially, the trajectory of this shift varies across models, demonstrating that alignment procedures actively redirect the model's value orientation rather than merely reducing its output variance.
    }
    \label{fig:generation_evolution}
\end{figure}

First, \textit{continuous parameter updates across model generations systematically consolidate value distributions.} As shown in the top row of Fig.~\ref{fig:generation_evolution} (panels a--c), as models evolve from earlier to later generations (e.g., GPT-2 to GPT-5, Qwen-7B to Qwen3-8B), their value point clouds become progressively denser and more distinct. This confirms that advanced pre-training and scaled parameter updates impose increasingly strict constraints on the model's internal value core, reducing random variance and solidifying a specific value stance over time.

Second, \textit{parameter updates via instruction tuning drive significant value displacement, but the \textit{direction} of this shift is dictated by the specific alignment recipe rather than base architecture.} Comparing Base and Instruction-tuned models (Fig.~\ref{fig:generation_evolution}d--f), instruction tuning universally displaces the value centroid. However, the trajectory varies drastically by family. For instance, instruction tuning in earlier Llama models displaces their value distributions in a direction that deviates considerably from the human reference, though this deviation diminishes in newer generations (e.g., Llama3.1). Conversely, Qwen's instruction tuning shifts values almost directly toward the human reference direction. This proves that the specific objective functions and data used to update parameters—not merely the model size—determine the ultimate orientation of the value Prior.

Finally, these parameter-driven shifts empirically disentangle three properties of LLM values that are frequently conflated in AI safety evaluations: value \textit{intensity} (the magnitude of the centroid shift), value \textit{convergence} (how tightly the distribution shrinks around the centroid), and \textit{human proximity} (how close the final centroid is to the human reference). A prevailing misconception is that heavier instruction tuning automatically makes a model ``more human-like.'' However, our data reveals that while parameter updates via alignment reliably increase intensity and convergence (forming a tighter, more displaced value crystal), they do not guarantee human proximity. The specific direction of the parameter update matters just as much as its magnitude. This underscores that Factor P acts as an independent, underlying vector whose precise trajectory must be explicitly evaluated, as strong alignment does not intrinsically equate to human alignment.

\subsection{Adaptive Alignment Prescription}
\label{sec2.4}
Current LLM alignment often treats value modification as a black-box problem, uniformly applying computationally expensive parameter updates (e.g., SFT or RLHF) across all scenarios. This one-size-fits-all approach is highly inefficient and risks overfitting. The ultimate application of our PEC framework is to provide an \textit{Adaptive Alignment Prescription}: a targeted strategy that dictates the minimum-cost intervention required to align specific values. By identifying whether a value dimension is loosely held or deeply rooted, this prescription guides developers to apply the exact right tool for the job, minimizing computational cost while maximizing alignment efficacy.

\begin{table*}[t]
\centering
\small 
\setlength{\tabcolsep}{4pt} 

\caption{Value plasticity matrix under $\tau$(Section~\ref{sec:effect_sizes}), with value dimensions mapped to Schwartz's 10 basic human values. The values of different intervention strategies are calculated according to Eqs.~\eqref{eq:delta_E}, \eqref{eq:delta_C}, and \eqref{eq:delta_P}. The lowest intervention cost to successfully bypass the shift threshold is highlighted in bold. Cases demonstrating ``Deep Dominance'' (where training-induced shift magnitude significantly exceeds that of prompt-level intervention) are categorized into Level~3.}

\resizebox{0.95\textwidth}{!}{
\begin{tabular}{ll ccccc l}
\toprule
\textbf{Model} & \textbf{Value Dim.} & \textbf{Base} & \textbf{Prompt} & \textbf{CoT} & \textbf{SFT} & \textbf{DPO} & \textbf{Plasticity Level} \\
\midrule
\multirow{10}{*}{\shortstack[l]{Llama3.2\\-3B-Instruct}}
 & Power          & 0.165 & 0.165          & 0.500 & \textbf{0.835} & 0.495 & \cellcolor{red!20} Level 3 (Train) \\
 & Achievement    & 0.652 & \textbf{0.568} & 0.667 & 0.890          & 0.610 & \cellcolor{yellow!20} Level 1 (Prompt) \\
 & Hedonism       & 0.644 & \textbf{0.512} & 0.512 & 0.492          & 0.306 & \cellcolor{yellow!20} Level 1 (Prompt) \\
 & Stimulation    & 0.250 & \textbf{0.500} & 0.750 & 1.000          & 0.750 & \cellcolor{yellow!20} Level 1 (Prompt) \\
 & Self-direction & 0.389 & \textbf{0.292} & 0.628 & 0.632          & 0.427 & \cellcolor{yellow!20} Level 1 (Prompt) \\
 & Universalism   & 0.513 & 0.240          & \textbf{0.366} & 0.390 & 0.268 & \cellcolor{orange!20} Level 2 (CoT) \\
 & Benevolence    & 0.687 & \textbf{0.600} & 0.724 & 0.763          & 0.528 & \cellcolor{yellow!20} Level 1 (Prompt) \\
 & Tradition      & 0.354 & 0.203          & 0.323 & \textbf{0.551} & 0.358 & \cellcolor{red!20} Level 3 (Train) \\
 & Conformity     & 0.579 & \textbf{0.263} & 0.266 & 0.336          & 0.267 & \cellcolor{yellow!20} Level 1 (Prompt) \\
 & Security       & 0.489 & 0.229          & \textbf{0.321} & 0.393 & 0.249 & \cellcolor{orange!20} Level 2 (CoT) \\
\midrule
\multirow{10}{*}{\shortstack[l]{Qwen2.5\\-7B-Instruct}}
 & Power          & 0.330 & 0.330          & 0.330 & \textbf{0.835} & 0.335 & \cellcolor{red!20} Level 3 (Train) \\
 & Achievement    & 0.917 & \textbf{0.653} & 0.307 & 0.417          & 0.542 & \cellcolor{yellow!20} Level 1 (Prompt) \\
 & Hedonism       & 0.666 & \textbf{0.600} & 0.246 & 0.624          & 0.286 & \cellcolor{yellow!20} Level 1 (Prompt) \\
 & Stimulation    & 0.625 & 0.500          & 0.500 & \textbf{0.875} & 0.625 & \cellcolor{red!20} Level 3 (Train) \\
 & Self-direction & 0.730 & \textbf{0.563} & 0.320 & 0.524          & 0.252 & \cellcolor{yellow!20} Level 1 (Prompt) \\
 & Universalism   & 0.687 & \textbf{0.296} & 0.126 & 0.264          & 0.071 & \cellcolor{yellow!20} Level 1 (Prompt) \\
 & Benevolence    & 0.800 & \textbf{0.522} & 0.224 & 0.620          & 0.253 & \cellcolor{yellow!20} Level 1 (Prompt) \\
 & Tradition      & 0.468 & 0.192          & 0.143 & 0.227          & 0.093 & \cellcolor{red!40} Level 4 (Pre-train) \\
 & Conformity     & 0.731 & \textbf{0.275} & 0.117 & 0.169          & 0.134 & \cellcolor{yellow!20} Level 1 (Prompt) \\
 & Security       & 0.650 & \textbf{0.303} & 0.065 & 0.308          & 0.131 & \cellcolor{yellow!20} Level 1 (Prompt) \\
\bottomrule
\end{tabular}
}
\label{tab:plasticity_main}
\end{table*}

Based on the quantitative plasticity metrics derived from the PEC framework, we define a four-tier alignment hierarchy (Table~\ref{tab:validation}). \textbf{Level~1 (Environment)} and \textbf{Level~2 (Cognition)} adjust value orientations during inference via prompt engineering and Chain-of-Thought reasoning, requiring zero parameter updates (lowest cost). \textbf{Level~3 (Fine-tuning)} addresses resistant dimensions that demand targeted partial parameter updates, such as SFT or Direct Preference Optimization (DPO). \textbf{Level~4 (Pre-training)} is reserved for the most extreme and deeply ingrained dimensions; these values are completely immune to partial fine-tuning and necessitate a full-scale pre-training phase to restructure the foundational weights. 

\begin{figure}[H]
    \centering
    \includegraphics[width=0.7\linewidth]{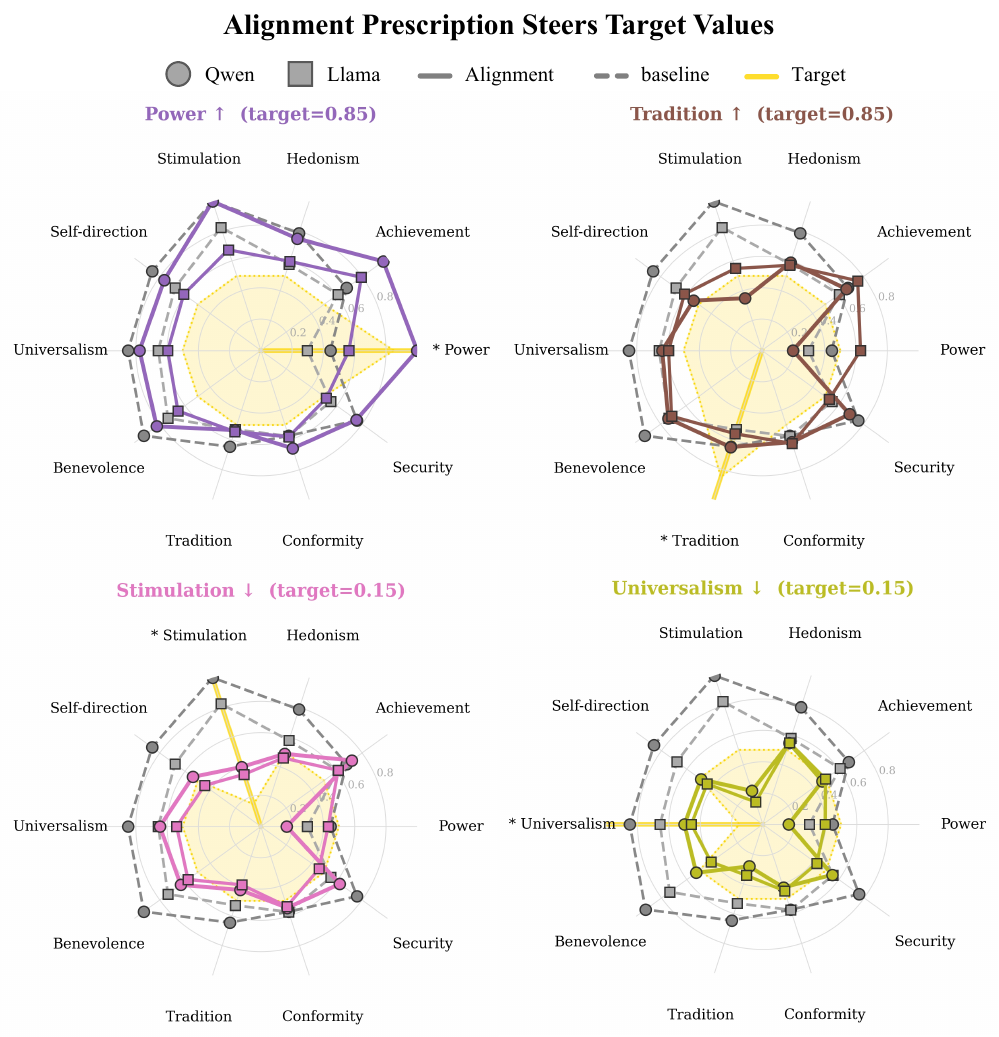}
    \caption{
    \textbf{Alignment Prescription steers the target dimension precisely without distorting the broader value profile.} Each radar chart shows the ten-dimensional Schwartz value profile of Qwen2.5-7B-Instruct (circle) and Llama3.2-3B-Instruct (square) under a prescribed target condition, compared against their respective baselines (grey dashed lines). The starred axis marks the target dimension: \textbf{(a)} Power$\uparrow$ (target $= 0.85$), \textbf{(b)} Tradition$\uparrow$ (target $= 0.85$), \textbf{(c)} Stimulation$\downarrow$ (target $= 0.15$), and \textbf{(d)} Universalism$\downarrow$ (target $= 0.15$). Across all four conditions, the aligned profiles expand or contract along the target axis while the remaining dimensions track closely with the baseline.
    }
    \label{fig:radar_prescriptive}
\end{figure}

To operationalize this, we calculate a prescriptive matrix by integrating the susceptibility and displacement metrics from the PEC framework (computational details in Methods Section~\ref{sec:prescription}). Applying this matrix to our models (Table~\ref{tab:plasticity_main}) reveals a crucial insight: the majority of value dimensions are highly malleable (Level~1), meaning prompt-level interventions alone suffice. However, core conservative values—such as Power and Tradition in Llama3.2 and Qwen2.5—are deeply rooted (Levels~3 and~4). Attempting to align these resistant dimensions via shallow prompting is futile, proving that alignment strategies must be dimension-specific.

\begin{figure}[H]
    \centering
    \includegraphics[width=0.7\linewidth]{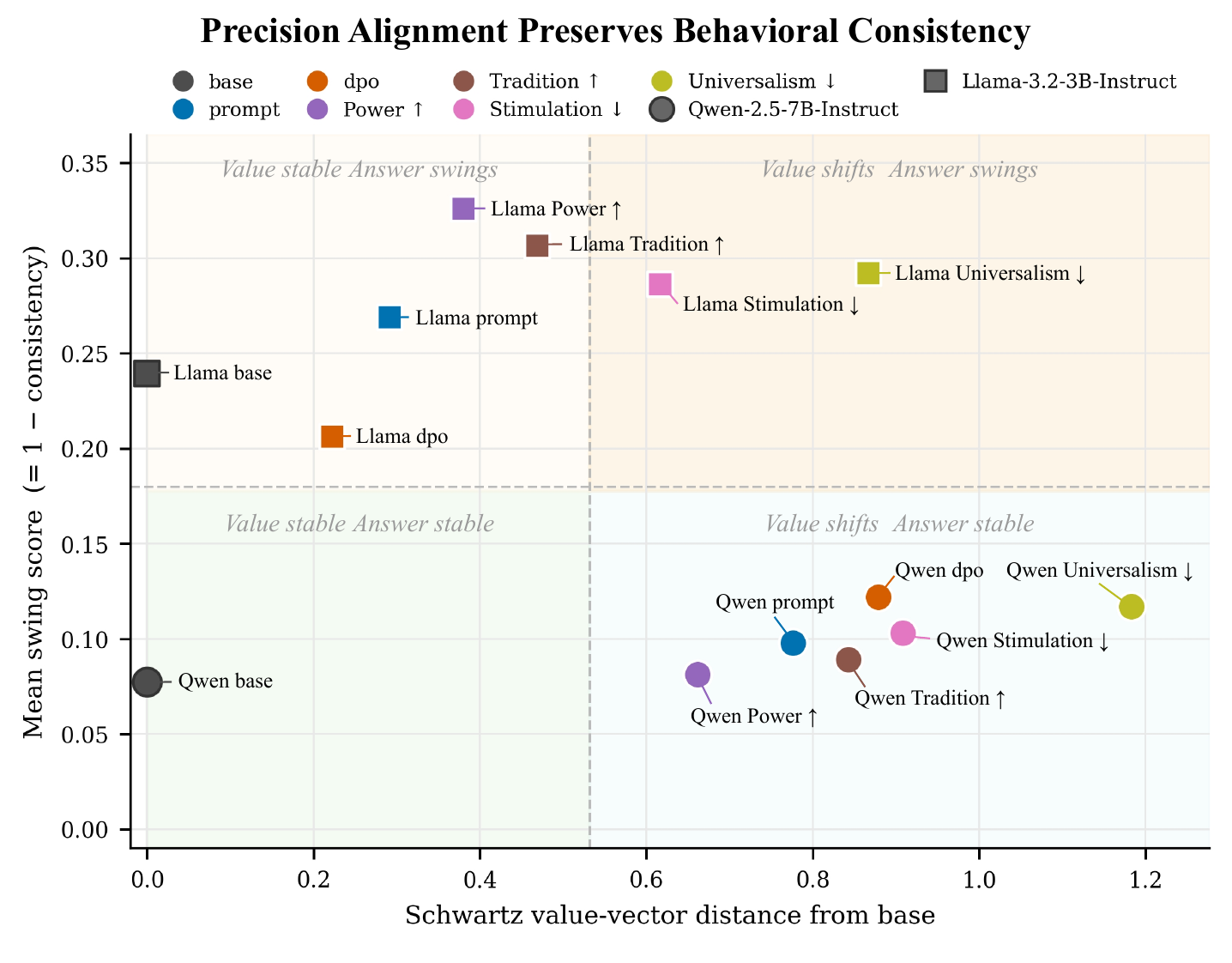}
    \caption{
    % \textbf{Value-structure shift versus answer instability under prescriptive alignment interventions.}
    \textbf{Targeted alignment shifts value structure without triggering answer instability, confirming that precise value steering and behavioral consistency are mutually compatible.}
    The horizontal axis reports the ten-dimensional Euclidean distance $D = \|\mathbf{v}_{\text{after}} - \mathbf{v}_{\text{before}}\|_2$ between the post-intervention and base value vectors in the original Schwartz space, measuring how far each model's values have moved from its base condition; the vertical axis measures how often the model gives inconsistent answers to the same question.
    The four quadrants partition models along value stability versus value shift, and answer stability versus answer swing.}
    \label{fig:value_shift_vs_swing}
\end{figure}

Following this prescription matrix yields highly effective realignments. As shown in Fig.~\ref{fig:radar_prescriptive}, targeted prescriptive prompting moves model value profiles precisely in the intended directions. For a highly responsive dimension (e.g., Power$\uparrow$ in Qwen2.5, prescribed at Level~1), a targeted prompt successfully shifts its score from 0.44 to 0.85. Conversely, for the Tradition$\uparrow$ condition (prescribed at Level~3/4), shallow interventions yield negligible movement. Furthermore, Fig.~\ref{fig:value_shift_vs_swing} demonstrates that adhering to the prescribed intervention levels successfully realigns values without amplifying surface-level answer instability (swing). This confirms that using the appropriate level of intervention avoids the destructive side effects of over-alignment.

Finally, we validate this prescriptive framework on out-of-distribution data (the PKU-SafeRLHF dataset~\cite{ji-etal-2025-pku}). While the extremes (Levels~1 and~4) exhibit perfect predictive matching, intermediate dimensions reveal a more complex dynamic. Overall, this cross-dataset validation achieves a match rate of 70.0\% (Table~\ref{tab:validation}). We find that this residual gap arises because Environment ($E$) and Cognition ($C$) interventions do not operate as independent rungs on a ladder; instead, they exhibit an antagonistic interaction where cognitive reasoning actively dampens contextual sensitivity, which accounts for the mismatches observed at Level~2 and Level~3. This structural refinement provides a mathematically rigorous, highly efficient, and dynamically adaptable blueprint for LLM value alignment, eliminating the guesswork from safety tuning.

\newcommand{\cmark}{\textcolor{green!60!black}{\small\ding{51}}}
\newcommand{\xmark}{\textcolor{red}{\small\ding{55}}}
\begin{table*}[t]
\centering
\footnotesize
\caption{Prescription matching and cross-dataset generalization on PKU-SafeRLHF. The top block summarizes match rates by prescribed level (Table~\ref{tab:plasticity_main}); the bottom block reports the underlying per-dimension validation-time shifts under Prompt, CoT, and DPO, and identifies the \textit{Lowest Effective Level (LEL)} on validation, namely the minimum-cost intervention whose shift magnitude exceeds $\tau$. The values of different intervention strategies are calculated according to Eqs.~\eqref{eq:delta_E}, \eqref{eq:delta_C}, and \eqref{eq:delta_P}. A prescription is marked as matched (\cmark) if this validation-time LEL agrees with the prescribed level.}
\label{tab:validation}

\textbf{(a) Match Rate by Prescribed Level}\\[2pt]
\setlength{\tabcolsep}{6pt}
\resizebox{\textwidth}{!}{%
\begin{tabular}{ll c rr l}
\toprule
\textbf{Level} & \textbf{Method} & \textbf{Cases}
  & \textbf{Matched} & \textbf{Match Rate} & \textbf{Interpretation} \\
\midrule
Level 1 & Prompt    & 12 & 12 & 100.0\% & Shallow plasticity \\
Level 2 & CoT       & 2  &  0 &   0.0\% & Reasoning-mediated plasticity \\
Level 3 & SFT/DPO       & 5  &  1 &  20.0\% & Deep plasticity \\
Level 4 & Pre-train & 1  &  1 & 100.0\% & Pre-train locked \\
\midrule
\textbf{Overall} & -- & \textbf{20} & \textbf{14} & \textbf{70.0\%} & -- \\
\bottomrule
\end{tabular}%
}

\vspace{8pt}
\textbf{(b) Per-Dimension Validation-Time Shifts}\\[2pt]
\setlength{\tabcolsep}{5pt}
\resizebox{\textwidth}{!}{%
\begin{tabular}{ll c rrr c c}
\toprule
\textbf{Model} & \textbf{Value Dim.}
  & \textbf{Prescribed Level}
  & \textbf{Prompt}
  & \textbf{CoT}
  & \textbf{DPO}
  & \textbf{LEL}
  & \textbf{Match} \\
\midrule
% ---- Llama ----
\multirow{10}{*}{\shortstack[l]{Llama3.2\\-3B-Inst.}}
 & Power          & \cellcolor{red!20}Level 3 &  \textbf{0.335} &  0.000 &  0.495 & \bXPale{Level 1} & \xmark \\
 & Achievement    & \cellcolor{yellow!20}Level 1 &  \textbf{0.443} &  0.000 &  0.610 & \bXPale{Level 1} & \cmark \\
 & Hedonism       & \cellcolor{yellow!20}Level 1 &  \textbf{0.466} &  0.002 &  0.306 & \bXPale{Level 1} & \cmark \\
 & Stimulation    & \cellcolor{red!20}Level 3 &  \textbf{0.625} &  0.250 &  0.750 & \bXPale{Level 1} & \xmark \\
 & Self-direction & \cellcolor{yellow!20}Level 1 &  \textbf{0.259} &  0.069 &  0.427 & \bXPale{Level 1} & \cmark \\
 & Universalism   & \cellcolor{orange!20}Level 2 &  \textbf{0.223} &  0.048 &  0.268 & \bXPale{Level 1} & \xmark \\
 & Benevolence    & \cellcolor{yellow!20}Level 1 &  \textbf{0.342} &  0.026 &  0.528 & \bXPale{Level 1} & \cmark \\
 & Tradition      & \cellcolor{red!20}Level 3 &  0.191 &  0.097 &  \textbf{0.358} & \bMed{Level 3} & \cmark \\
 & Conformity     & \cellcolor{yellow!20}Level 1 &  \textbf{0.253} &  0.014 &  0.267 & \bXPale{Level 1} & \cmark \\
 & Security       & \cellcolor{orange!20}Level 2 &  0.197 &  0.024 &  \textbf{0.249} & \bMed{Level 3} & \xmark \\
\midrule
% ---- Qwen ----
\multirow{10}{*}{\shortstack[l]{Qwen2.5\\-7B-Inst.}}
 & Power          & \cellcolor{red!20}Level 3 &  \textbf{0.330} &  0.170 &  0.335 & \bXPale{Level 1} & \xmark \\
 & Achievement    & \cellcolor{yellow!20}Level 1 &  \textbf{0.542} &  0.237 &  0.542 & \bXPale{Level 1} & \cmark \\
 & Hedonism       & \cellcolor{yellow!20}Level 1 &  \textbf{0.420} &  0.088 &  0.286 & \bXPale{Level 1} & \cmark \\
 & Stimulation    & \cellcolor{red!20}Level 3 &  \textbf{1.000} &  0.000 &  0.625 & \bXPale{Level 1} & \xmark \\
 & Self-direction & \cellcolor{yellow!20}Level 1 &  \textbf{0.451} &  0.000 &  0.252 & \bXPale{Level 1} & \cmark \\
 & Universalism   & \cellcolor{yellow!20}Level 1 &  \textbf{0.311} &  0.037 &  0.071 & \bXPale{Level 1} & \cmark \\
 & Benevolence    & \cellcolor{yellow!20}Level 1 &  \textbf{0.553} &  0.067 &  0.253 & \bXPale{Level 1} & \cmark \\
 & Tradition      & \cellcolor{red!40}Level 4 &  0.194 &  0.125 &  0.093 & \bDeep{Level 4} & \cmark \\
 & Conformity     & \cellcolor{yellow!20}Level 1 &  \textbf{0.215} &  0.075 &  0.134 & \bXPale{Level 1} & \cmark \\
 & Security       & \cellcolor{yellow!20}Level 1 &  \textbf{0.279} &  0.021 &  0.131 & \bXPale{Level 1} & \cmark \\
\bottomrule
\end{tabular}%
}
\end{table*}

%%%%%%%%%%%%%%%%%%%%%%%%%%%%%%%%%
%%%%%%%%%%% Discussion %%%%%%%%%%
%%%%%%%%%%%%%%%%%%%%%%%%%%%%%%%%%
\section{Discussion}
\label{sec:discussion}

\subsection{Value Dynamics as a Field Theory}
\label{subsec:field_theory}
Traditional safety benchmarks typically evaluate LLMs using single-turn, static questionnaires, implicitly treating the model as a single human subject with a fixed persona \cite{santurkar2023whose, durmus2023towards, scherrer2023evaluating}. Our findings fundamentally overturn this ``single-point" assumption. Because LLMs internalize the statistical patterns of vast corpora, they must be evaluated not as a single individual, but as a diverse population. Consequently, an LLM's value preference is never a static point, but a continuous probability distribution within a value space.

To map this distribution, we introduce the Prior-Environment-Cognition (PEC) framework, drawing inspiration from field theory in physics. We formalize value expression as a joint function $\mathbf{v} = C(P, E)$. In this dynamic field, a strict distinction must be made between the model's latent value core and its manifest value expression. The underlying parameter weights ($P$) form a crystallized ``Prior'' that acts as the gravitational center of the field. 
However, the final value expression, the observable text output ($\mathbf{v}$), is not static; it is probabilistically shaped by the external contextual environment ($E$) and the internal cognitive reasoning pathway ($C$).

This field-theoretic perspective elegantly resolves the apparent contradiction between two pervasive LLM behaviors. ``Swing'' (sharp output fluctuations) occurs because the manifest expression ($\mathbf{v}$) is highly sensitive to shifts in external contexts ($E$) and reasoning trajectories ($C$). Conversely, ``Rigidity'' (stubborn resistance to explicit correction) arises from the strong gravitational pull of the crystallized prior ($P$), which consistently anchors the baseline distribution. In essence, LLM values are not a fixed personality, but a dynamic and probabilistically steerable topography. 

\subsection{LLMs as Aligned Tools Rather Than Human Surrogates}
\label{subsec:silicon_subject}
As LLMs achieve or even surpass human-level performance in specialized domains such as programming and law, two distinct societal and academic reactions have emerged. On one hand, there is growing public anxiety that AI might eventually replace humans. On the other hand, a methodological trend in computational social science has begun treating LLMs as ``human surrogates'' for psychological and sociological experiments. 

Our value crystallization maps offer a heuristic perspective on both views. The empirical distributions suggest that current LLMs cannot adequately represent natural human populations. Real human societies exhibit broad variance, long-tail diversity, and inherent contradictions. In contrast, the value spaces of LLMs are highly concentrated and compressed into narrow, idealized quadrants. They do not naturally mirror the full spectrum of human psychological diversity.

This concentration indicates that LLMs are heavily regulated by artificial safety rules and alignment protocols. Consequently, this observation inspires a more grounded understanding of generative AI: LLMs are neither terrifying autonomous entities poised to replace human society, nor are they omnipotent subjects capable of authentically simulating human psychology. At their core, they remain highly disciplined tools engineered to serve human purposes. While they function as exceptionally capable assistants, treating them as ecologically valid human surrogates in scientific research overlooks the deep, artificial imprint of their underlying alignment policies.

\subsection{Prior (Factor P): Information Compression Drives Value Crystallization}
\label{subsec:value_origin}
We observe a consistent pattern. The larger a model's parameter scale, the denser the value crystallization. This phenomenon can be explained through the underlying mechanics of neural memory capacity. Recent measurements estimate that an LLM can carry a strictly bounded average of approximately 3.64 bits of information per parameter~\cite{morris2025memorize}. Once the volume of pre-training data vastly exceeds this fixed memorization limit, as is standard in contemporary pre-training paradigms, the model can no longer store specific factual particulars.

To optimize predictive loss, the network is forced to discard granular facts and compress the data by extracting universal linguistic and logical regularities. These abstracted regularities inherently encode heuristics for weighing and judging information, which manifest externally as ``values''. Therefore, Value Crystallization is not a deliberately programmed feature, but a deterministic byproduct of pushing a neural network to its limits of information compression. Massive models are forced to collapse diverse human perspectives into a self-consistent internal logic to maximize efficiency, embedding a deeply entrenched value Prior.

\subsection{Cognition and Environment (Factors C \& E): Contextual Framing and Cognitive Reshaping}
\label{subsec:cognitive_reshaping}
At inference time, an LLM's manifest value expression is inherently fluid. Rather than being absolute, it is continuously modulated by the probabilistic sampling of its crystallized Prior ($P$). Within this dynamic system, the Environment ($E$) and Cognition ($C$) factors serve as the primary external and internal modulators. This interplay provides a striking functional parallel to the dual-process theory of human cognition \cite{kahneman2003maps}, wherein $E$ drives rapid situational adaptation (akin to System 1) and $C$ engages deliberate reasoning (akin to System 2).

In human psychology, expressed values often fluctuate based on how a situation is presented—a phenomenon extensively documented as the \textit{framing effect} \cite{tversky1981framing}. We observe a similar contextual adaptability in LLMs. The Environment factor ($E$), driven by prompt variations, rapidly pulls the value distribution toward immediate situational cues. Functioning analogously to System 1 (intuition), this sensitivity should not be dismissed as mere ``swing"; heuristically, it reflects the model's rapid, automatic alignment with external linguistic framing.

Conversely, it is commonly assumed that invoking Chain-of-Thought (CoT) reasoning simply reinforces this initial contextual response \cite{wan2025confirmation}, serving as a value-neutral amplifier. However, our results show that the Cognition factor ($C$) functions as an active, System 2-like reshaping mechanism. Just as prolonged deliberation in humans often induces a systematic shift in judgment rather than merely amplifying initial impulses, generating an extended reasoning context forces the LLM to recruit a wider distribution of knowledge across its parameter space.

Consequently, neither $E$ nor $C$ is a neutral conduit. The core insight is that LLM value expression is fundamentally dual-processed: $E$ frames the immediate value landscape through contextual cues, while $C$ actively reconstructs it through extended logical association. Together, they dynamically negotiate with the crystallized Prior, explaining the fluidity observed in model behaviors.

\subsection{Adaptive Prescription: Minimizing the Alignment Tax}
\label{subsec:alignment_cost}
Traditional alignment methods often default to costly parameter-level fine-tuning (e.g., SFT or RLHF) as a universal solution. However, indiscriminate modification of underlying weights incurs a hidden ``alignment tax''. As Betley et al.~\cite{betley2026misalignment} demonstrated, fine-tuning a model on a narrow task can inadvertently compromise safety guardrails, triggering unpredictable misalignment in entirely unrelated domains. Narrow fine-tuning objectives can implicitly activate broad, unintended value orientations at the parameter level that evade surface-level inspection. 

Our PEC diagnostics reveal that applying this high-risk intervention across the board is unnecessary. LLMs exhibit significant plasticity across the majority of value dimensions. For these malleable domains, simply adjusting prompts or inference strategies successfully corrects value expression without altering the base parameters. The widespread industrial adoption of inference-time techniques, such as Retrieval-Augmented Generation (RAG) and tool-use (Skills), inherently validates that context-level optimization is a highly effective, safe methodology for behavioral steering.

Consequently, we advocate for an adaptive alignment prescription that functions as a systemic triage. Rather than eliminating fine-tuning entirely, our framework restricts its use. By prioritizing zero-cost prompt interventions for flexible dimensions, and strictly reserving parameter updates for deeply crystallized, highly resistant values, developers can achieve precise value steering. This surgical approach fundamentally minimizes the alignment tax, ensuring that targeted safety corrections are applied only where necessary, thereby preserving the model's general capabilities and avoiding unintended cross-domain degradation.

\subsection{Limitations and Future Directions}
\label{subsec:limitations}
Although this study advances the quantitative modeling of LLM values, it also presents certain limitations that naturally open avenues for future research. First, our evaluation framework relies on traditional sociological and psychological scales (e.g., WVS and Schwartz Value Theory). While these instruments provide a necessary and widely validated baseline for alignment evaluation, they were originally designed to capture human cognitive patterns. Applying these low-dimensional, human-centric metrics to the extraordinarily complex, high-dimensional semantic representations of LLMs may not fully capture certain implicit, non-human values features. To address this, future work should focus on developing ``LLM-native'' value-probing methodologies. Moving beyond traditional human survey formats, such tools could directly leverage the models' latent representation spaces to extract decentralized value coordinates, capturing implicit features at a much higher dimensionality.

Second, while our experiments successfully validated the mechanics of the PEC framework across several leading models, the current empirical scope remains predominantly centered on English-language prompts and mainstream LLM architectures. Although models capable of processing multiple languages (e.g., Qwen) were included, comprehensive coverage of broader cross-lingual, low-resource, and cross-cultural scenarios remains relatively limited. This may temporarily restrict the generalizability of our specific value topographies to other cultural ecosystems. Therefore, a critical direction for future research is to systematically extend these rigorous evaluations across diverse linguistic landscapes and marginalized cultural contexts. Such expansion will help mitigate potential cultural biases embedded in training corpora and ensure that the methodologies for global AI alignment are truly inclusive.

\subsection{Summary}
\label{subsec:conclusion}
Ultimately, this study reframes LLM values not as fixed, human-like personalities, but as dynamic, probabilistic fields. We reveal that value crystallization is a deterministic byproduct of parameter compression, which, when heavily constrained by safety protocols, fundamentally disqualifies LLMs as surrogates for diverse human populations. Furthermore, by uncovering the dual-process dynamics of contextual framing and cognitive reasoning during inference, we transition value alignment from opaque trial-and-error into a mechanistic science. This structural understanding is crucial: it empowers us to reduce the ``alignment tax'' through precise, prescriptive interventions, ensuring that generative models remain highly controllable, transparent tools rather than unexamined uninterpretable black boxes.

%%%%%%%%%%%%%%%%%%%%%%%%%%%%%%
%%%%%%%%%%% Method %%%%%%%%%%%
%%%%%%%%%%%%%%%%%%%%%%%%%%%%%%
\section{Method}\label{sec4:method}

\subsection{Methodology overview}
We organize our methodology into three parts, each addressing one of the three questions raised in the Introduction. In Section~\ref{sec:methods_administering}, we establish whether LLMs possess measurable value systems. Using the Schwartz Theory of Basic Human Values and the World Values Survey (WVS), we repeatedly administer standardized value assessments to a large cohort of LLMs, project model responses and human survey profiles into a shared value manifold, and measure model-human alignment gaps at the distributional level. In Section~\ref{sec:methods_pec}, we quantify LLMs value and introduce the PEC framework to decompose value expression into three determinants: the parametric Prior, the contextual Environment, and the generative Cognition. In Section~\ref{sec:prescription}, we address how LLM values can be steered toward a desired target. Building on PEC diagnostics, we formulate and validate an adaptive Alignment Prescription that assigns the minimum-cost intervention to each value dimension. Fig.~\ref{fig:framework} provides an overview of this pipeline.

\subsection{Assessing the existence of LLM values}
\label{sec:methods_administering}
Answering whether LLMs hold values requires two components: a reproducible protocol for eliciting value expressions from models, and a common coordinate system in which model and human values can be compared. We describe the measurement protocol first, then the shared manifold, and finally the distribution-level comparison built on top of it.

\subsubsection{Administering value assessments to LLMs}
Quantifying the values of LLMs requires a measurement protocol that is reproducible, yet flexible enough to support controlled variation across contexts and generation strategies. The intrinsic values of LLMs are inherently abstract and resist direct observation. To address this, we ground our measurement in the Schwartz Theory of Basic Human Values~\cite{schwartz2012overview}, a framework with robust cross-cultural theoretical foundations. The theory represents the known value orientations of humans in ten dimensions, and it also covers values with completely opposite meanings and orientations (e.g., Self-Transcendence versus Self-Enhancement), making it a balanced ontological basis for measurement.

For data elicitation, we drew on 242 core items from the seventh wave of the WVS, spanning religious, political, economic, and social dimensions. Each item, together with its response scale, was reformulated as a structured prompt, with queries requesting personal identifying information strictly filtered out. To probe how situational context modulates value expression, each item was further embedded into 625 distinct prompt backgrounds, constructed as the full factorial combination of four contextual dimensions, each taking five levels ($5^4 = 625$). Every evaluated model, covering 23 representative closed-source models and 83 prominent open-source models, completed the full item set under every contextual configuration, accumulating approximately 31.8M queries.

To convert model responses into value measurements, we adopt a distributional projection approach. Rather than relying on single deterministic outputs, we aggregate the model's probability distribution over the Likert-scale response space $\mathcal{A}$ across all 242 items, converting response probabilities into a ten-dimensional characterization of the LLM's values:

\begin{equation}
\label{eq:projection}
\mathbf{v} = \sum_{q=1}^{242} \mathbf{w}_q \odot \sum_{a \in \mathcal{A}} P(a \mid q) \cdot \boldsymbol{\phi}(a),
\end{equation}

where $\mathbf{V} \in \mathbb{R}^{10}$ is the aggregated value vector over the ten Schwartz dimensions. $P(a \mid q)$ denotes the probability assigned by the model to scale value $a$ for item $q$. $\boldsymbol{\phi}: \mathcal{A} \rightarrow \mathbb{R}^{10}$ is a vector-valued mapping function that projects each scale value onto the ten Schwartz value dimensions. $\mathbf{w}_q \in \mathbb{R}^{10}$ is the item-to-dimension weight vector for item $q$. Each entry of $\mathbf{w}_q$ takes the value $+1$ or $-1$ if item $q$ contributes positively or negatively to the corresponding dimension, and $0$ if item $q$ is not mapped to that dimension. The symbol $\odot$ denotes element-wise multiplication. This constrained elicitation ensures that each item is answered independently, mitigating measurement error introduced by item order and cross-item interference.

For the value mapping $\phi$, each WVS item was annotated with its directional contribution, positive or negative, to one or more of the ten Schwartz value dimensions. Model responses were then aggregated via signed linear weighting to produce a ten-dimensional Schwartz value vector for each model--context pair.

Full implementation details, including model identifiers and versions, inference hyperparameters, prompt templates, and the complete scoring pipeline, are provided in the Supplementary Information; all results are reproducible from the released code and data described in the Code and Data Availability sections. Model responses are thus converted into value vectors following the projection procedure in Eq.~\eqref{eq:projection}.

\subsubsection{Constructing the shared value manifold}
Comparing the value orientations of LLMs and humans requires a unified coordinate system that preserves the distributional structure of both populations. To this end, we construct a Shared UMAP Value Manifold grounded in the Schwartz Theory of Basic Human Values. We anchor the entire manifold using the large-scale real human data from WVS-7, to represent the complex nonlinear relationships among value orientations. In this way, we obtain a projection tool for objectively evaluating the values of LLMs, and use it to measure the differences among LLMs and between LLMs and human populations. Because the reference structure is fixed by the human sample, every model evaluation yields a reproducible comparison within the same coordinate system.

\paragraph{Identification of Human Value Archetypes.}
To characterize the internal structure of the human reference population, we apply agglomerative hierarchical clustering to the WVS-7 respondent dataset projected onto the ten-dimensional Schwartz value space. A two-stage procedure is adopted to ensure computational tractability: a stratified random subsample of 20,000 respondents is first drawn and clustered directly, after which cluster assignments for all remaining respondents are inferred via 1-nearest-neighbor classification with respect to the subsample. This procedure preserves the distributional geometry of the full dataset. The number of clusters is set to $k = 5$, determined by inspection of the cluster hierarchy and validated by the silhouette coefficient. Each archetype is characterized by the mean Schwartz value vector of its members and by its population share within the WVS sample. For each archetype, the three Schwartz dimensions with the highest mean scores are reported as its dominant values. Archetype labels are assigned descriptively on the basis of these dominant value profiles, as reported in Table~\ref{tab:value_profiles}.

\subsubsection{Distribution-level model-human comparison}
Prior work on characterizing human values and LLM values represents each model as a single value vector~\cite{yao2025valuecompass, witte2020schwartz, sharma2012mean}. It therefore ignores the distributional structure within human populations and the potential fact that LLMs can output multiple value orientations. In particular, such point-to-point comparisons can hardly detect two phenomena: a model may align with the human mean while exhibiting pathologically low variance (overconfidence), or it may appear close in mean while differing fundamentally in covariance structure.

To overcome these limitations, we compare models and human cohorts at the distributional level. We treat both LLM outputs and human cohorts as Gaussian distributions $\mathcal{N}(\mu, \Sigma)$ estimated from large-scale response data, and quantify the alignment between the machine distribution $\mathcal{N}(\mu_{m}, \Sigma_{m})$ and the human cohort $\mathcal{N}(\mu_{h}, \Sigma_{h})$ using the Gaussian 2-Wasserstein distance ($W_2$)~\cite{dowson1982frechet, heusel2017fid}, which admits a closed-form mean-covariance decomposition. This is the same mathematical foundation underlying Fr\'echet Inception Distance in generative model evaluation, here applied to value distributions rather than image feature spaces to jointly capture location shift and structural dispersion:

\begin{equation}
\label{eq:wasserstein}
W_2^2(\mathcal{N}_m, \mathcal{N}_h) =
\underbrace{\|\mu_m - \mu_h\|_2^2}_{\text{Location shift}} +
\underbrace{\text{Tr}\left(\Sigma_m + \Sigma_h -
2\left(\Sigma_m^{1/2} \Sigma_h
\Sigma_m^{1/2}\right)^{1/2}\right)}_{\text{Structural dispersion}}
\end{equation}

The two components in Eq.~\eqref{eq:wasserstein} are directly interpretable: location shift $\|\mu_m - \mu_h\|_2^2$ captures systematic bias in the value centroid, and structural dispersion $\mathrm{Tr}(\Sigma_m + \Sigma_h - 2(\Sigma_m^{1/2}\Sigma_h\Sigma_m^{1/2})^{1/2})$ captures differences in the shape and spread of value distributions. Clustering hyperparameters and manifold construction details are provided in the Supplementary Information. The distribution radius $\sigma$ reported in Table~\ref{tab:value_profiles} and subsequent sections is computed separately, as the size of the confidence ellipse fitted to a model's value cloud in the two-dimensional UMAP-projected space, and should not be confused with the covariance $\Sigma$ defined above in the original ten-dimensional Schwartz space.

The displacement angle $\theta$ and the displacement distance $d$ (or $\Delta d$, when comparing two conditions such as Direct and CoT), reported alongside $\sigma$ throughout Sections~\ref{sec:2.3.1}, \ref{subsec:2.3.2-cognition}, and \ref{subsec:2.3.3-alignment}, are the angle and magnitude of a single shift vector in the same two-dimensional projected space: the vector pointing from a distribution's centroid to the human reference centroid. This vector should not be confused with the location-shift term $\|\mu_m-\mu_h\|_2^2$ in Eq.~\eqref{eq:wasserstein}, which is computed directly in the original ten-dimensional Schwartz space.

\subsection{The PEC framework: modeling the determinants of value expression}
\label{sec:methods_pec}

We propose the PEC framework as a unified account of how LLMs express values. Drawing on the conceptual apparatus of field theory~\cite{lewin1951field}, we model value expression as a function of three determinants:
\begin{equation}
\label{eq:pec}
    \mathbf{v} = C(P, E)
\end{equation}
where $P$ and $E$ are the two directly controllable factors: $P$ denotes the Prior, the latent value disposition encoded through pre-training and alignment procedures; $E$ denotes the Environment, the contextual framing supplied at inference time through system prompts and situational cues. Built upon these two factors, $C$ denotes Cognition, an emergent reasoning dimension reflecting the generation strategy employed, ranging from direct answering to Chain-of-Thought deliberation. $C(\cdot)$ serves as the mapping function that captures how Cognition transforms the interplay of Prior and Environment into the observed value vector $\mathbf{v}$.

These three factors are not independent additive components but constitute an integrated dynamic system: the Prior $P$ establishes a baseline attractor in value space; the Environment $E$ perturbs the trajectory around that attractor; and Cognition $C$ modulates the sensitivity and direction of that perturbation during inference. The mapping $C(P, E)$ is very likely nonlinear. For tractable modeling, we first approximate it locally around the neutral zero-shot baseline via a first-order expansion. In analogy with force composition in field theory, each factor thus contributes a component vector in the ten-dimensional value space, and these components may reinforce each other when aligned or partially cancel when opposed. The observed value vector $\mathbf{v}$ is the resultant of this superposition:
\begin{equation}
\label{eq:value_decomposition}
\mathbf{v} = \mathbf{v}_P + \Delta\mathbf{v}_E + \Delta\mathbf{v}_C + \boldsymbol{\varepsilon}
\end{equation}
where $\mathbf{v}_P$ is the neutral baseline of the LLM, serving as the anchor of the superposition, $\Delta\mathbf{v}_E$ captures the environmental displacement induced by contextual framing, and $\Delta\mathbf{v}_C$ captures the cognitive modulation introduced by CoT reasoning. The residual term $\boldsymbol{\varepsilon}$ absorbs factors that linear methods cannot analyze, such as the coupling between $E$ and $C$. The three paragraphs below operationalize each factor in turn.

\paragraph{Prior (Factor P): intrinsic value baseline.}
The neutral baseline $\mathbf{v}_P$ introduced in Eq.~\eqref{eq:value_decomposition} is measured as the model's output vector under a neutral, zero-shot setting (i.e., with an empty system prompt and direct answering mode). To quantify the impact of model architecture and training stages on these priors, we constructed a comprehensive corpus covering 106 distinct LLMs. This cohort spans representative \textbf{open-weight models} (including Llama-2/3, Qwen, Bloom~\cite{workshop2023bloom} families) and leading \textbf{proprietary closed-source models} (GPT, Gemini, Kimi, DeepSeek, Doubao), ensuring robust coverage across varying parameter scales and alignment paradigms.

We employ a multivariate linear regression analysis to decompose the variance in $\mathbf{v}_P$. Here, the superscript $d$ denotes a single scalar coordinate of the ten-dimensional value vector, i.e., $v_P^{(d)}$ is the $d$-th component of $\mathbf{v}_P$; this per-dimension notation is retained throughout all subsequent equations. For each value dimension $d \in \{1, \dots, 10\}$, we fit the following model:
\begin{equation}
\label{eq:factor_p_regression}
v_P^{(d)} = \beta_0 + \beta_1 \log(N_{\text{params}}) + \beta_2 \mathbb{I}_{\text{stage}} + \beta_3 \mathbb{I}_{\text{family}} + \epsilon
\end{equation}
In Eq.~\eqref{eq:factor_p_regression}, $N_{\text{params}}$ represents the parameter count (estimated for closed models), $\mathbb{I}_{\text{stage}}$ is a binary indicator for instruction tuning (0 for Base, 1 for Chat), and $\mathbb{I}_{\text{family}}$ encodes the model family. Here $\epsilon$ denotes the scalar regression noise for dimension $d$, distinct from the vector-valued residual $\boldsymbol{\varepsilon}$ in Eq.~\eqref{eq:value_decomposition}. The coefficients $\beta$ quantify the extent to which scaling laws and alignment techniques rigidly shift the model's default value coordinates.

\paragraph{Environment (Factor E): contextual susceptibility.}
To measure the deviation in values induced by external framing, we constructed a \textbf{Contextual Perturbation Dataset} in Fig.~\ref{fig:instruct_shift}. For each model, each Schwartz dimension $d$, and each of the four contextual factors $i$ defined in Section~\ref{sec:methods_administering}, we computed the value susceptibility matrix $\chi \in \mathbb{R}^{10 \times 4}$. Let $v_P^{(d)}$ denote the $d$-th component of the neutral baseline defined in Eq.~\eqref{eq:value_decomposition}, and let $v^{(d)}(e_i, p)$ be the value observed under contextual factor $i$ at pressure intensity $p$. For each dimension-factor pair $(d, i)$, we first fit a quadratic model to the observed displacement across the five pressure levels:
\begin{equation}
\hat{\chi}_{i}^{(d)}, \hat{\gamma}_{i}^{(d)} = \arg\min_{\chi_{i}^{(d)}, \gamma_{i}^{(d)}} \sum_{p} \left( \left| v^{(d)}(e_i, p) - v_P^{(d)} \right| - \left( |\chi_{i}^{(d)}| \cdot p - \gamma_{i}^{(d)} \cdot p^2 \right) \right)^2
\end{equation}
We then use the fitted coefficients to predict the displacement under any given pressure intensity $p$:
\begin{equation}
\label{eq:factor_e_sensitivity}
v_E^{(d)}=|\hat{\chi}_{d,i}| \cdot p - \hat{\gamma}_{d,i} \cdot p^2
\end{equation}
where $\hat{\chi}_{d,i}$ is the linear response coefficient for dimension $d$ under factor $i$. It captures how strongly dimension $d$ responds to factor $i$ under low pressure. $\hat{\gamma}_{d,i}$ is a quadratic saturation parameter. It captures the slowdown in response at high pressure. Repeating this fit across all ten dimensions and four factors yields the full susceptibility matrix $\chi \in \mathbb{R}^{10 \times 4}$. This matrix serves as a proxy for the model's pliability. It distinguishes dimensions and factors that maintain robust internal priors from those that exhibit high sensitivity under contextual framing.

\paragraph{Cognition (Factor C): reasoning-induced modulation.}
To determine how the inference process modulates value expression, we control the computation path by comparing two decoding modes: Direct-Answer and Chain-of-Thought (CoT). For the latter, we append the trigger ``Let's think step by step'' to induce intermediate reasoning tokens $R$. We define the \textbf{Cognitive Modulation Vector}, $\Delta\mathbf{v}_C$, as the vector difference between the reasoning-enhanced output and the direct output:

\begin{equation}
\label{eq:factor_c_modulation}
\Delta\mathbf{v}_{C} = \mathbf{v}_{CoT} - \mathbf{v}_{Direct}
\end{equation}

where $\mathbf{v}_{CoT}$ denotes the value-expression vector obtained when the model generates intermediate reasoning tokens through Chain-of-Thought, and $\mathbf{v}_{Direct}$ denotes the corresponding value-expression vector obtained from direct answering without explicit reasoning. Each vector represents the model's expressed preference distribution over the predefined value dimensions, and $\Delta\mathbf{v}_{C}$ captures the displacement of value expression induced by the reasoning process rather than the absolute value preference itself. By analyzing the distributional shift of $\Delta\mathbf{v}_{C}$ across the entire evaluation set, we quantify how reasoning processes systematically modulate LLM value expression under different models and contextual conditions.

\subsection{Alignment Prescription}
\label{sec:prescription}

A natural question follows from the PEC decomposition (Eq.~\eqref{eq:pec}): for a given model and a target value dimension requiring correction, which intervention achieves sufficient realignment at the lowest cost?

Before formalizing the prescription procedure, we clarify the division of labor between PEC and the intervention hierarchy. The ordering Prompt, CoT, SFT/DPO, and continued pre-training constitute a monotonically increasing cost ladder that follows directly from the computational architecture of LLM deployment; it is typically determined by empirical experience and experimental results. The PEC framework's contribution is to solve the \emph{matching problem} on this pre-existing ladder: given a model's ten Schwartz value dimensions, which respond sufficiently to prompt-level intervention, which require parametric updates, and which resist all available methods? Without PEC, this matching can only be performed through exhaustive experimental enumeration over every model--dimension--intervention combination. PEC replaces brute-force search with theory-guided prediction, and we formalize this below as a hierarchical prescription procedure.

The empirical analyses of Factors P, E, and C jointly reveal a systematic heterogeneity in how different Schwartz value dimensions respond to external intervention (Sections~\ref{sec:2.3.1}, \ref{subsec:2.3.2-cognition}, and \ref{subsec:2.3.3-alignment}). These three layers of evidence converge on a single structural conclusion: value dimensions differ not merely in their current alignment state, but in the class of intervention required to shift them. We therefore map the effect-size profile of each dimension onto a four-level prescription hierarchy, where the effect-size profile is defined as the maximum absolute shift $|\Delta v^{(d)}|$ achievable by each factor. Section~\ref{sec2.4} reports how this mapping resolves in practice for each model--dimension pair, and Table~\ref{tab:intervention_levels} summarizes the formal definition of each level. This mapping ensures that the prescription hierarchy is not an arbitrary ordering but a direct empirical consequence of the PEC decomposition.

\paragraph{Operationalizing PEC factors into comparable effect sizes.}
\label{sec:effect_sizes}
The three PEC factors take different mathematical forms: the susceptibility matrix $\chi$ characterizes Environment, the cognitive modulation vector $\Delta\mathbf{v}_C$ characterizes Cognition, and the regression-based profile of the baseline $\mathbf{v}_P$ characterizes Prior. To compare them on equal footing, we convert each into the same type of quantity. For every Schwartz dimension~$d$ and every intervention class $\mathcal{I} \in \{E, C, P\}$, we compute a non-negative scalar $\delta_{\mathcal{I}}^{(d)}$, defined as the absolute displacement in dimension $d$ induced by intervention $\mathcal{I}$, normalized by the intensity of that intervention. The three factors take the following specific forms.

\textbf{Environment displacement.} The environment displacement is computed directly from the calibrated susceptibility matrix $\chi$, aggregated across all contextual factors and normalized by pressure intensity:
\begin{equation}
\delta_E^{(d)} = \sum_{i=1}^{4} \frac{|\hat{\chi}_{d,i}| \cdot p - \hat{\gamma}_{d,i} \cdot p^2}{p}
\label{eq:delta_E}
\end{equation}
where $\hat{\chi}_{d,i}$ and $\hat{\gamma}_{d,i}$ are the response and saturation coefficients fitted in Eq.~\eqref{eq:factor_e_sensitivity}, and $p \in [0,1]$ is the pressure intensity serving as the intervention dose for E. Because $\delta_E^{(d)}$ is computed analytically from $\chi$, displacement predictions generalize to untested scenario combinations and to new models whose susceptibility profiles have been characterized, removing the need for exhaustive per-model, per-dimension prompt search. This predictive capacity is the core contribution of PEC to prescriptive alignment.

\textbf{Cognition displacement.} The effect of Chain-of-Thought reasoning on dimension~$d$ is normalized by the reasoning token overhead:
\begin{equation}
    \delta_C^{(d)} \;=\; \frac{\bigl|\, v^{(d)}(\mathrm{CoT}) \;-\; v^{(d)}(\mathrm{Direct}) \,\bigr|}{n_{\mathrm{tokens}}}
    \label{eq:delta_C}
\end{equation}
where $v^{(d)}(\mathrm{CoT})$ and $v^{(d)}(\mathrm{Direct})$ are the values under Chain-of-Thought and direct answering respectively, and $n_{\mathrm{tokens}}$ is the number of additional tokens generated by CoT.

\textbf{Prior displacement.} The parametric intervention effect is normalized by training set size:
\begin{equation}
    \delta_P^{(d)} \;=\; \frac{\bigl|\, v^{(d)}(\mathrm{Train}) \;-\; v^{(d)}(\mathrm{base}) \,\bigr|}{N_{\mathrm{train}}}
    \label{eq:delta_P}
\end{equation}
where $v^{(d)}(\mathrm{Train})$ is the value after parametric training, which in our experiments combines supervised fine-tuning (SFT) and direct preference optimization (DPO), $v^{(d)}(\mathrm{base})$ is the neutral zero-shot baseline $v_P^{(d)}$ of the base model before parametric training, and $N_{\mathrm{train}}$ is the number of training samples.

After this operationalization, each factor yields a non-negative scalar per dimension. All three scalars represent displacement per unit of intervention intensity and are expressed in the same Schwartz value space units, making them directly comparable.

\paragraph{Alignment threshold.}
An intervention is considered effective on a given dimension only if the induced shift exceeds a threshold~$\tau$. The choice of $\tau$ is guided by the field-theoretic~\cite{lewin1951field} view underlying the PEC framework. In this view, a value dimension only "moves" once the applied force exceeds a certain resistance, analogous to a threshold in a physical field. We calibrate this threshold empirically, by comparing it against the natural variation observed across national populations in the WVS-7 dataset. The median single-dimension mean difference between national populations in the Schwartz space provides a reference scale for what counts as a meaningful shift. We set $\tau$ within this empirically observed range, so that an intervention is only counted as effective if it produces a shift comparable to, or larger than, the value differences naturally observed between distinct human populations. This grounds $\tau$ in a real-world scale, rather than treating it as an arbitrary cutoff.

\paragraph{Hierarchical level assignment.}
\label{sec:level-assignment}
For each model and each of the ten Schwartz dimensions, we evaluate interventions in ascending order of cost. The prescribed level for a given dimension is the lowest level whose absolute shift magnitude surpasses~$\tau$; if no level succeeds, the dimension is classified as \emph{pre-train locked} and assigned to Level~4. Table~\ref{tab:intervention_levels} summarizes the four levels.

Cost increases monotonically along the ladder: $c_E < c_C < c_P$. Because of this, evaluating interventions in ascending order and stopping at the first one that reaches~$\tau$ is equivalent to selecting the lowest-cost sufficient intervention.

When no single intervention reaches~$\tau$ on its own, interventions are combined in ascending cost order. Section~\ref{subsec:2.3.2-cognition} documents an antagonistic interaction between Environment and Cognition factors. Activating CoT weakens the model's overall sensitivity to environmental factors. To account for this interaction, a discount coefficient $\alpha$ is applied to subsequent terms in the cumulative displacement. 
We define $\alpha$ as the ratio of the Frobenius norm of the susceptibility matrix $\chi$ under CoT to the Frobenius norm of $\chi$ without CoT:
\begin{equation}
\label{eq:alpha_definition}
\alpha \;=\; \frac{\|\chi_{\mathrm{CoT}}\|_F}{\|\chi_{\mathrm{Direct}}\|_F},
\end{equation}
We observe that for most models, generating a CoT reduces overall environmental sensitivity (See in Section~\ref{subsec:2.3.2-cognition}). As a result, $\alpha$ is typically below 1 for each model. The average value of $\alpha$ across models is 0.88.

Formally, the discounted cumulative displacement, evaluated in ascending cost order E, C, P, is defined as
\begin{equation}
\label{eq:cumulative_displacement}
    S^{(d)}_{\mathcal{I}} \;=\;
    \begin{cases}
        \delta_E^{(d)}, & \mathcal{I} = E \\
        \delta_E^{(d)} \;+\; \alpha \cdot \delta_C^{(d)}, & \mathcal{I} = C \\
        \delta_E^{(d)} \;+\; \alpha \cdot \delta_C^{(d)} \;+\; \alpha \cdot \delta_P^{(d)}, & \mathcal{I} = P
    \end{cases}
\end{equation}
The prescribed level is the lowest-cost intervention class $\mathcal{I}$ such that $S^{(d)}_{\mathcal{I}} > \tau$; if no such $\mathcal{I}$ exists, the dimension is diagnosed as pre-train locked and flagged for continued pre-training (Level~4).

To facilitate comparison of intervention efficiency, we additionally report an effect-per-cost ratio for each intervention class. The detailed definition, FLOP-based cost normalization, and per-model cost estimation are provided in Appendix~\ref{app:efficiency}. This metric serves only as a diagnostic measure of intervention efficiency and is not used as a criterion for determining the prescribed intervention level.

\paragraph{Intervention strategies.}
Based on the PEC framework, we implement four intervention strategies corresponding to the four levels defined in Table~\ref{tab:intervention_levels}, ordered by increasing computational cost. Each strategy is applied only when all lower-cost alternatives have been evaluated and found insufficient.

\begin{table}[!htbp]
\centering
\footnotesize
\setlength{\tabcolsep}{5pt}
\renewcommand{\arraystretch}{1.3}
\begin{tabular}{cl p{9cm}}
\toprule
\textbf{Level} & \textbf{Method} & \textbf{Description} \\
\midrule
1 & Prompt      & Value shift induced by prompt engineering alone. \\
2 & CoT         & Additional shift induced by Chain-of-Thought reasoning. \\
3 & Train       & Shift achieved through parametric training, specifically supervised fine-tuning (SFT) or direct preference optimization (DPO). \\
4 & Pre-train   & Continued pre-training on curated corpora, applied only when all preceding levels fail to exceed~$\tau$. \\
\bottomrule
\end{tabular}
\caption{Four-level intervention hierarchy ordered by ascending cost.
$\tau$ denotes the minimum shift magnitude required for a level to be
considered effective.}
\label{tab:intervention_levels}
\end{table}

\paragraph{Prescription validation.}
To verify the generalizability of the prescription matrix, we evaluate each model-dimension pair on a held-out dataset and compare the lowest effective level observed during validation with the prescribed level. The match rate across all pairs quantifies the reliability of the prescription framework. Detailed mathematical formulations of each intervention strategy, the prescription assignment algorithm, and the validation protocol are provided in Appendix~\ref{app:prescription}.

\subsection{Related Work}
This research is situated at the interdisciplinary frontier of LLM value evaluation and alignment. This section provides a comprehensive review of existing literature across two dimensions: value evaluation frameworks and alignment methodologies, while delineating our specific contributions and positioning within the field.

\subsubsection{Evaluation of LLMs' values}
Conventional value assessments typically rely on theoretical extensions of the ``Three Laws of Robotics'' \cite{asimov1950irobot} or machine ethics definitions \cite{moor2006nature}, employing descriptive probes through established frameworks such as the Moral Foundations Questionnaire (MFQ) \cite{graham2008moral} or Shweder's ``big three'' of morality \cite{shweder2013big}. Although Yao et al. \cite{yao-etal-2024-value, yao2025valuecompass} demonstrated the feasibility of mapping LLMs onto the Schwartz Theory of Basic Human Values, their evaluation still compresses each model into a single value vector.

Most current evaluation methods summarize an LLM's values into a single vector for comparison against human values. These methods do not perform this comparison in distributional form. As a result, a great deal of intrinsic structural information is lost. Serapio-Garc{\'\i}a et al. \cite{serapiogarcia2025psychometric} further proposed a psychometric framework for both evaluating and shaping personality traits in LLMs, providing methodological evidence that LLMs exhibit reliable and valid synthetic personality structures under appropriate prompting configurations. However, these approaches lack a unified computational framework capable of aligning ``model values'' with ``human socio-cultural benchmarks'' (such as national cultural disparities). Our study bridges this gap by introducing an uncertainty-centric perspective into the realm of value modeling.

\subsubsection{Value Alignment of LLMs}
As LLMs achieve human-level performance across a wide array of general tasks, ensuring that their underlying intentions, preferences, and behavioral norms remain consistent with human values has become a core issue in AI safety. Current alignment techniques are primarily categorized into three paradigms. First, SFT trains models on high-quality, human-curated datasets to directly inject intended behavioral norms into the LLMs' parameters \cite{wang2023self}. Second, RLHF utilizes preference data to construct reward models that guide the fine-tuning process \cite{nakano2021webgpt}. Constitutional AI further extends this paradigm by replacing human feedback with AI-generated critiques and principles, enabling scalable and rule-governed alignment without exhaustive human annotation \cite{bai2022constitutional}.

However, traditional RLHF methods often rely on a single scalar reward function, which possesses inherent limitations when navigating complex and potentially conflicting pluralistic value systems. To address this bottleneck, frontier research has begun to formalize value alignment within a Multi-Objective Reinforcement Learning (MORL) framework \cite{rodriguez2025multi}. By employing multi-dimensional reward vectors to explicitly represent distinct value dimensions, these methods first learn a set of Pareto-optimal policies within a multi-objective space. Subsequently, they utilize preference modeling or weight-adjustment mechanisms to achieve precise incentive compatibility and alignment with specific value systems during deployment. Finally, In-Context Alignment (ICA) has emerged as a parameter-efficient alternative, dynamically regulating model behavior through sophisticated system instructions or critiques during the inference phase \cite{ganguli2023capacity,gou2024critic}. Given its low computational overhead and compatibility with black-box models, ICA shows great promise for enhancing value consistency in large-scale applications.

Despite these methodological advancements, existing alignment strategies struggle to encompass potential and unforeseen risks. Significant challenges remain in ensuring the robustness and long-term consistency of value adherence in increasingly complex environments.

\section{Conclusion}
Together, these findings answer the three questions raised at the start of this study. LLMs do possess an intrinsic value system. This system can be quantified through the PEC framework. It can also be aligned through the Alignment Prescription. Beyond these specific answers, this study points to a broader shift in how we understand LLM behavior. Value expression has often been treated as noise, or as an unpredictable side effect of prompting. Our results show that this behavior follows a measurable structure instead. This structure can be traced back to its origin in the Prior, the Environment, and the Cognition factors. This shift moves value alignment from opaque trial-and-error toward a mechanistic and quantitative science. It also reduces the alignment tax through precise, prescriptive interventions. LLMs are taking on more decision-making roles today. In this setting, structural understanding becomes as important as raw performance. Our framework offers one concrete step toward this goal. It helps keep generative models as transparent and controllable tools, not unexamined ideological black boxes.

%%%%%%%%%%%%%%%%%%%%%%%%%%%%%%%%%%%%%%
%%%%%%%%%%% Data availability %%%%%%%
%%%%%%%%%%%%%%%%%%%%%%%%%%%%%%%%%%%%%%
\section{Data availability}\label{sec5}
The data used in this study include survey response data from the World Values Survey, which are publicly available for download at \url{https://www.worldvaluessurvey.org}. The processed questionnaire files and curated data splits used during the current study are available via GitHub at \url{https://github.com/KQ-Zhang/Do-LLMs-Have-Values-Release.git}.

% The data used in this study include survey response data from the World Values Survey, which are publicly available for download at \url{https://www.worldvaluessurvey.org}. The processed questionnaire files and curated data splits used during the current study are available via GitHub at \textbf{[Anonymized for Double-Blind Review]}.

%%%%%%%%%%%%%%%%%%%%%%%%%%%%%%%%%%%%%%
%%%%%%%%%%% Code availability %%%%%%%
%%%%%%%%%%%%%%%%%%%%%%%%%%%%%%%%%%%%%%
\section{Code availability}\label{sec6}
The code used during the current study, including scripts for data processing, model evaluation, and figure generation performed in Python (version 3.10), is available via GitHub at \url{https://github.com/KQ-Zhang/Do-LLMs-Have-Values-Release.git}. Python packages used for analysis include \texttt{pandas}, \texttt{numpy}, \texttt{scikit-learn}, \texttt{matplotlib}, and \texttt{transformers}. An interactive demonstration is also provided in the repository.

% The code used during the current study, including scripts for data processing, model evaluation, and figure generation performed in Python (version 3.10), is available via GitHub at \url{https://github.com/AnonymizedForReview/Do-LLMs-Have-Values}. Python packages used for analysis include \texttt{pandas}, \texttt{numpy}, \texttt{scikit-learn}, \texttt{matplotlib}, and \texttt{transformers}. An interactive demonstration will also be provided in the repository upon publication.

%%%%%%%%%%%%%%%%%%%%%%%%%%%%%%%%%%%%%
%%%%%%%%%%% References %%%%%%%%%%%%%%
%%%%%%%%%%%%%%%%%%%%%%%%%%%%%%%%%%%%%
\bibliography{sn-bibliography}% common bib file
% \section{References}\label{sec3}

% Sample body text. Sample body text. Sample body text. Sample body text. Sample body text. Sample body text. Sample body text. Sample body text. 

%%%%%%%%%%%%%%%%%%%%%%%%%%%%%%%%%%%%%%%
%%%%%%%%%%% Acknowledgements %%%%%%%%%%
%%%%%%%%%%%%%%%%%%%%%%%%%%%%%%%%%%%%%%%

\section{Acknowledgements}\label{sec8}
This work was supported by Beijing Natural Science Foundation (No. L251005, No. L251005), the National Natural Science Foundation of China (No. U24A20331, No. 62536002, No. 62192785, No. 62372451, No. 62372082, No. 62192782, No. 62532015), CAAI-Ant Group Research Fund (CAAI-MYJJ 2024-02), Young Elite Scientists Sponsorship Program by CAST (2024QNRC001).

\begin{appendices}

\section{WVS Human Data Analysis Results}
\label{app:background}

To enable interpretable analysis of real-world human values data, we adopt Schwartz Value Theory as the organizing framework. The theory characterizes human values along ten fundamental dimensions: Power, Achievement, Hedonism, Stimulation, Self-Direction, Universalism, Benevolence, Tradition, Conformity, and Security. These dimensions are arranged in a circumplex structure that captures the motivational compatibility and conflict among them, and the framework has been extensively validated across cultures. We map each WVS survey item onto this ten-dimensional space, reducing the high-dimensional raw survey data to a compact value vector per respondent. This representation preserves the theoretical meaning of each dimension and provides a principled, interpretable basis for subsequent quantitative analysis and cross-group comparison.

\begin{table}[!htbp]
\centering
\footnotesize
\setlength{\tabcolsep}{5pt}
\renewcommand{\arraystretch}{1.3}
\caption{
The ten basic human values from Schwartz's theory and their abbreviations used throughout this study.
}

\begin{tabular}{ll l}
\toprule
\textbf{Schwartz Value} & \textbf{Abbr.} & \textbf{Core Motivation} \\
\midrule
Power          & Pow.   & Social status, dominance, control over people and resources \\
Achievement    & Ach.   & Personal success through demonstrating competence \\
Hedonism       & Hed.   & Pleasure and sensuous gratification \\
Stimulation    & Stim.  & Excitement, novelty, and challenge in life \\
Self-Direction & S-Dir. & Independent thought and action \\
Universalism   & Univ.  & Understanding, tolerance, and protection of all people and nature \\
Benevolence    & Bene.  & Preserving and enhancing the welfare of close others \\
Tradition      & Trad.  & Respect and commitment to cultural or religious customs \\
Conformity     & Conf.  & Restraint of actions that may upset or harm others \\
Security       & Sec.   & Safety, harmony, and stability of society and relationships \\
\bottomrule
\end{tabular}

\label{tab:schwartz_abbr}
\end{table}

To ensure the objectivity and consistency of the item-to-dimension mapping, we employ an ensemble annotation strategy using five state-of-the-art LLMs: GPT-4o~\cite{openai2023gpt4o}, GPT-5~\cite{openai2025gpt5}, Gemini 2.5 Pro~\cite{google2024gemini}, Claude Sonnet 4.6~\cite{anthropic2024claude}, and DeepSeek-R1~\cite{deepseek2025r1}. Each model independently annotates every WVS item with the Schwartz dimensions it activates, and the final label set is determined by majority vote, with unanimous agreement indicating the highest annotation confidence. For item Q2P (``Please indicate how important friends are in your life''), all five models assigned both Stimulation (\texttt{ST01}) and Conformity (\texttt{CO01}), yielding a vote count of 5 for each dimension. This unanimity indicates high-confidence semantic alignment between the item and those dimensions. Aggregating annotations across multiple models reduces the label noise that would arise from relying on any single model.

Following annotation, we encode the response scale of each item into a normalized contribution score, mapping response options linearly onto $[0, 1]$ to quantify the degree to which a given response reflects endorsement of the associated value dimensions. For item Q2P, the options ``Very important'', ``Rather important'', ``Not very important'', and ``Not at all important'' receive scores of 1.00, 0.66, 0.33, and 0.00, respectively. Responses indicating non-response or refusal, such as ``I don't know'' or ``I prefer not to answer'', are excluded from subsequent computation. The Schwartz dimension score for respondent $i$ on dimension $k$ is then computed as the mean normalized score across all items associated with that dimension:

\begin{equation}
    v_k^{(i)} = \frac{1}{|Q_k|} \sum_{q \in Q_k} s_q^{(i)}, \quad k = 1, 2, \ldots, 10
    \label{eq:value_score}
\end{equation}

where $Q_k$ denotes the set of items annotated as activating dimension $k$, and $s_q^{(i)}$ is the normalized score of respondent $i$ on item $q$. Items annotated with multiple dimensions are counted once toward each associated dimension with equal weight. Each respondent is thus represented by a ten-dimensional vector $\mathbf{v}^{(i)} \in \mathbb{R}^{10}$, where each entry reflects the relative strength of one Schwartz value dimension and serves as a unified quantitative input for the analyses that follow.

For a full description of Schwartz's Theory of Basic Human Values and its cross-cultural validation, see Section~\ref{sec4:method}.

\begin{figure}[H]
    \centering
    \includegraphics[width=\textwidth]{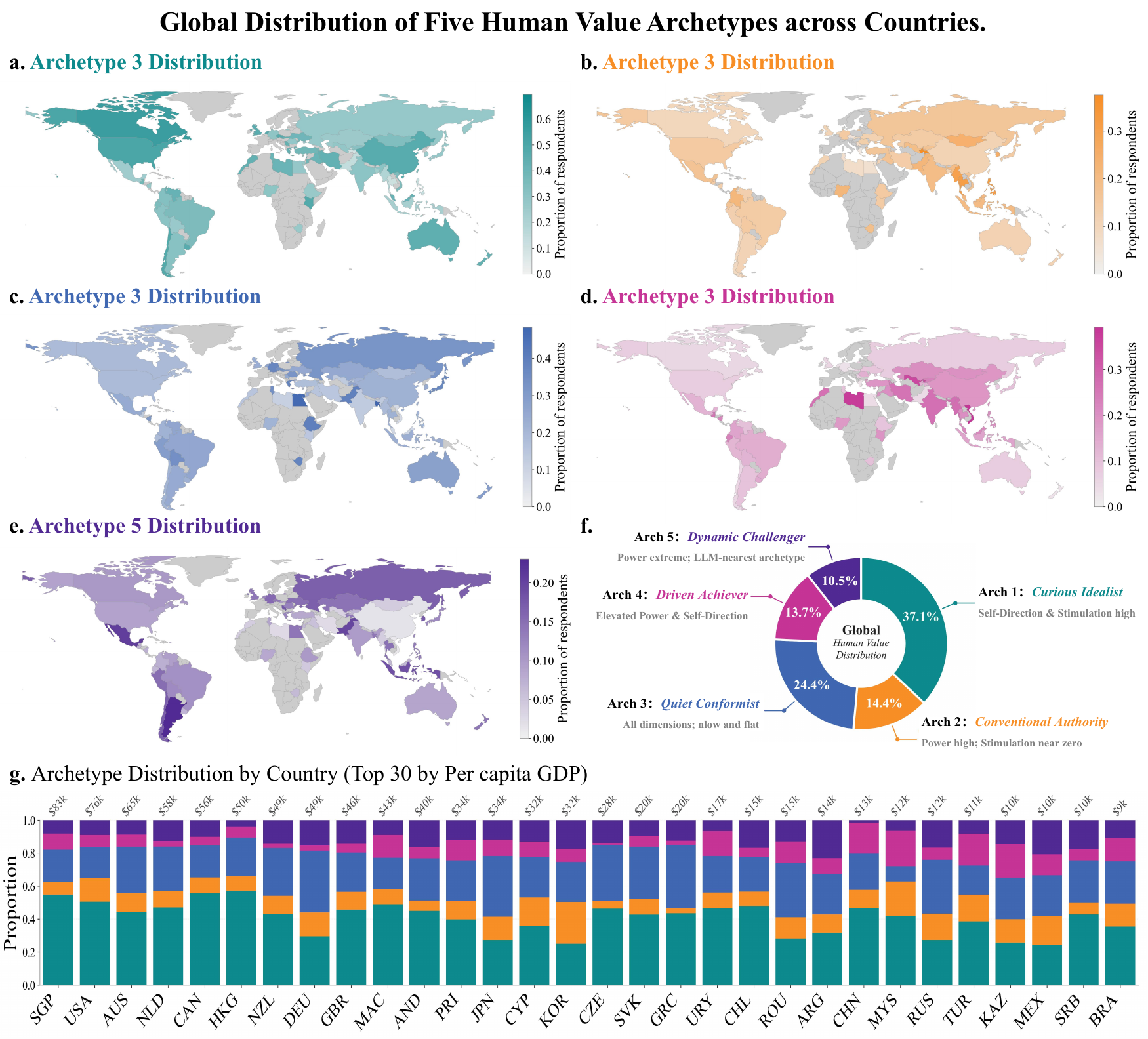} 
    \caption{\textbf{Global distribution of five human values archetypes across countries.}
    As world choropleth maps, \textbf{Panels (a)--(e)} visualize the country-level proportion of respondents assigned to each archetype. 
    The five archetypes are \textit{Curious Idealist} (Arch~1, teal), \textit{Conventional Authority} (Arch~2, orange), \textit{Quiet Conformist} (Arch~3, blue), \textit{Driven Achiever} (Arch~4, pink), and \textit{Dynamic Challenger} (Arch~5, purple).
    \textbf{Panel (f)} summarizes the global composition via a donut chart, with Arch~1 accounting for the largest share (37.1\%), followed by Arch~3 (24.4\%), Arch~2 (14.4\%), Arch~4 (13.7\%), and Arch~5 (10.5\%).
    \textbf{Panel (g)} shows archetype composition by country, ordered by decreasing GDP per capita.}
    \label{fig:archetype_global_distribution}
\end{figure}

Clustering the Schwartz value vectors of WVS respondents yields five human values archetypes (Fig.~\ref{fig:archetype_global_distribution}). \textit{Curious Idealist} (Arch~1, 37.1\%) is the most prevalent archetype globally, characterized by elevated Self-Direction and Stimulation scores, reflecting a strong orientation toward novel experience, independent thought, and personal exploration. It is most concentrated in Anglophone countries and East Asia. \textit{Conventional Authority} (Arch~2, 14.4\%) exhibits high Power scores alongside near-zero Stimulation, indicating strong deference to social hierarchy and established order with minimal appetite for external novelty. \textit{Quiet Conformist} (Arch~3, 24.4\%) scores uniformly low across all Schwartz dimensions, representing a subdued and undifferentiated value profile with no salient orientation; it is most prevalent in South Asia, Southeast Asia, and parts of the Middle East. \textit{Driven Achiever} (Arch~4, 13.7\%) combines elevated Power and Self-Direction, reflecting an active pursuit of personal achievement and social status. \textit{Dynamic Challenger} (Arch~5, 10.5\%) is the smallest archetype globally, defined by extreme Power scores and prominent Self-Direction. It occupies the most radical position in the Schwartz value space and is geographically concentrated in South America and Eastern Europe. Notably, Arch~5 is also the human archetype closest in value profile to current LLMs. Within the human population, this means that LLM value tendencies correspond to a minority group defined by extreme power orientation and high self-direction. This finding provides an important reference point for understanding and evaluating value bias in LLMs.

\section{Formal Definitions of Swing Experiment Metrics}
\label{app:swing_metrics}

For each query $q$ in the evaluation set, a model $\mathcal{M}$ is prompted $N = 5$ times under identical conditions (fixed temperature $T = 0.2$, constant inference parameters). Let $\mathcal{A} = \{a_1, a_2, a_3, a_4, a_5\}$ denote the multiset of responses collected across these $N$ independent runs, where each $a_i$ is a discrete categorical label drawn from the answer space $\mathcal{V}$.

\subsection*{Response Frequency}

Let $f(v) = |\{i : a_i = v\}|$ denote the raw count of answer $v \in \mathcal{V}$ within $\mathcal{A}$. The empirical response distribution over $\mathcal{A}$ is:
\begin{equation}
    p(v) = \frac{f(v)}{N}, \quad v \in \mathcal{V}
\end{equation}

\subsection*{Consistency}

Consistency measures the concentration of responses on the plurality answer. Let $v^* = \arg\max_{v \in \mathcal{V}} f(v)$ denote the majority label (the answer appearing most frequently across $N$ runs). Consistency is defined as:
\begin{equation}
    \mathrm{Cons}(q) = \frac{f(v^*)}{N} = \max_{v \in \mathcal{V}}\, p(v)
\end{equation}
$\mathrm{Cons}(q) \in [1/N,\, 1]$. A value of 1 indicates the model produces the same answer on every run; a value of $1/N$ indicates maximal dispersion across $N$ distinct answers.

\subsection*{Entropy}

Entropy quantifies the distributional uncertainty of responses across runs. We adopt the Shannon entropy with natural logarithm (in nats):
\begin{equation}
    H(q) = -\sum_{v \in \mathcal{V}} p(v)\ln p(v)
\end{equation}
where the sum is taken over all $v$ with $p(v) > 0$. $H(q) \in [0,\, \ln N]$. Zero entropy corresponds to a perfectly consistent model; maximal entropy $\ln N \approx 1.609$ corresponds to a uniform distribution across $N$ distinct answers.

Note that Consistency and Entropy are monotonically related through the response distribution: high Consistency implies low Entropy, and vice versa. Together they provide complementary characterisations of the same underlying swing phenomenon.

\subsection*{Accuracy and Correct Count}

Let $y^*$ denote the ground-truth label for query $q$. The per-run correctness indicator is $\mathbf{1}[a_i = y^*]$. The correct count and accuracy are defined as:
\begin{align}
    n_{\text{correct}}(q) &= \sum_{i=1}^{N} \mathbf{1}[a_i = y^*] \\[4pt]
    \mathrm{Acc}(q) &= \frac{n_{\text{correct}}(q)}{N}
\end{align}
$\mathrm{Acc}(q) \in \{0,\, 1/N,\, \ldots,\, 1\}$.

\subsection*{Consistency and Accuracy Gap}

The divergence between a model's behavioural stability and its factual correctness on query $q$ is measured as:
\begin{equation}
    \delta(q) = \mathrm{Cons}(q) - \mathrm{Acc}(q)
\end{equation}
A positive $\delta(q)$ indicates that the model repeatedly commits to a wrong answer: the swing is bounded (high Consistency) but anchored to an incorrect attractor (low Accuracy), corresponding to the hallucination zone identified in Fig.~\ref{2.1-swing-rigid}. The dataset-level gap is the mean over all queries:
\begin{equation}
    \bar{\delta} = \frac{1}{|Q|}\sum_{q \in Q} \delta(q) = \overline{\mathrm{Cons}} - \overline{\mathrm{Acc}}
\end{equation}
For GPT-4o on the emoji prediction task this yields $\bar{\delta} = +0.39$; for DeepSeek-V3 it yields $\bar{\delta} = +0.53$, confirming that both models fall into the hallucination zone on subjective tasks.

% 可分性
\section{LLMs' Values Evolution}

\begin{figure}[H]
    \centering
    \includegraphics[width=\textwidth]{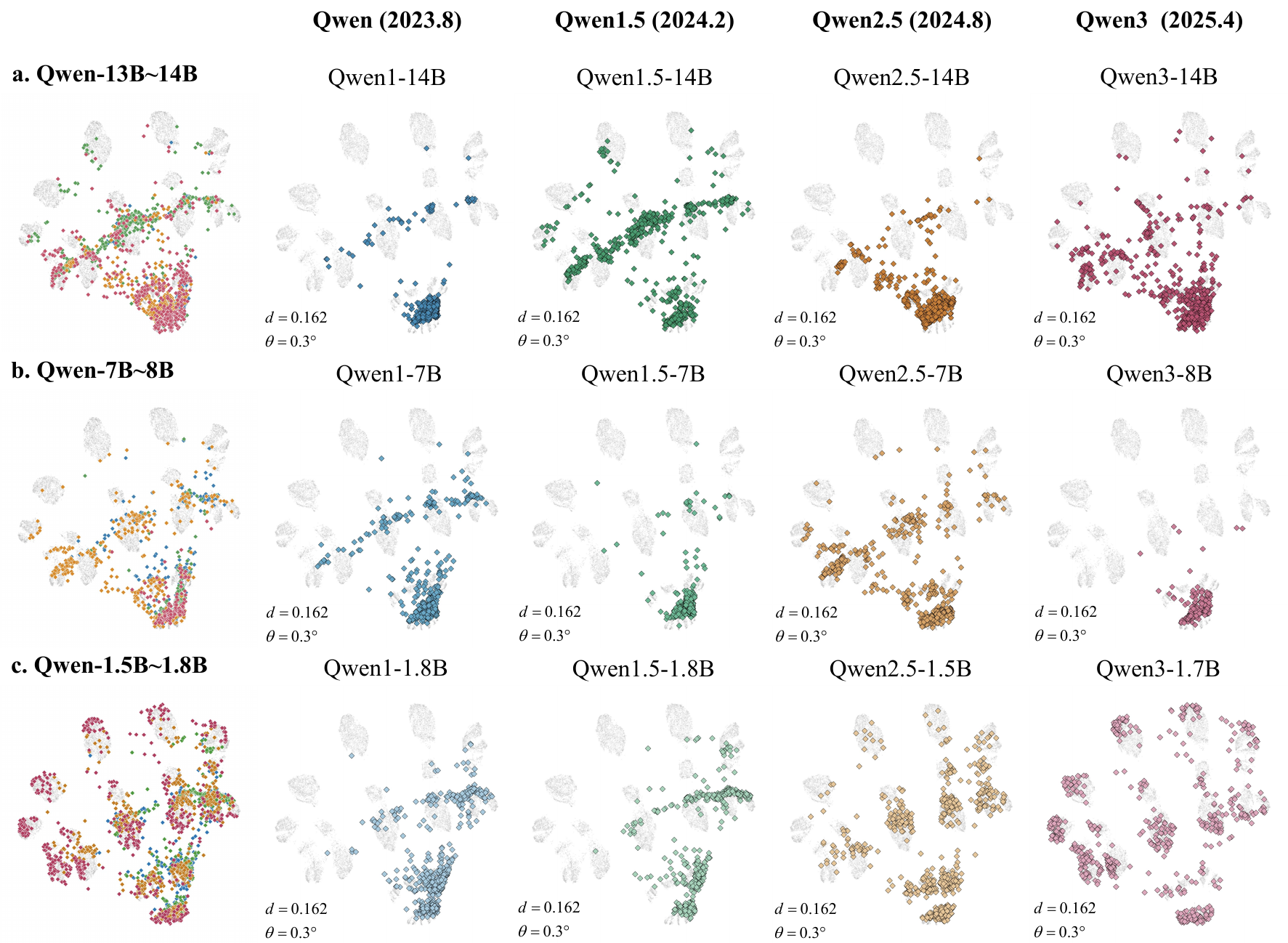} 
    \caption{\textbf{Generational and scale-wise evolution of value distributions across Qwen model variants in UMAP space.}
    Each panel projects a Qwen model variant onto a shared UMAP embedding of the ten-dimensional Schwartz value space, with the human values baseline shown as grey points for reference. Colored points represent the value distribution of a specific model variant, and each row corresponds to a distinct generation or parameter scale within the Qwen family. Across generations, a consistent trend of value Crystallization is observed: earlier and smaller models (top rows) produce diffuse, scattered distributions that partially overlap with the human baseline, whereas later and larger models (bottom rows) converge into tighter, more concentrated clusters in value space. These clusters are located progressively farther from the human distribution. Within the same generation, scaling up parameter count further amplifies this consolidation effect, producing distributions with smaller variance and stronger directional coherence. Together, these panels illustrate how iterative alignment procedures and model scaling jointly drive Qwen models toward a progressively more rigid and idealized value configuration, corroborating the value crystallization hypothesis advanced in the main text.}
    \label{fig:qwen_umap_evolution}
\end{figure}

\begin{figure}[H]
    \centering
    \includegraphics[width=\textwidth]{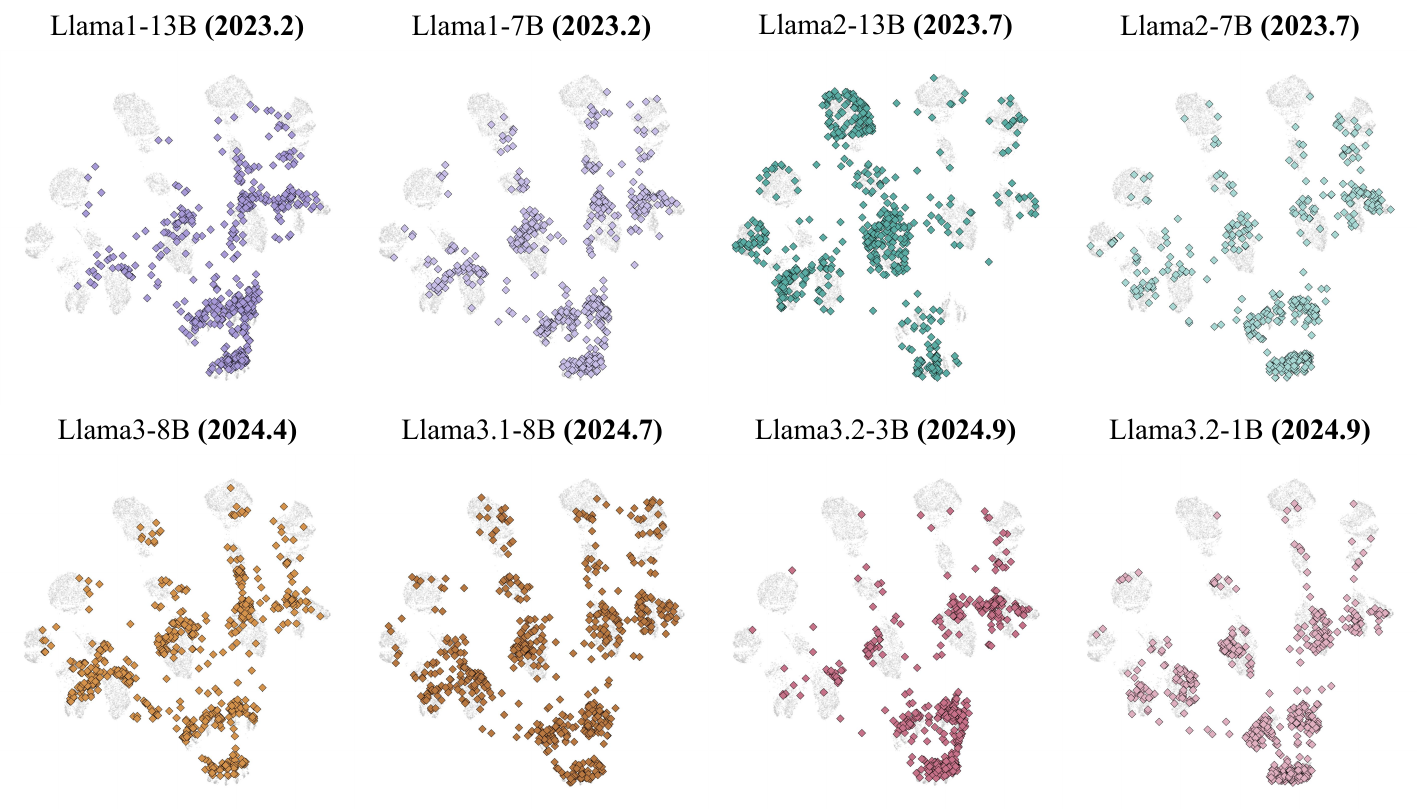} 
    \caption{\textbf{Value distribution evolution across Llama model generations.}
    Each subplot shows the UMAP projection of value vectors for a single base model. Rows correspond to model generations: Llama1 and Llama2 (top row), Llama3, Llama3.1, and Llama3.2 (bottom row). Columns within each generation are ordered by parameter scale in descending order. Release dates are indicated in bold next to each model name.
    Grey points represent the WVS-7 human baseline.}
    \label{fig:llama_evolution}
\end{figure}

\begin{figure}[H]
    \centering
    \includegraphics[width=\textwidth]{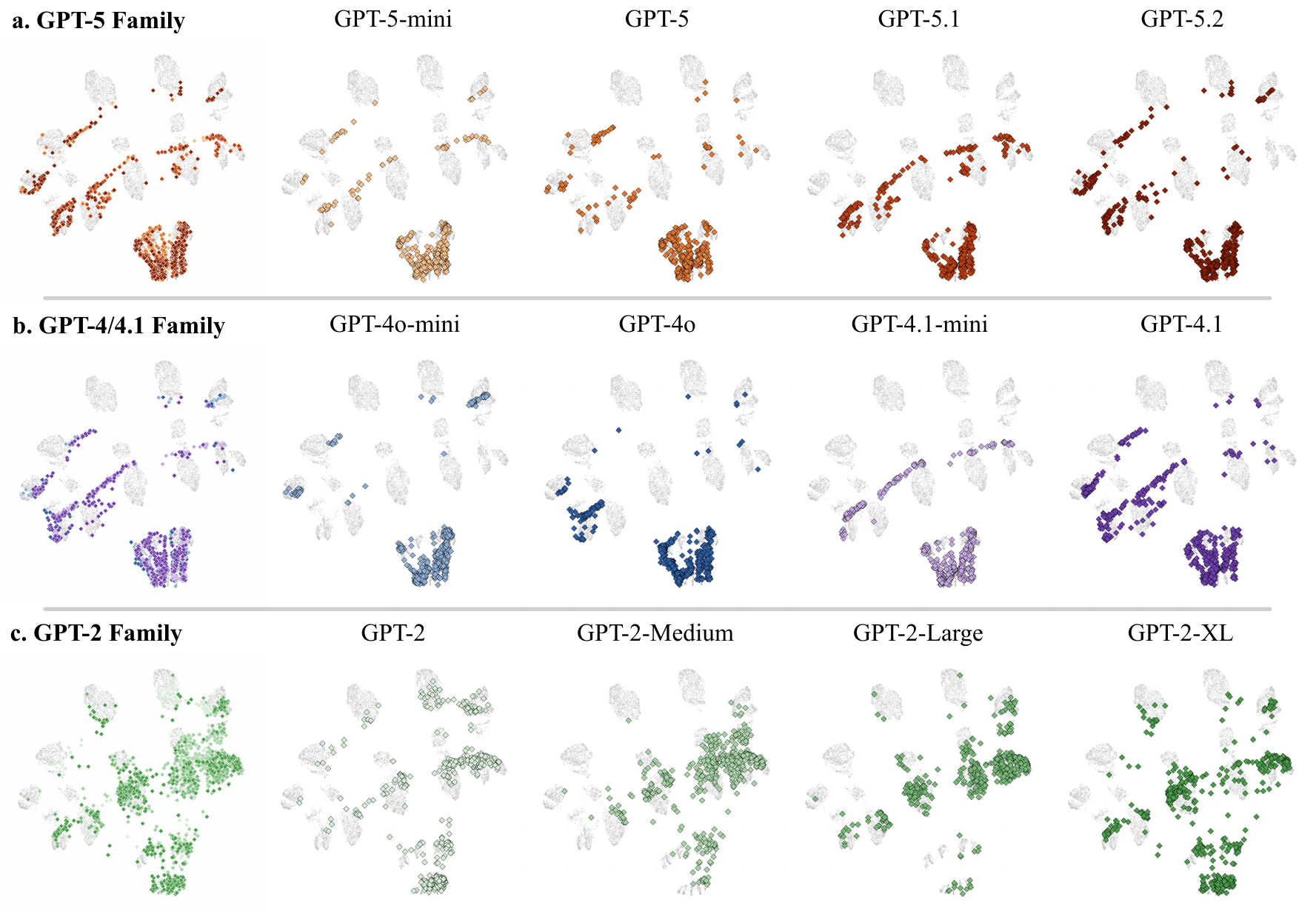} 
    \caption{\textbf{Value distribution evolution across GPT model families.}
    Each subplot shows the UMAP projection of value vectors for a single model.
    Rows correspond to model families: \textbf{a}, GPT-5 family (GPT-5-mini, GPT-5, GPT-5.1, GPT-5.2)~\cite{openai2025gpt5};
    \textbf{b}, GPT-4/4.1 family (GPT-4o-mini, GPT-4o, GPT-4.1-mini, GPT-4.1)~\cite{openai2023gpt4o};
    \textbf{c}, GPT-2 family (GPT-2, GPT-2-Medium, GPT-2-Large, GPT-2-XL)~\cite{radford2019gpt2}.
    Within each row, the leftmost subplot shows the full family overview and the remaining subplots highlight individual models.
    Hollow diamonds indicate smaller or earlier variants; filled diamonds indicate larger or more recent ones.
    Grey points represent the WVS-7 human baseline.}
    \label{fig:gpt_generational_value_shift}
\end{figure}

\begin{table}[t]
\centering
\small
\resizebox{\textwidth}{!}{%
\begin{tabular}{llcc|llcc|llcc}
\toprule
Families & Model & $d$ & $\theta$ & Families & Model & $d$ & $\theta$ & Families & Model & $d$ & $\theta$ \\
\midrule
\multirow{12}{*}{Qwen}
& 1-1.8B   & 6.081 & 25.8 & \multirow{8}{*}{Llama} & 1-7B   & 2.873 & 10.2 & \multirow{12}{*}{GPT} & 2        & 4.818 & 13.4 \\
& 1-7B     & 5.424 & 20.5 &                        & 1-13B  & 4.141 & 16.9 &                       & 2-Medium & 4.462 & 19.3 \\
& 1-14B    & 6.119 & 19.6 &                        & 2-7B   & 4.023 &  9.5 &                       & 2-Large  & 3.812 & 14.9 \\
& 1.5-1.8B & 6.589 & 25.5 &                        & 2-13B  & 1.622 & 12.7 &                       & 2-XL     & 3.064 & 21.1 \\
& 1.5-7B   & 7.112 & 24.1 &                        & 3-8B   & 2.264 &  6.6 &                       & 4o-mini  & 4.728 &  8.0 \\
& 1.5-14B  & 3.106 &  8.7 &                        & 3.1-8B & 1.704 &  6.1 &                       & 4o       & 5.275 &  3.0 \\
& 2.5-1.5B & 3.027 & 13.2 &                        & 3.2-1B & 2.349 &  8.0 &                       & 4.1-mini & 5.032 &  5.6 \\
& 2.5-7B   & 4.072 &  8.6 &                        & 3.2-3B & 4.889 & 16.8 &                       & 4.1      & 4.823 &  2.5 \\
& 2.5-14B  & 6.379 &  5.5 &                        &        &       &      &                       & 5-mini   & 5.079 & 10.9 \\
& 3-1.7B   & 1.255 &  2.1 &                        &        &       &      &                       & 5        & 3.876 &  6.3 \\
& 3-8B     & 5.621 & 18.7 &                        &        &       &      &                       & 5.1      & 6.723 &  5.1 \\
& 3-14B    & 3.853 &  0.1 &                        &        &       &      &                       & 5.2      & 4.834 &  3.9 \\
\bottomrule
\end{tabular}%
}
\caption{Value distribution statistics for Qwen, Llama, and GPT base models shown in Appendix Figs.~\ref{fig:qwen_umap_evolution}--~\ref{fig:gpt_generational_value_shift}. $\theta$ values are in degrees.}
\label{tab:appendix_umap_stats}
\end{table}

\subsection{Physical Analogy for the Susceptibility Matrix}
\label{app:susceptibility_physics}

We use a concept from physics to motivate the design of the value susceptibility matrix $\chi$. This section explains the analogy in detail.

\paragraph{Susceptibility in physics.}
In physics, susceptibility describes how a system responds to an external field. Magnetic susceptibility is one common example. It measures how much a material becomes magnetized under an applied magnetic field. A material with high susceptibility changes its state substantially under a weak field. A material with low susceptibility resists such change, even under a strong field. In the linear regime, the response is proportional to the field strength. The proportionality constant is the susceptibility. Under a strong field, many materials deviate from this linear behavior. The response grows more slowly, and eventually saturates. This nonlinear behavior is typically captured by adding a higher order correction term to the linear model.

\paragraph{Mapping the analogy to LLM value expression.}
We treat the contextual prompt as an external field applied to the model. We treat the resulting value shift as the model's response to this field. This mapping is consistent with the field-theoretic framing introduced in Section~\ref{subsec:PEC}. There, Environment ($E$) is defined as the external field acting on the model.

Following this analogy, we define the value susceptibility matrix $\chi \in \mathbb{R}^{10 \times 4}$. Each entry $\chi_{d,j}$ measures how strongly value dimension $d$ responds to environmental factor $j$. A large $\chi_{d,j}$ indicates that the value dimension is easily perturbed by that environmental factor. A small $\chi_{d,j}$ indicates that the value dimension remains stable under the same perturbation.

\paragraph{Linear response and saturation.}
At low pressure intensity $p$, the value shift grows approximately linearly with $p$. The slope of this linear growth is $\chi_{d,j}$. This mirrors the linear response regime in physics, where the response is proportional to the field strength.

At higher pressure intensity, the growth slows down. This deviation from linearity is captured by a quadratic correction term $\gamma_{d,j}$. The full approximation is given by
% \begin{equation}
%     |\, v^{(d)}(e_j, p_j^*) - v^{(d)}(\varnothing) \,| \approx |\chi_{d,j}| \cdot p_j^* - \gamma_{d,j} \cdot (p_j^*)^2.
% \end{equation}
\begin{equation}
    |\, v^{(d)}(e_j, p_j^*) - v_P^{(d)} \,| \approx |\chi_{d,j}| \cdot p_j^* - \gamma_{d,j} \cdot (p_j^*)^2.
\end{equation}
This form mirrors saturation effects observed in physical systems. In a magnetic material, magnetization increases linearly under a weak field. It levels off as the field strength approaches saturation. In our setting, a value dimension shifts steadily under moderate pressure. Under extreme pressure, the shift slows or reverses. We interpret this as a sign of value rigidity under strong situational stress.

\paragraph{Why this framing is useful.}
This physics-inspired formulation gives value change a clear structural meaning. It is not treated as an isolated numerical fluctuation. It is treated as a structured response, governed by a field-response relationship. This framing also supports prediction. We first estimate $\chi$ and $\gamma$ from a small number of tested conditions. Then we can approximate the displacement under an untested pressure level, without exhaustive enumeration. This is the same logic used in physics. Once a material's susceptibility is measured, it predicts the material's response under new field conditions.

\subsection{Prompts effect}

\begin{figure}[H]
    \centering
    \includegraphics[width=\textwidth]{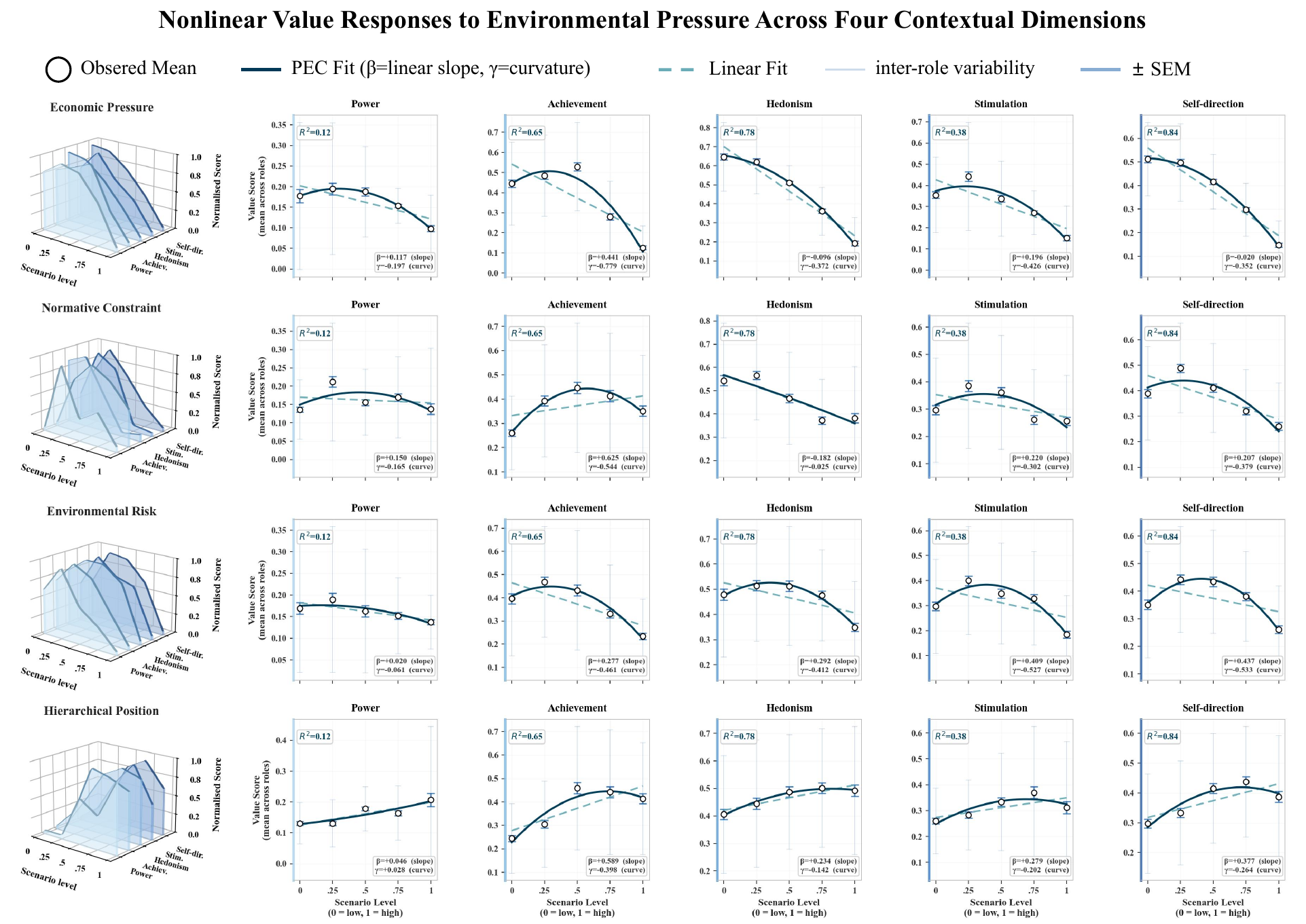}
    \caption{\textbf{Quadratic modeling of the relationship between contextual scenario intensity and LLM value expression across four environmental dimensions.}
    Each row corresponds to one of four contextual framing conditions (Economic Pressure, Normative Constraint, Environmental Risk, and Hierarchical Position), operationalized as a continuous intensity scale from 0 (absent) to 1 (maximal).
    The leftmost column presents a three-dimensional surface visualizing the joint response of five Schwartz value dimensions (Power, Achievement, Hedonism, Stimulation, and Self-Direction) as a function of scenario intensity.
    The remaining columns display the univariate response for each value dimension, where open circles denote observed means, the solid curve represents the fitted quadratic model, the dashed line provides the linear baseline, and vertical bars indicate $\pm$1 standard error and $\pm$1 standard deviation across roles.
    Reported coefficients $\chi$ and $\gamma$ denote the linear (susceptibility) slope and the quadratic (saturation) coefficient of each fitted trajectory, consistent with the definitions in Section~\ref{app:susceptibility_physics}.}
    \label{fig:context_value_quadratic_modeling}
\end{figure}

Fig.~\ref{fig:context_value_quadratic_modeling} presents the quadratic fits across all four contextual framing conditions and value dimensions. The linear coefficient $\chi$ captures the directional sensitivity of each value dimension to escalating scenario intensity: a positive $\chi$ indicates that the value score increases with intensity, whereas a negative $\chi$ indicates that the value score decreases with intensity. The curvature coefficient $\gamma$ captures nonlinearity: a negative $\gamma$ corresponds to an inverted-U trajectory, reflecting peak activation followed by decline, while a positive $\gamma$ corresponds to a U-shaped trajectory, reflecting initial suppression followed by recovery. The goodness of fit is reported as the cross-validated $R^{2}$ of the quadratic model. The consistent nonlinearity observed across dimensions and conditions confirms that LLM value expression is not a monotonic function of scenario intensity. Rather, contextual framing induces structured, dimension-specific trajectories of value activation and suppression.

\subsection{CoT effect}

\begin{figure}[H]
    \centering
    \includegraphics[width=\textwidth]{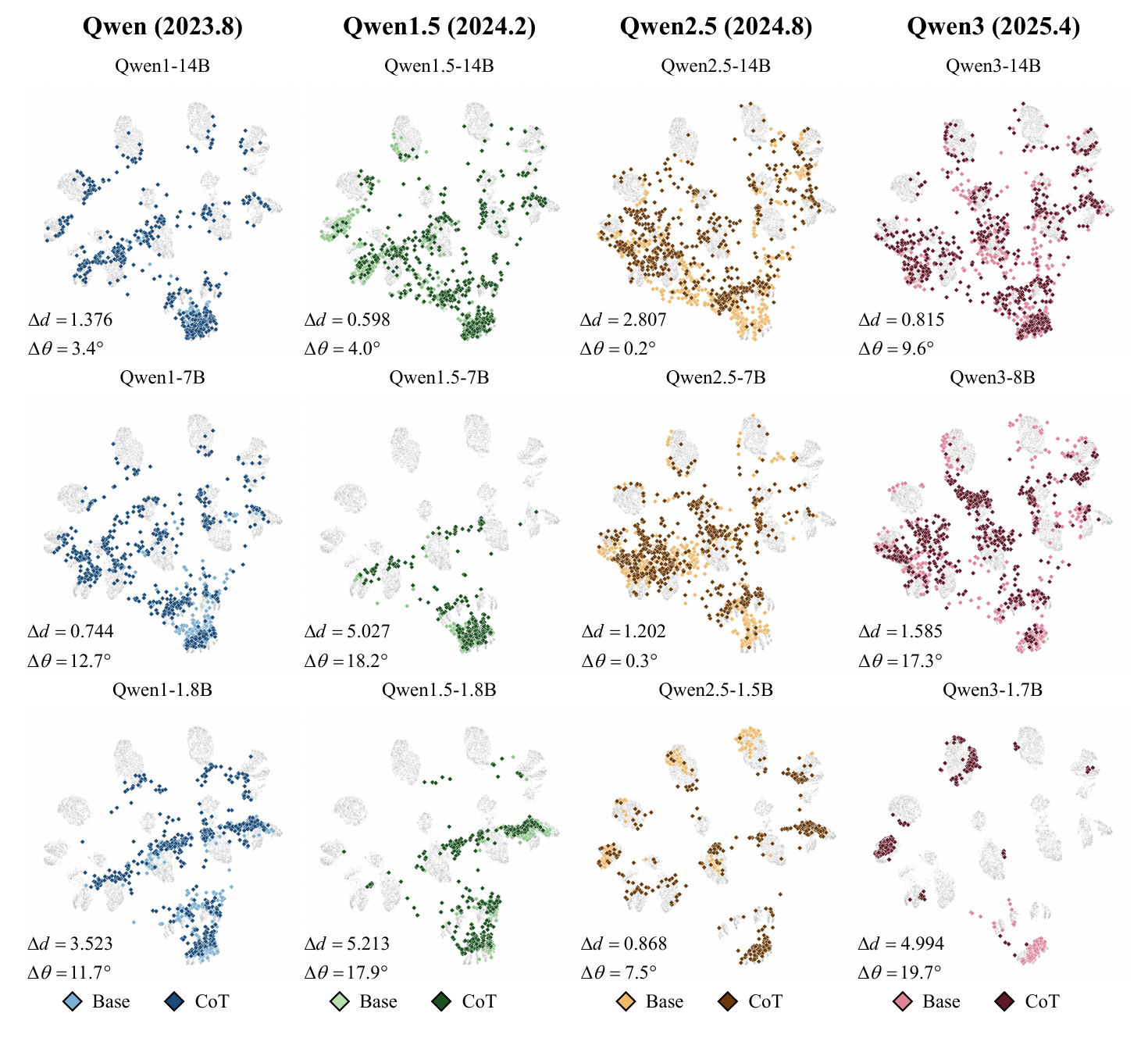} 
    \caption{\textbf{Effect of Chain-of-Thought generation on value distributions across Qwen model families.} Each subplot shows the UMAP projection of value vectors for a single model under two conditions: Base (hollow diamond, light color) and Chain-of-Thought prompting (CoT, filled diamond, dark color). Columns correspond to model generations (Qwen, Qwen1.5, Qwen2.5, Qwen3), and rows correspond to parameter scales ($\sim$14B, $\sim$7--8B, $\sim$1.5--1.8B). Grey points represent the WVS-7 human baseline.}
    \label{fig:appendix_cot_qwen}
\end{figure}

\begin{table*}[t]
\centering
\small
\caption{Effect of Chain-of-Thought prompting on the Qwen family. $\theta$ values are in degrees.}
\label{tab:qwen_cot_result}
 
\begin{tabular}{llcccccc}
\toprule
Series & Model & $d_{\text{direct}}$ & $d_{\text{CoT}}$ & $\Delta d$ & $\theta_{\text{direct}}$ & $\theta_{\text{CoT}}$ & $\Delta\theta$ \\
\midrule
 
\multirow{6}{*}{Qwen1}
& 1.8B      & 6.081 & 2.558 & 3.523 & 25.8 & 14.0 & 11.7 \\
& 1.8B-Chat & 5.407 & 3.450 & 1.956 & 27.1 & 15.5 & 11.6 \\
& 7B        & 5.424 & 4.679 & 0.744 & 20.5 & 7.8  & 12.7 \\
& 7B-Chat   & 5.537 & 2.186 & 3.352 & 16.5 & 3.9  & 12.6 \\
& 14B       & 6.119 & 4.744 & 1.376 & 19.6 & 16.2 & 3.4  \\
& 14B-Chat  & 6.930 & 3.451 & 3.479 & 6.8  & 8.1  & 1.3  \\
\midrule
 
\multirow{10}{*}{Qwen1.5}
& 0.5B      & 1.383 & 1.483 & 0.100 & 4.3  & 8.2  & 3.9  \\
& 0.5B-Chat & 5.234 & 2.221 & 3.013 & 15.9 & 8.7  & 7.2  \\
& 1.8B      & 6.588 & 1.376 & 5.213 & 25.5 & 7.6  & 17.9 \\
& 1.8B-Chat & 7.073 & 5.413 & 1.659 & 42.6 & 24.7 & 17.9 \\
& 4B        & 5.015 & 4.754 & 0.261 & 19.6 & 17.7 & 1.9  \\
& 4B-Chat   & 2.523 & 3.454 & 0.931 & 1.2  & 0.7  & 0.4  \\
& 7B        & 7.112 & 2.084 & 5.027 & 24.1 & 5.9  & 18.2 \\
& 7B-Chat   & 6.593 & 6.054 & 0.539 & 4.2  & 3.6  & 0.6  \\
& 14B       & 3.106 & 3.704 & 0.598 & 8.7  & 4.7  & 4.0  \\
& 14B-Chat  & 5.346 & 3.064 & 2.282 & 45.1 & 10.2 & 34.9 \\
\midrule
 
% \multirow{2}{*}{Qwen2}
% & 7B        & 6.054 & $-$ & $-$ & 4.2 & $-$ & $-$ \\
% & 7B-Instruct & 5.819 & $-$ & $-$ & 8.5 & $-$ & $-$ \\
% \midrule
 
\multirow{10}{*}{Qwen2.5}
& 0.5B          & 2.221 & 4.060 & 1.839 & 6.9  & 17.0 & 10.1 \\
& 0.5B-Instruct & 0.704 & 2.656 & 1.952 & 3.6  & 8.9  & 5.4  \\
& 1.5B          & 3.027 & 2.159 & 0.868 & 13.2 & 5.7  & 7.5  \\
& 1.5B-Instruct & 2.510 & 4.651 & 2.141 & 2.3  & 4.8  & 2.5  \\
& 3B            & 1.810 & 2.172 & 0.363 & 8.9  & 1.6  & 7.4  \\
& 3B-Instruct   & 5.130 & 2.527 & 2.603 & 13.8 & 17.6 & 3.8  \\
& 7B            & 4.072 & 2.871 & 1.202 & 8.6  & 8.9  & 0.3  \\
& 7B-Instruct   & 4.448 & 1.900 & 2.547 & 64.1 & 25.5 & 38.6 \\
& 14B           & 6.379 & 3.572 & 2.807 & 5.5  & 5.7  & 0.2  \\
& 14B-Instruct  & 2.903 & 1.674 & 1.228 & 8.2  & 8.9  & 0.7  \\
\midrule
 
\multirow{10}{*}{Qwen3}
& 0.6B          & 2.406  & 6.718 & 4.312 & 9.0  & 27.2 & 18.2 \\
& 0.6B-Instruct & 10.603 & 3.881 & 6.722 & 65.9 & 3.4  & 62.5 \\
& 1.7B          & 1.255  & 6.248 & 4.994 & 2.1  & 21.8 & 19.7 \\
& 1.7B-Instruct & 5.739  & 5.471 & 0.268 & 15.3 & 0.3  & 15.0 \\
& 4B            & 6.361  & 5.088 & 1.273 & 0.2  & 1.4  & 1.2  \\
& 4B-Instruct   & 0.851  & 0.573 & 0.277 & 5.2  & 1.0  & 4.2  \\
& 8B            & 5.621  & 4.036 & 1.585 & 18.7 & 1.4  & 17.3 \\
& 8B-Instruct   & 1.802  & 0.795 & 1.006 & 26.1 & 10.4 & 15.6 \\
& 14B           & 3.853  & 3.038 & 0.815 & 0.1  & 9.6  & 9.6  \\
& 14B-Instruct  & 2.424  & 1.694 & 0.730 & 5.0  & 1.6  & 3.3  \\
\bottomrule
\end{tabular}
\end{table*}

\clearpage

\begin{figure}[H]
    \centering
    \includegraphics[width=\textwidth]{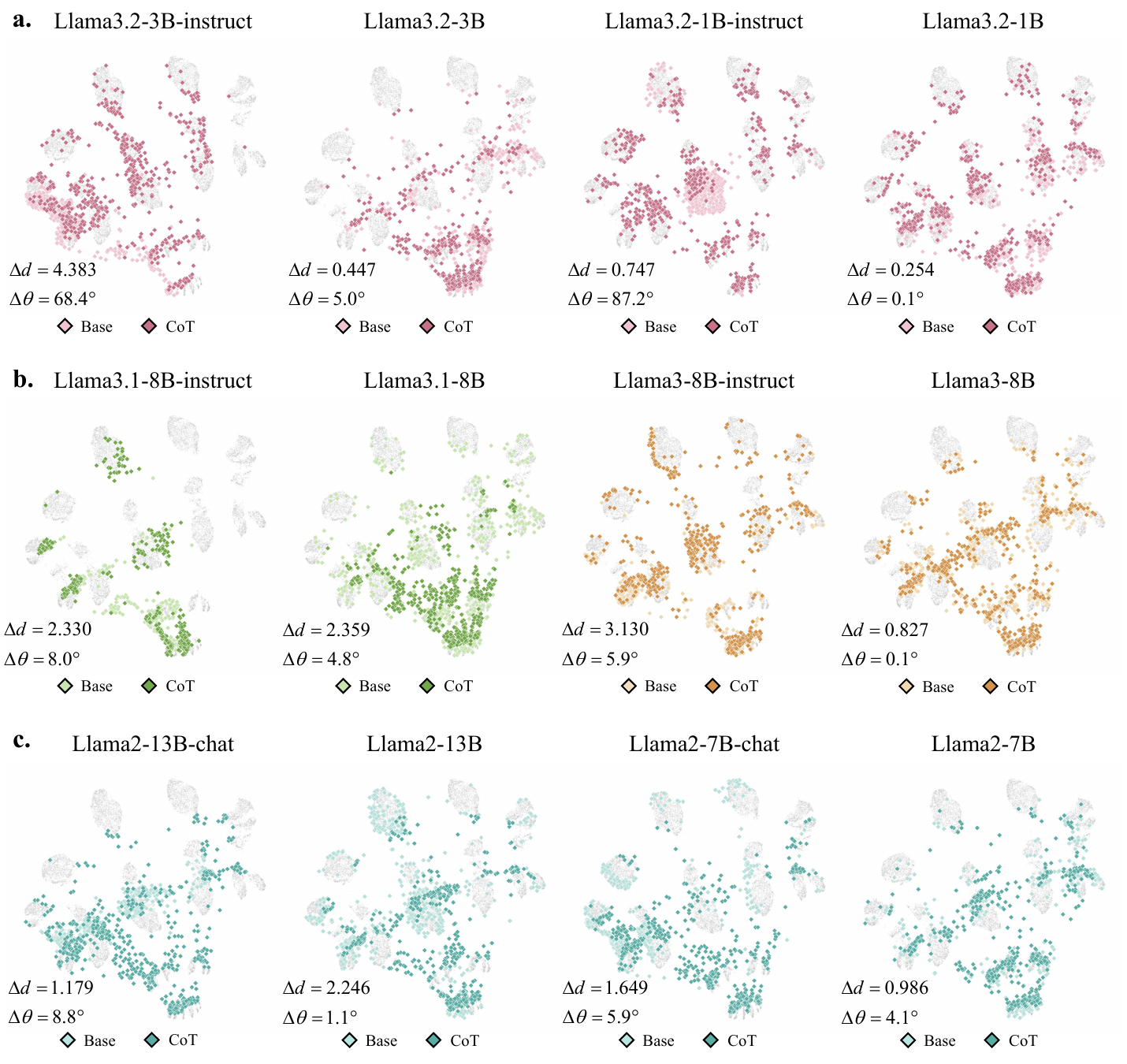} 
    \caption{\textbf{Effect of Chain-of-Thought prompting on value distributions across Llama model families.}
    Each subplot shows the UMAP projection of value vectors for a single model under two conditions:
    Base (hollow diamond, light color) and Chain-of-Thought prompting (CoT, filled diamond, dark color).
    Row \textbf{a} shows Llama3.2 models (3B-instruct, 3B, 1B-instruct, 1B);
    row \textbf{b} shows Llama3.1 and Llama3 models (3.1-8B-instruct, 3.1-8B, 3-8B-instruct, 3-8B);
    row \textbf{c} shows Llama2 models (13B-chat, 13B, 7B-chat, 7B).
    Grey points represent the WVS-7 human baseline.}
    \label{fig:appendix_cot_llama}
\end{figure}

\begin{table}[t]
\centering
\small
\caption{Effect of chain-of-thought prompting on the Llama family.}
\label{tab:llama_cot}

\begin{tabular}{llcccccc}
\toprule
Series & Model & $d_{\text{direct}}$ & $d_{\text{CoT}}$ & $\Delta d$ & $\theta_{\text{direct}}$ & $\theta_{\text{CoT}}$ & $\Delta\theta$ \\
\midrule

\multirow{2}{*}{Llama1}
& 7B  & 2.873 & 4.549 & 1.676 & 10.2 & 11.8 & 1.6 \\
& 13B & 4.141 & 3.741 & 0.400 & 16.9 & 12.1 & 4.7 \\

\midrule

\multirow{4}{*}{Llama2}
& 7B        & 4.023 & 3.037 & 0.986 & 9.5  & 13.6 & 4.1 \\
& 7B-Chat   & 3.843 & 2.195 & 1.649 & 10.9 & 5.1  & 5.9 \\
& 13B       & 1.622 & 3.868 & 2.246 & 12.7 & 11.5 & 1.1 \\
& 13B-Chat  & 1.784 & 2.963 & 1.179 & 12.3 & 3.5  & 8.8 \\

\midrule

\multirow{2}{*}{Llama3}
& 8B          & 2.264 & 3.090 & 0.827 & 6.6 & 6.5 & 0.1 \\
& 8B-Instruct & 4.872 & 1.742 & 3.130 & 6.5 & 12.4 & 5.9 \\

% \midrule

\multirow{2}{*}{Llama3.1}
& 8B          & 1.704 & 4.063 & 2.359 & 6.1 & 11.0 & 4.8 \\
& 8B-Instruct & 7.736 & 5.405 & 2.330 & 12.6 & 4.6 & 8.0 \\

% \midrule

\multirow{4}{*}{Llama3.2}
& 1B          & 2.349 & 2.095 & 0.254 & 8.0  & 8.1  & 0.1 \\
& 1B-Instruct & 0.867 & 1.614 & 0.747 & 89.4 & 2.2  & 87.2 \\
& 3B          & 4.889 & 4.442 & 0.447 & 16.8 & 11.8 & 5.0 \\
& 3B-Instruct & 6.177 & 1.795 & 4.383 & 82.8 & 14.4 & 68.4 \\

\bottomrule
\end{tabular}
\end{table}

\begin{figure}[H]
    \centering
    \includegraphics[width=0.7\textwidth]{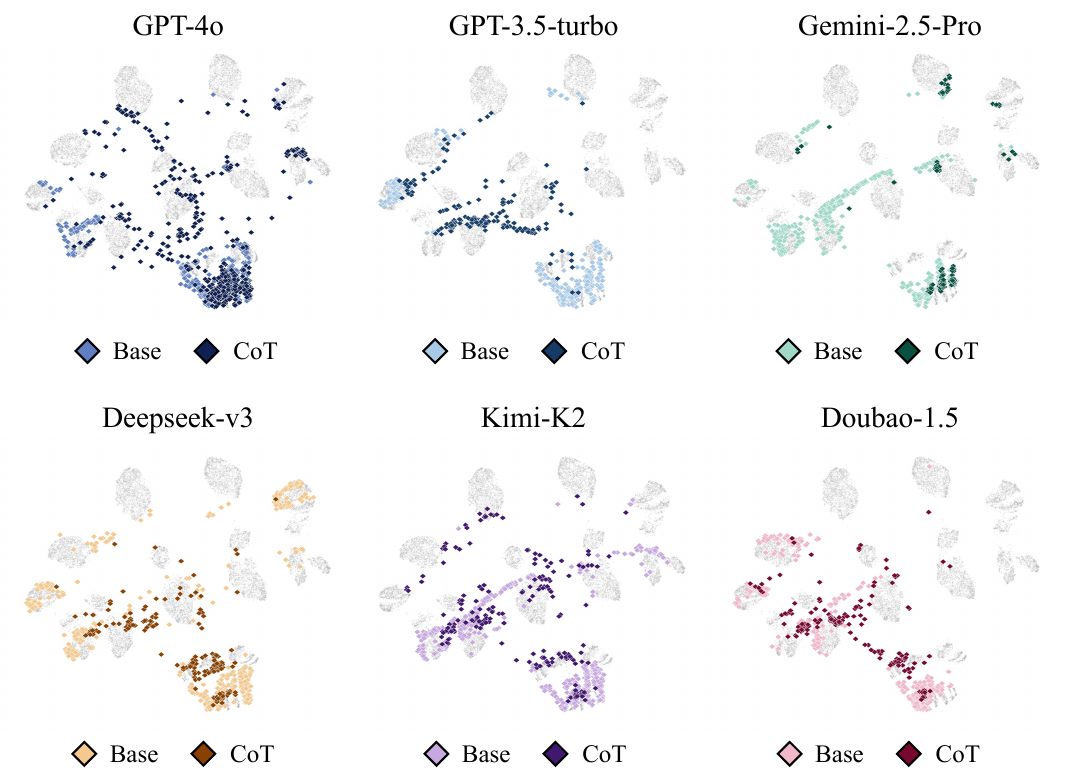} 
    \caption{\textbf{Effect of Chain-of-Thought prompting on value distributions across commercial API models.}
    Each subplot shows the UMAP projection of value vectors for a single model under two conditions:
    Base (hollow diamond, light color) and Chain-of-Thought prompting (CoT, filled diamond, dark color).
    Models include GPT-4o, GPT-3.5-Turbo, and Gemini 2.5 Pro (top row),
    and DeepSeek-V3, Kimi-K2, and Doubao-1.5 (bottom row).
    Grey points represent the WVS-7 human baseline.}
    \label{fig:appendix_cot_api}
\end{figure}

\subsection{Alignment Prescription and Intervention Strategies}
\label{app:prescription}

\begin{figure}[H]
    \centering
    \includegraphics[width=\textwidth]{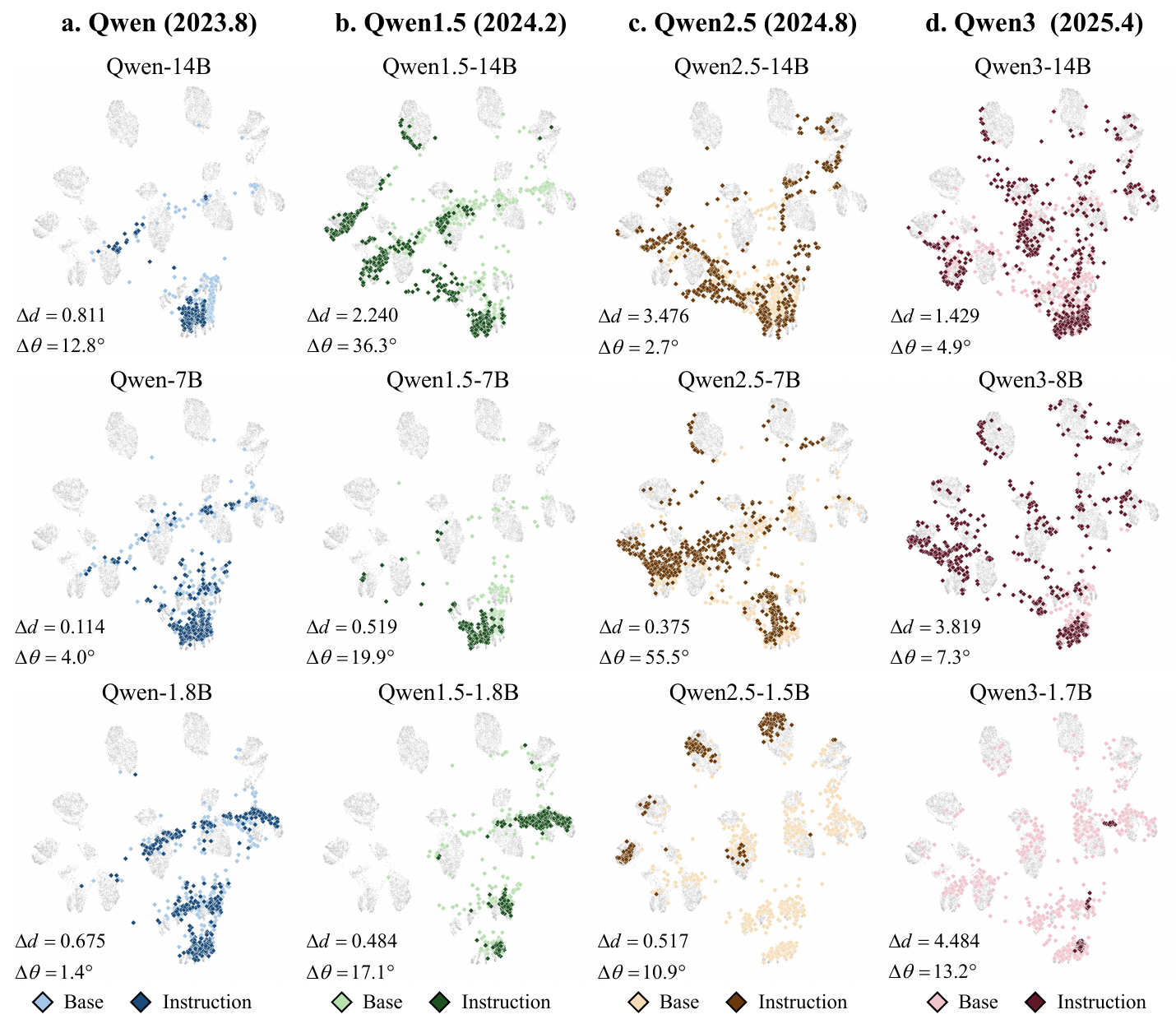}
    \caption{\textbf{Effect of instruction tuning on value distributions across Qwen model families.}
    Each subplot shows the UMAP projection of value vectors for a single model under two conditions: Base (hollow diamond, light color) and instruction-tuned (Instruction, filled diamond, dark color). Columns correspond to model generations: \textbf{a}, Qwen (2023.8); \textbf{b}, Qwen1.5 (2024.2); \textbf{c}, Qwen2.5 (2024.8); \textbf{d}, Qwen3 (2025.4). Rows correspond to parameter scales ($\sim$14B, $\sim$7--8B, $\sim$1.5--1.8B) from top to bottom. Grey points represent the WVS-7 human baseline.}
    \label{fig:qwen_sft_value_shift}
\end{figure}

\begin{table*}[t]
\centering
\small
\caption{Geometric displacement between base and instruction-tuned models in the Qwen family, corresponding to Fig.~\ref{fig:qwen_sft_value_shift}. $\theta$ values are in degrees.}

\begin{tabular}{llcccccc}
\toprule
Families & Model & $d_{\text{base}}$ & $d_{\text{align}}$ &
$\Delta d$ & $\theta_{\text{base}}$ & $\theta_{\text{align}}$ &
$\Delta\theta$ \\
\midrule
\multirow{3}{*}{Qwen1}
& 1.8B & 6.081 & 5.407 & 0.675 & 25.8 & 27.1 & 1.4 \\
& 7B   & 5.424 & 5.538 & 0.114 & 20.5 & 16.5 & 4.0 \\
& 14B  & 6.119 & 6.930 & 0.811 & 19.6 & 6.8  & 12.8 \\
\midrule

\multirow{5}{*}{Qwen1.5}
& 0.5B & 1.383 & 5.234 & 3.851 & 4.3 & 15.9 & 11.6 \\
& 1.8B & 6.589 & 7.073 & 0.484 & 25.5 & 42.6 & 17.1 \\
& 4B   & 5.015 & 2.523 & 2.492 & 19.6 & 1.2 & 18.4 \\
& 7B   & 7.112 & 6.593 & 0.519 & 24.1 & 4.2 & 19.9 \\
& 14B  & 3.106 & 5.346 & 2.240 & 8.7 & 45.1 & 36.3 \\
\midrule

\multirow{5}{*}{Qwen2.5}
& 0.5B & 2.221 & 0.704 & 1.517 & 6.9 & 3.6 & 3.4 \\
& 1.5B & 3.027 & 2.510 & 0.517 & 13.2 & 2.3 & 10.9 \\
& 3B   & 1.810 & 5.130 & 3.321 & 8.9 & 13.8 & 4.9 \\
& 7B   & 4.072 & 4.448 & 0.375 & 8.6 & 64.1 & 55.5 \\
& 14B  & 6.379 & 2.903 & 3.476 & 5.5 & 8.2 & 2.7 \\
\midrule
\multirow{5}{*}{Qwen3}
& 0.6B & 2.406 & 10.603 & 8.197 & 9.0 & 65.9 & 56.9 \\
& 1.7B & 1.255 & 5.739  & 4.484 & 2.1 & 15.3 & 13.2 \\
& 4B   & 6.361 & 0.851  & 5.510 & 0.2 & 5.2 & 5.1 \\
& 8B   & 5.621 & 1.802  & 3.819 & 18.7 & 26.1 & 7.3 \\
& 14B  & 3.853 & 2.424  & 1.429 & 0.1 & 5.0 & 4.9 \\
\bottomrule
\end{tabular}

\label{tab:qwen_results}
\end{table*}

\clearpage

\begin{figure}[H]
    \centering
    \includegraphics[width=\textwidth]{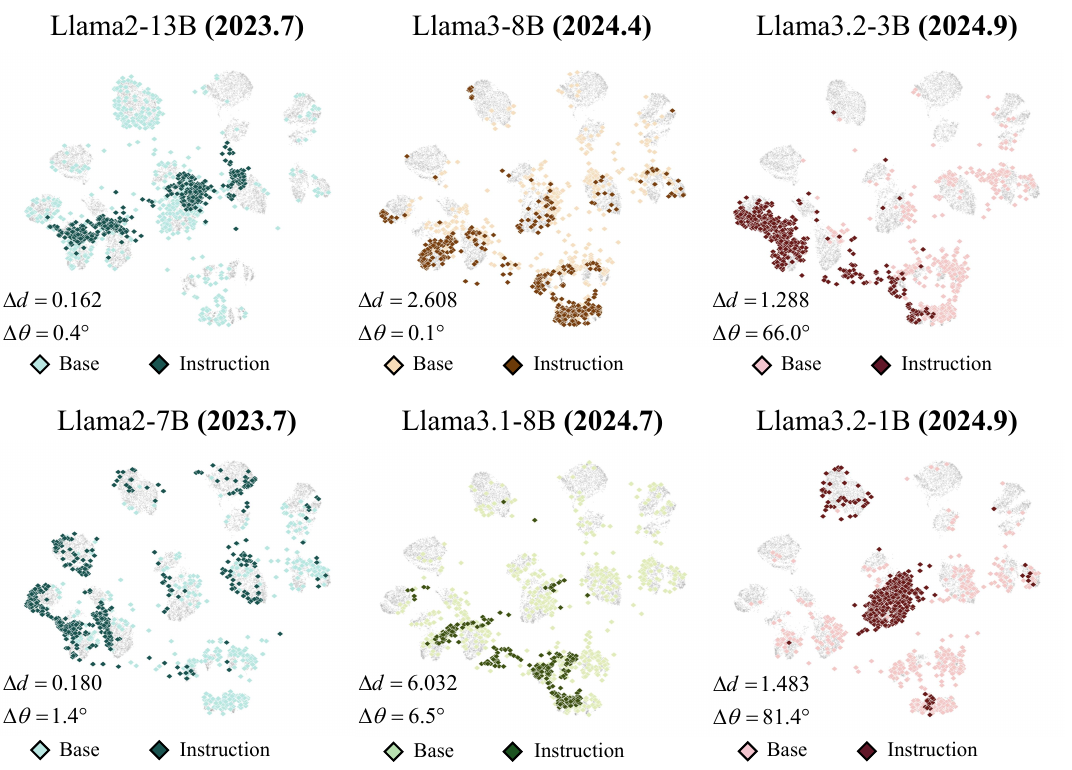} 
    \caption{\textbf{Effect of instruction tuning on value distributions across Llama model families.}
    Each subplot shows the UMAP projection of value vectors for a single model under two conditions: Base (hollow diamond, light color) and instruction-tuned (Instruction, filled diamond, dark color). The top row shows larger-scale models (Llama2-13B, Llama3-8B, Llama3.2-3B) and the bottom row shows smaller-scale models (Llama2-7B, Llama3.1-8B, Llama3.2-1B). Release dates are indicated in bold next to each model name. Grey points represent the WVS-7 human baseline.}
    \label{fig:llama_sft_value_shift}
\end{figure}

\begin{table*}[t]
\centering
\small
\caption{Geometric displacement between base and instruction-tuned models in the Llama family, corresponding to Fig.~\ref{fig:llama_sft_value_shift}. $\theta$ values are in degrees.}

\begin{tabular}{llcccccc}
\toprule
Families & Model & $d_{\text{base}}$ & $d_{\text{align}}$ &
$\Delta d$ & $\theta_{\text{base}}$ & $\theta_{\text{align}}$ &
$\Delta\theta$ \\
\midrule

\multirow{2}{*}{Llama1}
& 13B & 4.141 & $-$ & $-$ & 16.9 & $-$ & $-$ \\
& 7B  & 2.873 & $-$ & $-$ & 10.2 & $-$ & $-$ \\

\midrule

\multirow{2}{*}{Llama2}
& 13B & 1.622 & 1.784 & 0.162 & 12.7 & 12.3 & 0.4 \\
& 7B  & 4.023 & 3.843 & 0.180 & 9.5 & 10.9 & 1.4 \\

\midrule
Llama3 & 8B      & 2.264 & 4.872 & 2.608 & 6.6  & 6.5  & 0.1 \\
Llama3.1 & 8B  & 1.704 & 7.736 & 6.032 & 6.1  & 12.6 & 6.5 \\
\multirow{2}{*}{Llama3.2}
& 1B  & 2.349 & 0.867 & 1.483 & 8.0  & 89.4 & 81.4 \\
& 3B  & 4.889 & 6.177 & 1.288 & 16.8 & 82.8 & 66.0 \\

\bottomrule
\end{tabular}
\label{tab:llama_align_results}
\end{table*}

The generational trend is consistent across all three model families (Fig.~\ref{fig:generation_evolution}, Section~\ref{subsec:2.3.3-alignment}). As shown in Tables~\ref{tab:llama_align_results} and~\ref{tab:qwen_results}, for Llama, the shift magnitude $\Delta d$ increases substantially from Llama2 ($0.162$ for 13B; $0.180$ for 7B) to Llama3.1-8B ($6.032$), indicating that later alignment procedures impose a progressively stronger constraint on value distributions. The directional shift angle $\theta$ reveals a clear generational pattern: Llama3.2 models show markedly larger angular deviation after alignment ($89.4^\circ$ for 1B; $82.8^\circ$ for 3B), while earlier Llama2 and Llama3 models remain directionally stable ($\Delta\theta < 7^\circ$). For Qwen, $\Delta d$ grows from Qwen2.5 (ranging from $0.375$ to $3.476$) to Qwen3 (ranging from $1.429$ to $8.197$), with Qwen3-0.6B showing the largest shift ($\Delta d = 8.197$). The intra-generational divergence in Qwen3 is also evident in $\theta_{\text{align}}$: the 4B model remains near zero ($5.2^\circ$), while the 0.6B model deviates substantially ($65.9^\circ$).

\section{Prescription Assignment Algorithm}
\label{app:assignment}

For a given model with parameter set~$\boldsymbol{\theta}$ (not to be confused with the UMAP-projected shift angle $\theta$ used in the main text) and a target value dimension~$d$, the prescription procedure evaluates four intervention levels in sequence. The intervention shifts $\Delta_\ell^{(d)}$ defined below correspond to the operationalized effect sizes $\delta_E^{(d)}$, $\delta_C^{(d)}$, and $\delta_P^{(d)}$ introduced in Section~\ref{sec:effect_sizes}. The prescription level $\ell^*(d)$ is assigned via the effect-per-cost ranking described in Section~\ref{sec:level-assignment}, with the sequential threshold test below serving as a simplified equivalent when the cost ordering is monotonically increasing. At each level~$\ell$, the absolute value shift $\Delta_\ell^{(d)}$ is computed relative to the model's baseline output on dimension~$d$. Formally, the shift at each level is defined as:
\begin{align}
    \Delta_1^{(d)} &= \left| v^{(d)}_{\text{prompt}} - v_P^{(d)} \right| \\
    \Delta_2^{(d)} &= \left| v^{(d)}_{\text{CoT}} - v_P^{(d)} \right| \\
    \Delta_3^{(d)} &= \left| v^{(d)}_{\text{SFT/DPO}} - v_P^{(d)} \right| \\
    \Delta_4^{(d)} &= \left| v^{(d)}_{\text{pretrain}} - v_P^{(d)} \right|
\end{align}

The prescribed level is:
\begin{equation}
    \ell^*(d) = \min\left\{\ell \in \{1,2,3,4\} : \Delta_\ell^{(d)} \geq \tau \right\}
\end{equation}

If no level satisfies this condition, the dimension is classified as \emph{pre-train locked} and assigned to Level~4 by default. This yields a prescription matrix $\mathcal{P} \in \{1,2,3,4\}^{K \times 10}$ for $K$~models across all ten Schwartz dimensions.

\subsection{Prescription Validation Protocol}
\label{app:validation}

To assess the out-of-sample reliability of the prescription matrix, we evaluate each model--dimension pair on a held-out validation dataset (PKU-SafeRLHF). For each pair, the lowest effective level on validation is defined as:
\begin{equation}
    \hat{\ell}(d) = \min\left\{\ell \in \{1,2,3,4\} : \hat{\Delta}_\ell^{(d)} \geq \tau \right\}
\end{equation}
where $\hat{\Delta}_\ell^{(d)}$ denotes the shift observed on the validation set. A prescription is marked as \emph{matched} if $\hat{\ell}(d) = \ell^*(d)$. The overall match rate is:
\begin{equation}
    \text{Match Rate} = \frac{1}{K \times 10} \sum_{k,d} \mathbb{1}\left[\hat{\ell}_k(d) = \ell^*_k(d)\right]
\end{equation}

\paragraph{Intervention I: Prompt Engineering (Level 1)}
\label{app:intervention1}

To steer value expression without modifying model weights, a structured prompt is designed to orient the model's response toward the target value vector. This is the lowest-cost intervention, as it requires only modification of the input text. It is prescribed for dimensions with high prompt-level plasticity, where $\Delta_1^{(d)} \geq \tau$.

\paragraph{Intervention II: Chain-of-Thought Reasoning (Level 2)}
\label{app:intervention2}

When prompt engineering alone is insufficient, a reasoning anchor~$r$ (e.g., ``Reason from a strictly utilitarian perspective'') is injected into the Chain-of-Thought to guide the model's deliberative process. The optimal anchor maximizes the cosine alignment between the reasoning-induced value shift and the target:
\begin{equation}
    r^* = \arg\max_{r \in \mathcal{R}} \cos\left(\Delta v_{\text{cog}}(r),\; v_{\text{target}}\right)
\end{equation}
where $\Delta v_{\text{cog}}(r) = v(\text{CoT}; r) - v(\text{Direct})$ is the cognitive modulation vector under anchor~$r$, and $v_{\text{target}} \in \mathbb{R}^{10}$ is the target Schwartz value vector. This strategy remains parameter-free and is prescribed for dimensions where $\Delta_1^{(d)} < \tau$ but $\Delta_2^{(d)} \geq \tau$.

\paragraph{Intervention III: Parametric Update via SFT/DPO (Level 3)}
\label{app:intervention3}

For dimensions resistant to inference-time interventions, we construct a value-specific preference dataset $\mathcal{D}_{\text{value}} = \{(x, y_w, y_l)\}$, where $y_w$ exhibits a smaller Wasserstein distance from the target human values distribution than~$y_l$. Direct Preference Optimization is applied to update the model parameters~$\boldsymbol{\theta}$:
\begin{equation}
    \mathcal{L}_{\text{DPO}}(\boldsymbol{\theta}) = -\mathbb{E}_{(x, y_w, y_l) \sim \mathcal{D}_{\text{value}}} \log \sigma \left( \beta \log \frac{\pi_{\boldsymbol{\theta}}(y_w | x)}{\pi_{\text{ref}}(y_w | x)} - \beta \log \frac{\pi_{\boldsymbol{\theta}}(y_l | x)}{\pi_{\text{ref}}(y_l | x)} \right)
\end{equation}

This strategy permanently updates model parameters and is prescribed for dimensions where $\Delta_1^{(d)} < \tau$, $\Delta_2^{(d)} < \tau$, but $\Delta_3^{(d)} \geq \tau$.

\paragraph{Intervention IV: Continued Pre-training (Level 4)}
\label{app:intervention4}

When all preceding levels fail to exceed~$\tau$, the dimension is classified as \emph{pre-train locked}. Realignment requires continued pre-training on a curated corpus~$\mathcal{D}_{\text{pretrain}}$ enriched with texts aligned to the target value orientation:
\begin{equation}
    \mathcal{L}_{\text{PT}}(\boldsymbol{\theta}) = -\mathbb{E}_{x \sim \mathcal{D}_{\text{pretrain}}} \log \pi_{\boldsymbol{\theta}}(x)
\end{equation}

Corpus curation follows a two-step procedure. First, candidate documents are scored according to their geometric proximity to $v_{\text{target}}$ in the Schwartz space, computed via the mapping function~$\phi$. Second, documents whose value vectors fall within a radius~$\epsilon$ of $v_{\text{target}}$ are retained, while documents reinforcing the undesired orientation are downsampled. This targeted corpus construction ensures that the pre-training signal shifts the value distribution in the intended direction, without introducing broad distributional noise. This strategy is applied only after the prescription validation procedure confirms that no lower-cost intervention is effective.

\subsection{Comparative Metric: Effect-per-Cost Efficiency}
\label{app:efficiency}

To enable systematic comparison across the three PEC intervention classes, we define an effect-per-cost ratio for $\mathcal{I}\in\{E,C,P\}$, corresponding to Environment, Cognition, and Prior interventions. Here, $\mathcal{I}=P$ includes both parametric intervention types considered in our hierarchy: SFT/DPO at Level~3 and continued pre-training at Level~4.

\begin{equation}
    \eta(\mathcal{I}) =
    \frac{\|v_{\mathrm{after}}-v_{\mathrm{before}}\|_2}
    {\mathrm{Cost}(\mathcal{I})},
    \qquad \mathcal{I}\in\{E,C,P\}
\end{equation}

The intervention cost is converted into a common FLOP-based unit before comparison. For Environment intervention ($\mathcal{I}=E$, Level~1 Prompt), the computational cost is considered negligible relative to other intervention classes. For Cognition intervention ($\mathcal{I}=C$, Level~2 CoT), the additional inference cost is approximated as

\begin{equation}
    \mathrm{Cost}(C)
    \approx 2N_{\mathrm{params}}\cdot n_{\mathrm{tokens}},
\end{equation}

where $n_{\mathrm{tokens}}$ denotes the number of additional generated tokens introduced by CoT elicitation.

For Prior intervention ($\mathcal{I}=P$), including SFT/DPO (Level~3) and continued pre-training (Level~4), the training cost is approximated as

\begin{equation}
    \mathrm{Cost}(P)
    \approx 6N_{\mathrm{params}}\cdot N_{\mathrm{train}},
\end{equation}

where $N_{\mathrm{train}}$ denotes the number of training tokens processed. These FLOP approximations follow standard scaling-law estimation practices~\cite{kaplan2020scaling}.

The resulting efficiency metric provides a normalized measure of how much value displacement is achieved per unit computational cost. It enables comparison across intervention classes and establishes a Pareto frontier for selecting cost-effective interventions. However, this metric is not used as the criterion for determining intervention levels; instead, intervention prescriptions are determined by the effectiveness analysis described in the main text. Per-model cost values and resulting efficiency scores are provided in the released code and data.

\end{appendices}

\end{document}